\documentclass[preprint,12pt]{elsarticle}

\usepackage{amssymb}
\usepackage{amsthm}

\usepackage{graphicx}
\graphicspath{{NDLS_paper_IEEE_resource/}}

\usepackage{array}
\usepackage[vlined,english,ruled,linesnumbered]{algorithm2e}
\usepackage{algpseudocode}
\usepackage{multirow}
\usepackage{subfigure}
\usepackage{amsmath,amsfonts}

\usepackage{url}

\theoremstyle{definition}
\newtheorem{defn}{Definition}[section]
\newtheorem*{defn*}{Definition}

\usepackage{color}

\journal{}
\begin{document}

\begin{frontmatter}



\title{Multi-objectivization Inspired Metaheuristics for the Sum-of-the-Parts Combinatorial Optimization Problems}


\author[addxjtu]{Jialong Shi}
\ead{jialong.shi@xjtu.edu.cn}

\author[addxjtu]{Jianyong Sun \corref{correspondingauthor}}
\cortext[correspondingauthor]{Corresponding author}
\ead{jy.sun@xjtu.edu.cn}

\author[addcityu]{Qingfu Zhang}
\ead{qingfu.zhang@cityu.edu.hk}

\address[addxjtu]{School of Mathematics and Statistics, Xi'an Jiaotong University, Xi'an, China}

\address[addcityu]{Department of Computer Science, City University of Hong Kong, Hong Kong SAR}

\begin{abstract}
  Multi-objectivization is a term used to describe strategies developed for optimizing single-objective problems by multi-objective algorithms. This paper focuses on multi-objectivizing the sum-of-the-parts combinatorial optimization problems, which include the traveling salesman problem, the unconstrained binary quadratic programming and other well-known combinatorial optimization problem. For a sum-of-the-parts combinatorial optimization problem, we propose to decompose its original objective into two sub-objectives with controllable correlation. Based on the decomposition method, two new multi-objectivization inspired single-objective optimization techniques called \emph{non-dominance search} and \emph{non-dominance exploitation} are developed, respectively. Non-dominance search is combined with two metaheuristics, namely iterated local search and iterated tabu search, while non-dominance exploitation is embedded within the iterated Lin-Kernighan metaheuristic. The resultant metaheuristics are called \emph{ILS+NDS}, \emph{ITS+NDS} and \emph{ILK+NDE}, respectively. Empirical studies on some TSP and UBQP instances show that with appropriate correlation between the sub-objectives, there are more chances to escape from local optima when new starting solution is selected from the non-dominated solutions defined by the decomposed sub-objectives. Experimental results also show that ILS+NDS, ITS+NDS and ILK+NDE all significantly outperform their counterparts on most of the test instances.
\end{abstract}


\end{frontmatter}


\section{Introduction}

The so-called multi-objectivization approach deals with a single-objective problem by converting it into a multi-objective one and then optimizing it by a multi-objective algorithm. Many related studies~\cite{knowles2001reducing,segura2016using,ishibuchi2007optimization,handl2008multiobjectivization,mouret2011novelty,garza2015multi} have confirmed that the idea of multi-objectivization can achieve better performance than some traditional single-objective algorithms. 

This paper differs from the existing multi-objectivization studies in two aspects.
\begin{itemize}
  \item This paper focuses on multi-objectivizing an important subclass of the Combinatorial Optimization Problems (COPs)~\cite{reeves1993modern}, the \emph{sum-of-the-parts COPs}, through decomposing its original objective into two sub-objectives.
  \item Different from most multi-objectivization studies which directly optimize the sub-objectives using multi-objective optimization algorithms, This paper uses the sub-objectives to help search-based single-objective metaheuristics\footnote{A search-based metaheuristic is an iterative optimization algorithm for COPs. At each iteration, it searches for a better solution in the neighborhood of current solution. If a local optimum is found, it tries to escape from it by starting the search from a new point.} escape from local optima and find better solutions.
\end{itemize}
In a sum-of-the-parts COP, its objective function can be represented as the summation of a finite number of sub-functions over unit costs. The well-known Traveling Salesman Problem (TSP)~\cite{applegate2006traveling},  Unconstrained Binary Quadratic Programming (UBQP)~\cite{kochenberger2014unconstrained} problem,  Quadratic Assignment Problem (QAP)~\cite{lawler1963quadratic} and Vehicle Routing Problem (VRP)~\cite{toth2002vehicle} all belong to this type. We propose to decompose a sum-of-the-parts COP's objective function $f$ into two sub-objectives $f_1$ and $f_2$ by splitting each unit cost of the problem into two parts. The decomposed $f_1$ and $f_2$ are subject to $f(x) = f_1(x) + f_2(x)$ for any solution $x$ in the solution space. The cost splitting follows a probability distribution, while the sub-objectives' correlation can be controlled by varying the probability distribution.

Based on the proposed objective decomposition method, two new multi-objectivization inspired techniques are developed, both of which are based on the idea that the sub-objectives can help search-based single-objective metaheuristics escape from local optima and find better solutions. The first multi-objectivization inspired method we propose is called \emph{Non-Dominance Search} (NDS) which is suitable to be combined within search-based metaheuristics with fixed neighborhood structure. The idea behind NDS is similar to that of Variable Neighborhood Search (VNS)~\cite{mladenovic1997variable} which enlarges the neighborhood size when a local optimum is found. NDS also tries to find better solutions in the neighborhoods of the neighboring solutions of the local optimum. Instead of searching the neighborhood's neighborhood exhaustively, NDS assumes that the improved solutions are more likely to be found in the neighborhood of a \emph{non-dominated neighboring solution} of the current local optimum $x_*$. We call this hypothesis the ``neighborhood non-dominance'' hypothesis. Here a non-dominated neighboring solution of $x_*$ is the neighboring solution that is non-dominated to $x_*$ in terms of $(f_1, f_2)$. We denote the set of all the non-dominated neighboring solutions of $x_*$ as ${\cal N}(x_*\mid f_1, f_2)$. When trapping in a local optimum $x_*$, NDS will only search the neighborhoods of the solutions in ${\cal N}(x_*\mid f_1, f_2)$, looking for an improving solution. If no improvement is found, NDS returns $x_*$.

To verify the hypothesis of NDS, we carry out empirically study on two kinds of the sum-of-the-parts COPs, namely, the TSP and the UBQP. Empirical results confirm that the hypothesis holds for all the considered TSP and UBQP instances. Further, it is found that the effectiveness of the decomposition depends highly on the correlation between the two sub-objectives.

A limitation of NDS is that it requires that the neighborhood structure is fixed, i.e., given a solution, one can list all the neighboring solutions. However, NDS cannot be combined within a local search with varied neighborhood structure, e.g., the Lin-Kernighan (LK) local search~\cite{lin1973effective} for the TSP. To overcome this problem, we propose another multi-objectivization inspired technique, called \emph{Non-Dominance Exploitation} (NDE). NDE looks for local optima that are non-dominated to the current best solution. The search region close to these local optima will be further exploited.

In this paper, we combine NDS with the well-known Iterated Local Search (ILS)~\cite{lourencco2010iterated} and Iterated Tabu Search (ITS)~\cite{palubeckis2006iterated}, while we combine NDE with the Iterated Lin-Kernighan algorithm (ILK)~\cite{johnson1997traveling,applegate2003chained}. The resultant algorithms are called ILS+NDS, ITS+NDS and ILK+NDE, respectively.
In the experimental studies, we compare ILS+NDS against the basic ILS and a variant of ILS+NDS (in which the guidance of $(f_1, f_2)$ is eliminated) on some TSP instances and UBQP instances, compare ITS+NDS against the basic ITS on some UBQP instances and compare ILK+NDE against the basic ILK and a variant of ILK+NDE (in which the guidance of $(f_1, f_2)$ is eliminated) on some middle- and large-size TSP instances. In addition, in the implementations of ILS+NDS, ITS+NDS and ILK+NDE, different levels of sub-objectives' correlation are tested. The experimental results show that ILS+NDS, ITS+NDS and ILK+NDE all significantly outperform their counterparts on most of the test instances with a proper value of correlation coefficient between the sub-objectives $(f_1, f_2)$.

Preliminary work of this paper has been published in a conference~\cite{shi2019multi}. This paper differs significantly from the conference version in the following aspects.
\begin{itemize}
  \item In this paper, the decomposition method is generalized to all the sum-of-the-parts COPs. Besides the TSP, the UBQP is also used as testbed.
  \item A method to control the correlation between the sub-objectives $f_1$ and $f_2$ is proposed.
  \item Systematic experiments are carried out to analyze the neighborhood non-dominance hypothesis.
  \item A new multi-objectivization inspired technique, NDE, is proposed for local search metaheuristics with varied neighborhood structure.
\end{itemize}

The rest of the paper is organized as follows. Section~\ref{sec:relate} presents the related work. Section~\ref{sec:decomp} formalizes the sum-of-the-parts COP and introduces the proposed objective decomposition method. The TSP and the UBQP are used to illustrate the procedure of the decomposition method. In Section~\ref{sec:new_methods}, two new multi-objectivization inspired techniques, NDS and NDE, based on the proposed objective decomposition method, are presented. Section~\ref{sec:epm} presents the empirical study on the neighborhood non-dominance hypothesis. Experimental results of the proposed methods for the TSP and the UBQP instances are also presented in Section~\ref{sec:epm}. Section~\ref{sec:conclu} concludes the paper and discusses future work.

\section{Related Work}\label{sec:relate}

The study of multi-objectivization can be dated back to 2001 when Knowles et al.~\cite{knowles2001reducing} first invented the term ``multi-objectivization''. In their work, a continuous optimization problem and the TSP were used as the testbeds. The authors proposed to decompose the objective function of the TSP by cutting a tour into two sub-tours. Their experimental results showed that the multi-objective algorithm can return better solutions compared to a broadly equivalent single-objective algorithm. The authors claimed that this is because the multi-objectivization technique reduced the number of local optima in the search space. Ishibuchi and Nojima~\cite{ishibuchi2007optimization} focused on single-objective problems which are in the form of a scalarizing function. They decomposed the original objective into sub-objectives that are similar to the scalarizing function and found that Evolutionary Multi-objective Optimization (EMO) helps a local solver escape from local optima.

The aforementioned work found that multi-objectivization is beneficial to single-objective optimization. However, Handl et al.~\cite{handl2008multiobjectivization} argued that multi-objectivization through decomposition can equally render a single-objective optimization problem easier or harder, since the incomparable nature of multiple objectives creates plateaus in the fitness landscape which may reduce the number of local optima, but may hinder the search. J{\"a}hne et al.~\cite{jahne2009evolutionary} proposed the so-called Multi-Objectivization via Segmentation (MOS) method for the TSP. In MOS, the original objective is decomposed into two sub-objectives by defining two sets of cities. The decomposition multi-objectivization method has also been successfully applied to logic circuit design~\cite{coello2002design}, sorting and shortest paths problem~\cite{scharnow2005analysis}, robotic control~\cite{mouret2011novelty}, protein structure prediction~\cite{garza2015multi} and others.

Compared to the decomposition-based multi-objectivization methods, much work have been carried out on using helper objectives to multi-objectivize a single-objective problem. Helper objectives are additional objectives that are optimized simultaneously with the original objective to ``maintain diversity in the population, guide the search away from local optima, or help the creation of good building blocks''~\cite{jensen2003guiding}. For constrained optimization problems, it is a common practice to transform the constraints into a helper objective. There has an extensive study of constrained optimization using EMO, see e.g.~\cite{coello2000treating, runarsson2005search, mezura2008constrained, singh2009performance, churchill2013multi, ferrandez2019preference, xu2019helper}.

For continuous problems, the helper objectives are usually related to the properties of the solution population or the properties of the problem. For example, Abbass and Deb~\cite{abbass2003searching} used the age of individual solution as an additional objective to maintain the population diversity and to slow down the selection pressure. Jiao et al.~\cite{jiao2017dynamic} converted a single-objective problem to a dynamic multi-objective problem by considering a niche-count objective to maintain the diversity. Deb and Saha~\cite{deb2012multimodal} converted a multi-modal problem into a bi-objective problem by considering the gradient or neighborhood information as the second objective.

For COPs, some studies defined the helper objectives based on the segments of the original objective. For the Job Shop Scheduling Problem (JSSP), Jensen~\cite{jensen2003guiding, jensen2004helper} used the flow-times of individual jobs as the helper objectives which are designed to dynamically change during the search in a random order. Experimental results showed that using NSGA-II~\cite{deb2002fast} to optimize the generated multi-objective problem significantly outperforms using a traditional GA. Syberfeldt and Rogstrom~\cite{syberfeldt2018two} proposed a two-step multi-objectivization method. In the first step, the helper objective was set to conflict with the original objective and in the second step the helper objective was in harmony with the original objective. Alsheddy~\cite{alsheddy2018penalty} proposed a helper objective by a penalty-based approach. Bleuler et al.~\cite{bleuler2008reducing} tried to reduce the `bloat' phenomenon in genetic programming by considering the program size as a second objective.

Lochtefeld and Ciarallo~\cite{lochtefeld2010deterministic, lochtefeld2011helper, lochtefeld2011multiobjectivization, lochtefeld2014analysis, lochtefeld2015multi} conducted a series of studies on helper objective based and decomposition-based methods. They found that problem-specific knowledge should be incorporated for a good helper objective sequence. In a recent study~\cite{lochtefeld2015multi}, they showed that the decomposition-based method has a better on-average performance compared to the helper objective method on their tested JSSP instances. Brockhoff et al.~\cite{brockhoff2007additional,brockhoff2009effects} showed that a multi-objectivized problem can become either harder or easier depending on the definitions of the helper objectives.

In addition to the previous studies, multi-objectivization has been proved to be helpful for dealing with the timetabling problem ~\cite{wright2001subcost}, dynamic environment problem~\cite{bui2005multiobjective}, minimum spanning tree problem ~\cite{neumann2006minimum,neumann2008can}, computational mechanics design problem~\cite{greiner2007improving}, chemical phase-equilibrium detection problem~\cite{preuss2007solving}, short-term unit commitment problem~\cite{trivedi2012multi}, compliant mechanism design problem~\cite{sharma2014customized} and other problems.

From the above literature review we observe that (1) few studies have been tried to develop a universal multi-objectivization method for a wide range of problems, and (2) existing EMO algorithms are applied in most studies. In this paper, we propose a universal objective decomposition method which is suitable for the sum-of-the-parts COPs. Further, based on the decomposition method, we propose two new techniques which can improve the global search ability of local-search-based metaheuristics. The well-known local-search-based metaheuristics includes ILS~\cite{lourencco2010iterated}, ITS~\cite{palubeckis2006iterated}, VNS~\cite{mladenovic1997variable}, Guided Local Search (GLS)~\cite{voudouris1999guided,shi2018eb}, etc. Generally speaking, those metaheuristics alternately execute a local search process and an escaping process. The basic procedure of ILS is shown in Algorithm~\ref{alg:ILS}, in which LocalSearch($x$) means starting a local search process from $x$ until a local optimum is encountered and output this local optimum. Perturbation($x_*$) means perturbing some parts of a local optimum $x_*$ so that the resulting solution is no longer in the attraction basin of the original local optimum. The basic procedure of ITS is shown in Algorithm~\ref{alg:ITS}. The framework of ITS is similar to that of ILS, except that in ITS the local search heuristic is replaced by the tabu search heuristic. VNS escapes from local optima by changing the neighborhood structure of the search (usually changing to a larger neighborhood) so that the original local optimum is no longer locally optimal in the new neighborhood structure. In GLS, when the agent is in a local optimum, some features that appear in the local optimum are selected and penalized. Then the objective function is augmented by the accumulated penalties and guides the agent to move out of the attraction basin of this local optimum.

\begin{algorithm}
\small
    $x_0' \gets $ randomly or heuristically generated solution\;
    set $x_{best} \gets x_0'$ and $j \gets 0$\;
    \While{stopping criterion is not met}{
        $x_j \gets$ LocalSearch($x_j'$)\;
        \If {$f(x_{j}) < f(x_{best})$} {
            $x_{best} \gets x_{j}$\;
        }
        $x_{j+1}' \gets$ Perturbation($x_j$)\; \label{ils1}
        $j\gets j+1$\;
    }
    \KwRet{\mbox{the historical best solution} $x_{best}$}
\caption{Iterated Local Search (ILS)}
\label{alg:ILS}
\end{algorithm}

\begin{algorithm}
\small\
    $x_0' \gets $ randomly or heuristically generated solution\;
    set $x_{best} \gets x_0'$ and $j \gets 0$\;
    \While{stopping criterion is not met}{
        $x_j \gets$ TabuSearch($x_j'$)\;
        \If {$f(x_{j}) < f(x_{best})$} {
            $x_{best} \gets x_{j}$\;
        }
        $x_{j+1}' \gets$ Perturbation($x_j$)\;
        $j\gets j+1$\;
    }
    \KwRet{\mbox{the historical best solution} $x_{best}$}
\caption{Iterated Tabu Search (ITS)}
\label{alg:ITS}
\end{algorithm}

\section{Objective Decomposition}\label{sec:decomp}

A Combinatorial Optimization Problem (COP) is defined as
\begin{equation}\label{eq:COP}
  \begin{array}{rl}
    \mbox{minimize / maximize} & \ \ f(x)\\
    \mbox{subject to} & \ \ x\in\mathcal{S},\\
  \end{array}
\end{equation}where $f:\mathcal{S}\rightarrow\mathbb{R}$ is the objective function and $\mathcal{S}$ is the solution space which is a finite discrete set, e.g., an $n$-dimensional binary vector space $\{0,1\}^n$ for the UBQP or an $n$-dimensional permutation space $\mathcal{P}_n$ for the TSP. In this paper we focus on the sum-of-the-parts COPs. A sum-of-the-parts COP satisfies the following constraints:
\begin{itemize}
\item[(i)] The problem is uniquely determined by a finite discrete set of units $U=\{u_i\mid i=1,2,\dots,|U|\}$ and each unit $u_i$ has a fixed cost $c_i$;
\item[(ii)] A feasible solution $x$ is a subset of $U$ and satisfies certain rules of composition;
\item[(iii)] The objective function $f(x)$ is the summation (or weighted summation) of the costs of units in $x$.
\end{itemize}
Formally, the sum-of-the-parts COP can be expressed as
\begin{equation}\label{eq:COP_subclass}
  \begin{array}{rl}
    \mbox{minimize / maximize} & f(x) = \sum\limits_{i\in\{j\mid u_j\in x\}} c_i,\\
    \mbox{subject to} & x\subset U,\\ 
    & x~\mbox{satisfies certain composition rules,}\\
  \end{array}
\end{equation}where $c_i$ is the cost associated with $u_i$.

The well-known TSP belongs to the sum-of-the-parts COPs. In a TSP, the edges between any two cities form a finite set and each edge has a fixed travel cost. A TSP solution is a subset of edges that forms a tour visiting every city exactly once, then returning to the first city. The function value of a TSP solution is the total cost of the edges in the tour. Hence, in the TSP, the total edge set can be seen as the unit set $U$ and the edge costs can be seen as the unit costs. Besides the TSP, we can deduce that the UBQP, the quadratic assignment problem, the vehicle routing problem and the knapsack problem all belong to the sum-of-the-parts COPs.

For the sum-of-the-parts COPs, we propose a new method to decompose the original objective function $f$ into two sub-objective functions $f_1$ and $f_2$ such that $f(x) = f_1(x) + f_2(x)$ for any solution $x$ in the solution space. The proposed decomposition method is quite simple. For each unit $u_i$ in the finite set, the method splits its cost into two new values $c^{(1)}_i$ and $c^{(2)}_i$ such that $c_i = c^{(1)}_i + c^{(2)}_i$ following a probability distribution $p$. The decomposition is independent of the unit set, i.e., the splitting of all the unit costs follows the same probability distribution. As a result, $f_1$ and $f_2$ are defined by the new unit costs $\{c^{(1)}_i\mid i=1,2,\dots,|U|\}$ and $\{c^{(2)}_i\mid i=1,2,\dots,|U|\}$, respectively:
\begin{equation}\label{eq:f1}
  f_1(x) = \sum\limits_{i\in\{j\mid u_j\in x\}} c^{(1)}_i,~\mbox{and}
\end{equation}
\begin{equation}\label{eq:f2}
  f_2(x) = \sum\limits_{i\in\{j\mid u_j\in x\}} c^{(2)}_i.
\end{equation}
It is obvious that $f(x) = f_1(x) + f_2(x)$ for any $x \in U$.

\begin{figure}
  \centering
  \includegraphics[width=0.3\linewidth]{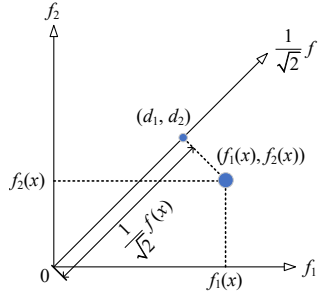}\\
  \caption{The relationship between $f$, $f_1$ and $f_2$. }\label{fig:f0f1f2} 
\end{figure}

The relationship between the original objective function $f$ and the two sub-objective functions $(f_1, f_2)$ can be illustrated in Figure~\ref{fig:f0f1f2}. In Figure~\ref{fig:f0f1f2}, we plot a new axis in the middle of the $f_1$ and $f_2$ axis. For any point $(f_1(x),f_2(x))$ in the bi-objective space, its projection on the middle axis is denoted as $(d_1,d_2)$. Then we can have $d_1 = d_2$ and $f_1(x) + f_2(x) = d_1 + d_2 = 2d_1$. The distance between $(d_1,d_2)$ and $(0,0)$ is $\sqrt{(d_1-0)^2+(d_2-0)^2}$ $ = \sqrt{2(d_1)^2}$ $ =\sqrt{2}|d_1|$ $ =\frac{1}{\sqrt{2}}\cdot 2|d_1|$ $ = \frac{1}{\sqrt{2}}|d_1+d_2|$ $ = \frac{1}{\sqrt{2}}|f_1(x)+f_2(x)| = \frac{1}{\sqrt{2}}|f(x)|$, which means the middle axis measures $\frac{1}{\sqrt{2}}f(x)$.

From Eq.~\ref{eq:f1} and Eq.~\ref{eq:f2}, it is seen that $f_1(x)$ and $f_2(x)$ are different only by their unit costs $\{c^{(1)}_i\mid i=1,2,\dots,|U|\}$ and $\{c^{(2)}_i\mid i=1,2,\dots,|U|\}$. We thus use their Pearson correlation coefficient to measure the correlation between the sub-objectives $f_1(x)$ and $f_2(x)$. The correlation coefficient is defined as:
\begin{equation}\label{eq:corr_coef}
  \rho = \frac{\mbox{cov}(\{c^{(1)}_i\},\{c^{(2)}_i\})}{\sigma(\{c^{(1)}_i\})\cdot \sigma(\{c^{(2)}_i\})},
\end{equation}
where $\mbox{cov}(\{c^{(1)}_i\},\{c^{(2)}_i\})$ is the covariance between the weight sets $\{c^{(1)}_i\mid i=1,2,\dots,|U|\}$ and $\{c^{(2)}_i\mid i=1,2,\dots,|U|\}$ which is calculated by
$$\mbox{cov}(\{c^{(1)}_i\},\{c^{(2)}_i\})=\frac{1}{|U| -1}\sum_{j=1}^{|U|} (c^{(1)}_j - \mu^{(1)})(c^{(2)}_j - \mu^{(2)}),$$
and $\sigma(\{c^{(k)}_i\}),~k=1,2$ is the standard deviation which is calculated by
$$\sigma(\{c^{(k)}_i\})=\sqrt{\frac{1}{|U| -1}\sum_{j=1}^{|U|} |c^{(k)}_j - \mu^{(k)}|^2}~,~~~~ k=1,2,$$
where $\mu^{(k)}=\frac{1}{|{U}| }\sum_{j=1}^{|U|}c^{(k)}_j ,~k=1,2,$ is the average value of $\{c^{(k)}_i\mid i=1,2,\dots,|U|\},~k=1,2$.


If $c^{(1)}_i = c^{(2)}_i = c_i/2$ for all $i\in\{1,2,\dots,|U|\}$, then we have $f_1(x)=f_2(x)=f(x)/2$ for any $x$ in the solution space and $\rho=1$. Conversely, if $|c^{(1)}_i - c^{(2)}_i|\gg0$ for all $i\in\{1,2,\dots,|U|\}$, then the probability that $|f_1(x)-f_2(x)|\gg0$ becomes relatively high and $\rho\to -1$. Hence, by controlling the ratio between $c^{(1)}_i$ and $c^{(2)}_i$, we can control the correlation between $f_1$ and $f_2$. 

In the following, we show how to apply the proposed decomposition method to the TSP and the UBQP respectively.

\subsection{Decomposition of the TSP}\label{sec:TSP_decomp}

Given $n$ cities and travel costs between every pair of cities, the TSP is to find the most cost-effective tour that visits every city exactly once and returns to the first city. Formally, let ${\cal G} =({\cal V}, {\cal E})$ be a fully connected graph with cities as vertexes, where $\cal V$ is the vertex set and $\cal E$ the edge set. Denote $c_{i,j}>0$ the cost of the edge $(i,j)$ between vertex $i$ and vertex $j$, the objective function of a TSP is defined as
\begin{equation}\label{eq:TSP}
\begin{split}
\mbox{minimize} ~&~ f(x)= \sum_{(i,j)\in x} c_{i,j},\\
\mbox{subject to} ~&~ x = \{(y_n,y_1),(y_1,y_2),(y_2,y_3),\dots,(y_{n-1},y_n)\}\subset {\cal E},\\
~&~ y=(y_1,y_2,\dots,y_n)\in \mathcal{P}_n,
\end{split}
\end{equation}
where a feasible solution $x$ is a set of edges defined by a permutation $y=(y_1,y_2,\dots,y_n)$ in the permutation space of $\{1, 2, \cdots, n\}$, i.e., $\mathcal{P}_n$. In this paper we focus on the symmetric TSPs, i.e., $c_{i,j} = c_{j,i}$ for all $i,j \in \{1,2,\dots,n\}$.

As mentioned above, the TSP belongs to the sum-of-the-parts COP. In the TSP, the edge set $\cal E$ can be seen as the finite unit set $U$ in Eq.~\ref{eq:COP_subclass} with the edge costs as the unit costs. A TSP solution $x$ is a subset of ${\cal E}$ and the function value of $x$ is the summation of the edge costs in $x$. Hence the proposed objective decomposition method can be directly applied to the TSP.

To decompose a TSP, for each edge $(i,j)$, first $c^{(1)}_{i,j}$ is randomly sampled from a pre-defined probability distribution $p$ in $(0, c_{i,j})$, where $c_{i,j}$ is the original edge cost, then $c^{(2)}_{i,j} = c_{i,j} - c^{(1)}_{i,j}$. It is obvious that $c^{(1)}_{i,j}>0$, $c^{(2)}_{i,j}>0$ and $c^{(1)}_{i,j}+c^{(2)}_{i,j}=c_{i,j}$ for any $i,j\in\{1,2,\dots,n\}$, hence $\left\{c^{(1)}_{i,j}\mid i,j\in \{1,2,\dots,n\}\right\}$ and $\left\{c^{(2)}_{i,j}\mid i,j\in \{1,2,\dots,n\}\right\}$ define two legal TSPs $f_1$ and $f_2$ and $f(x) = f_1(x) + f_2(x)$ for any $x$ in the solution space. Note here that, the generated TSPs $f_1$ and $f_2$ are both non-Euclidean TSP, i.e., the edge costs in $f_1$ and $f_2$ are not Euclidean distances in the 2D space.

\begin{figure}
  \centering
  \subfigure[``bell'']{
    \label{fig:p_fun_mountain} 
    \includegraphics[width=0.3\linewidth]{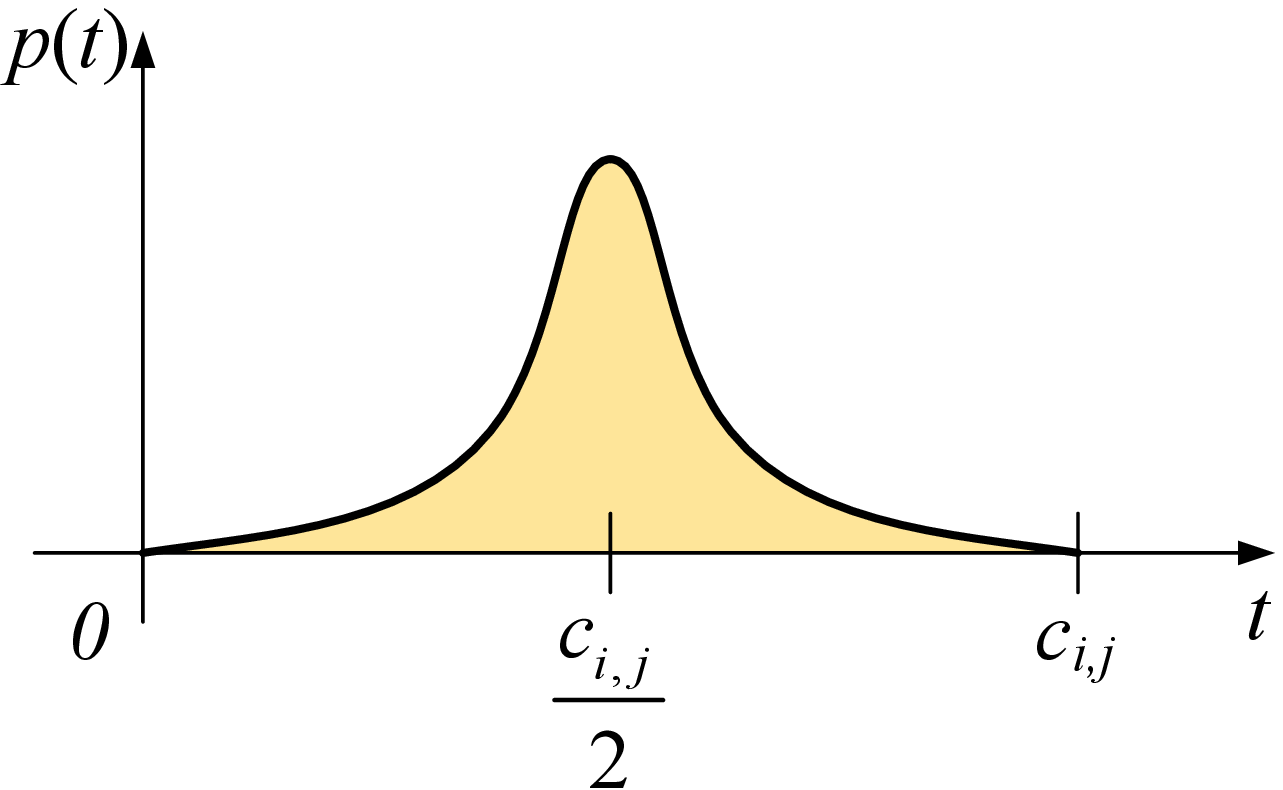}}
    \hspace{0.0\linewidth}
  \subfigure[``valley'']{
    \label{fig:p_fun_valley}
    \includegraphics[width=0.3\linewidth]{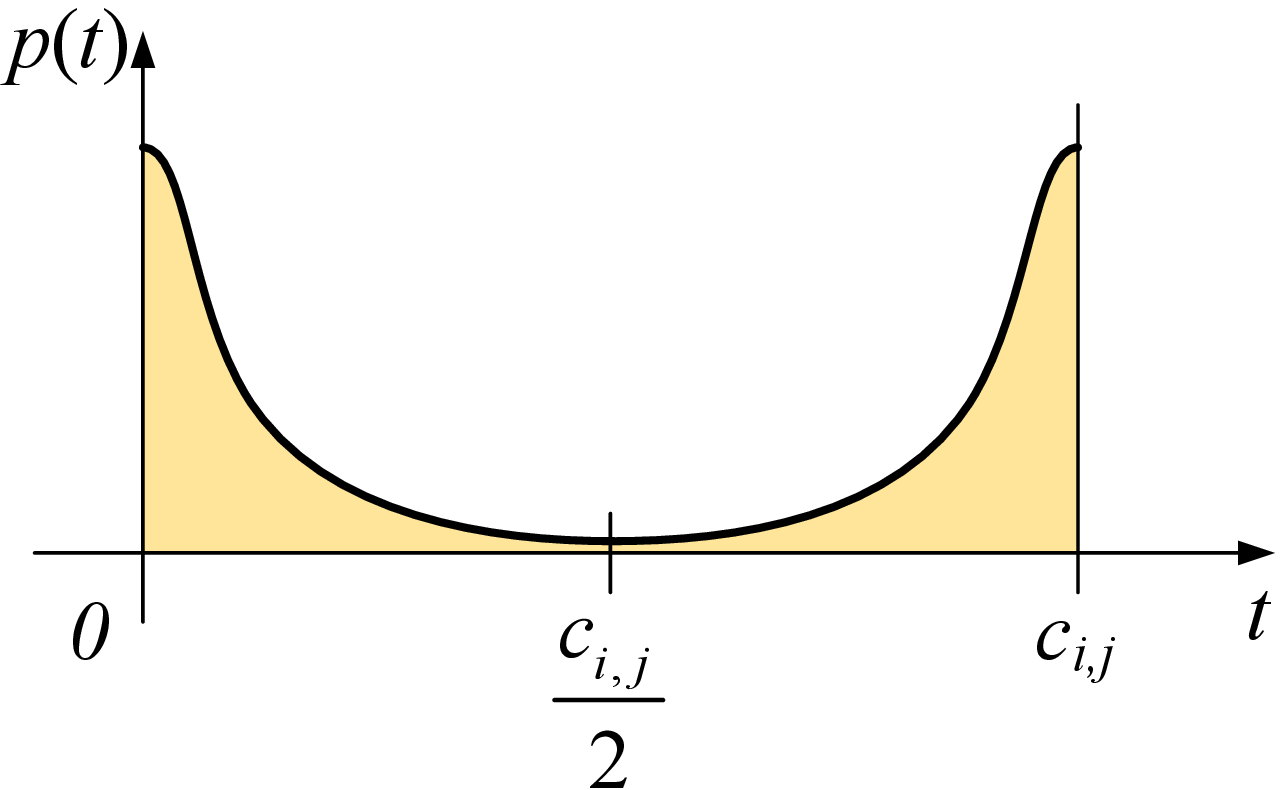}}
  \subfigure[``line'']{
    \label{fig:p_fun_plain}
    \includegraphics[width=0.3\linewidth]{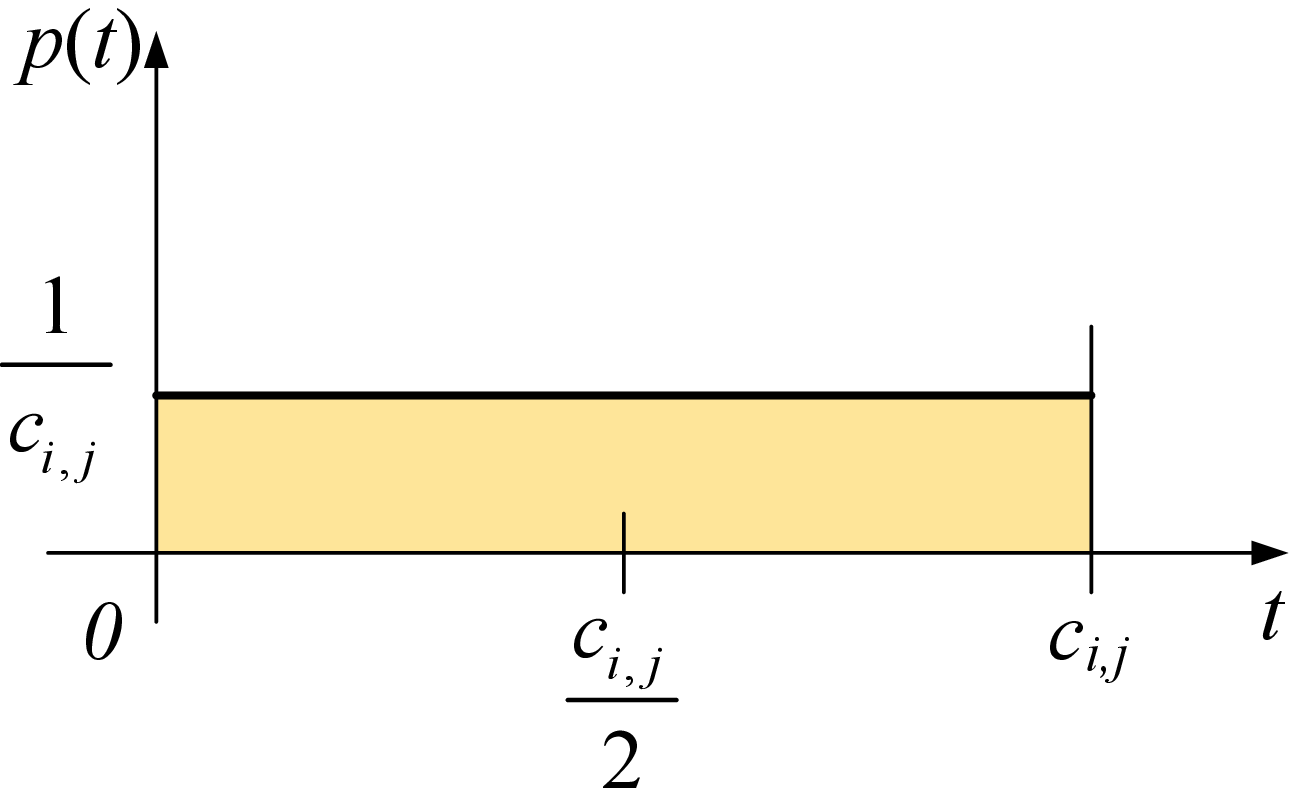}}
 \caption{Examples of probability distribution $p$ used to decompose each edge in the TSP.}\label{fig:shape_of_p}
\end{figure}

We find that the correlation between $f_1$ and $f_2$ by such a decomposition can be controlled by the shape of $p$. In Figure~\ref{fig:shape_of_p}, we show three examples of $p$ with different shapes, namely ``bell'', ``valley'' and ``line''.

When $p$ is of the shape of a ``bell'' (Figure~\ref{fig:p_fun_mountain}), the greatest probability is obtained in the middle of $(0, c_{ij})$. Sampling $c^{(1)}_{i,j}$ from the bell distribution is therefore of high probability to be $\approx c_{i,j}/2$. Since $c^{(2)}_{i,j} = c_{i,j}-c^{(1)}_{i,j} \approx c_{i,j}/2$, it means that the probability that $c^{(1)}_{i,j} \approx c^{(2)}_{i,j}$ is very high and the correlation coefficient $\rho$ will be roughly 1.

When $p$ is of the shape of a ``valley'' (Figure~\ref{fig:p_fun_valley}), the probability that $c^{(1)}_{i,j} \approx 0$ or $c^{(1)}_{i,j} \approx c_{i,j}$ is very high. Hence it is very likely that the difference between $c^{(1)}_{i,j}$ and $c^{(2)}_{i,j}$ is relatively large, which means $\rho$ is close to $-1$.

When $p$ is of the shape of a ``line'', it is actually the uniform distribution (Figure~\ref{fig:p_fun_plain}).  $c^{(1)}_{i,j}$ takes any value in $(0, c_{i,j})$ with equal probability, so does $c^{(2)}_{i,j}$. As a result, the correlation coefficient $\rho \approx 0$.

\begin{figure}
  \centering
  \subfigure[$\rho=-0.5657$]{
    \label{fig:corr_example_1}
    \includegraphics[width=0.23\linewidth]{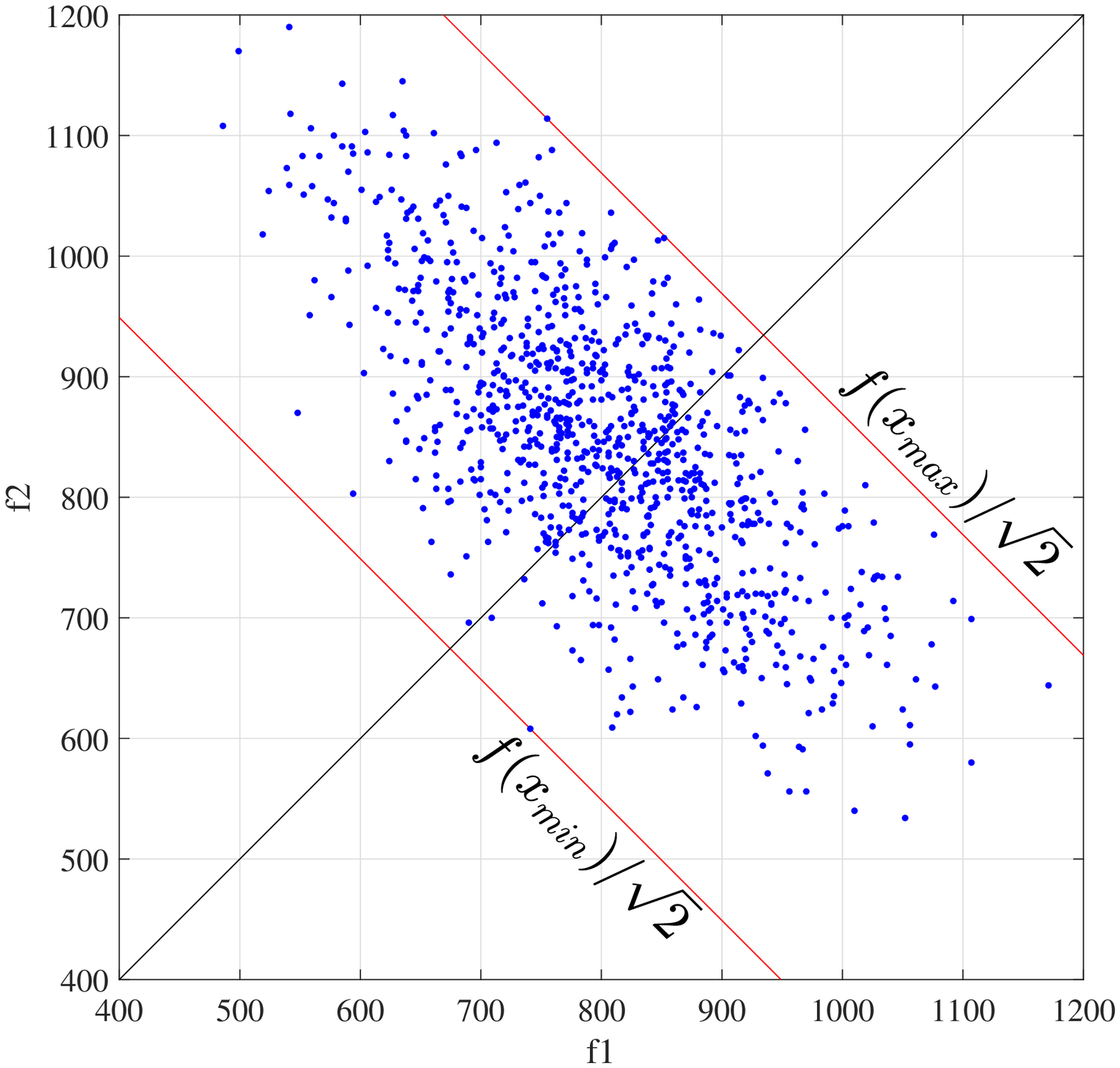}}
    \hspace{-0.005\linewidth}
  \subfigure[$\rho=-0.3586$]{
    \label{fig:corr_example_2}
    \includegraphics[width=0.23\linewidth]{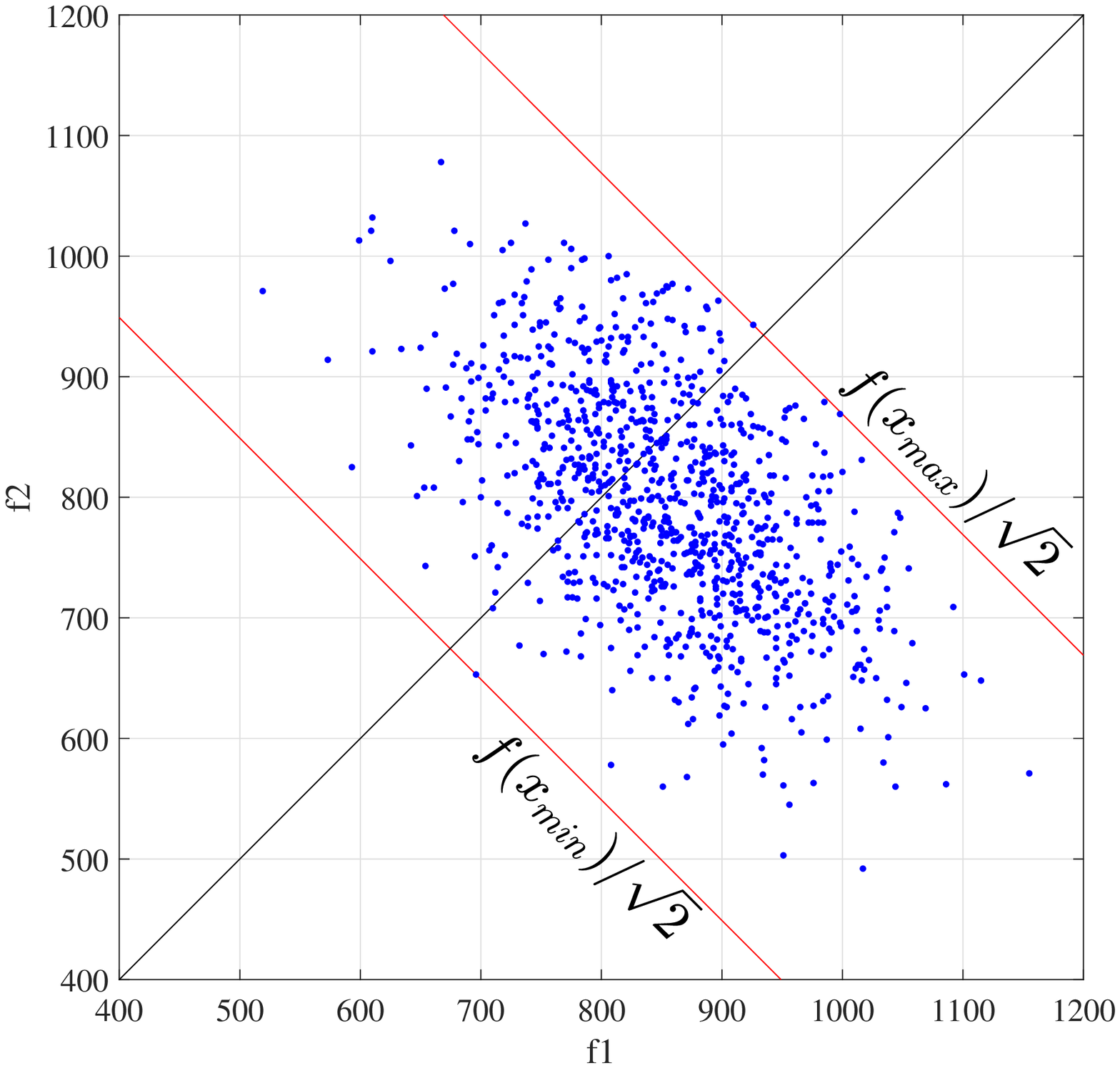}}
    \hspace{-0.005\linewidth}
  \subfigure[$\rho=-0.2271$]{
    \label{fig:corr_example_3}
    \includegraphics[width=0.23\linewidth]{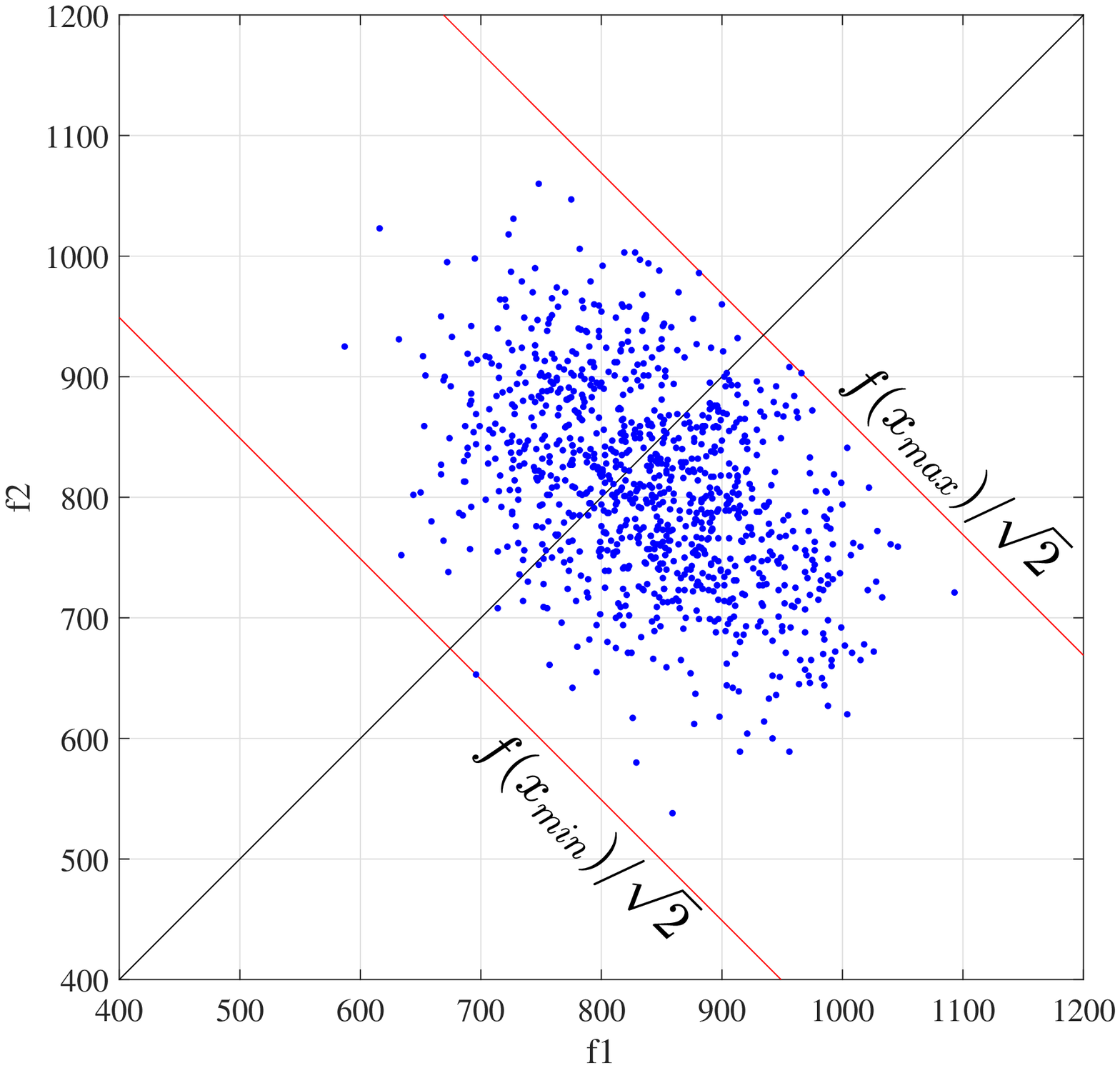}}
    \hspace{-0.005\linewidth}
  \subfigure[$\rho=-0.1087$]{
    \label{fig:corr_example_4}
    \includegraphics[width=0.23\linewidth]{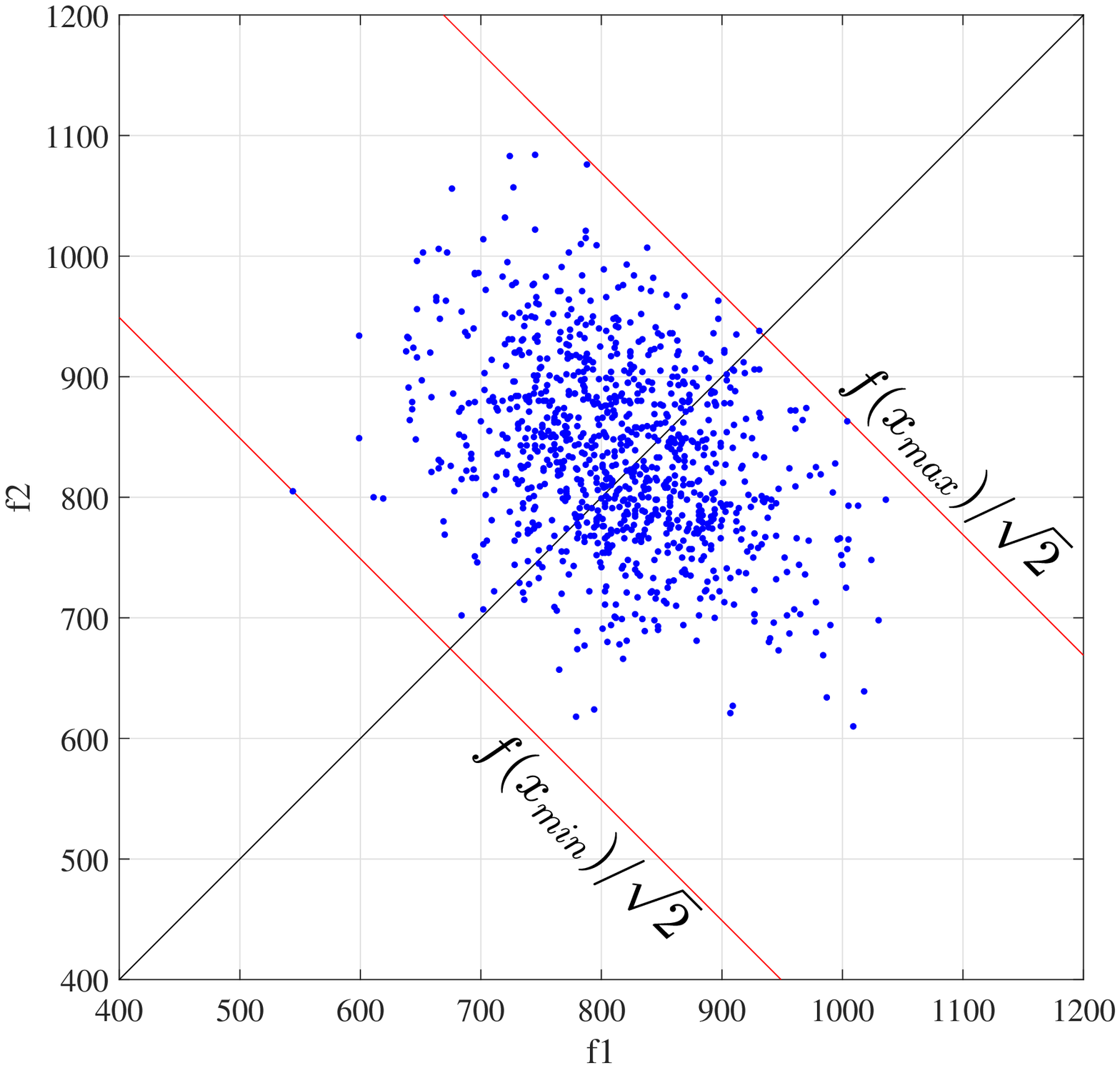}}\\
    \vspace{-0.005\linewidth}
  \subfigure[$\rho=0.3122$]{
    \label{fig:corr_example_5}
    \includegraphics[width=0.23\linewidth]{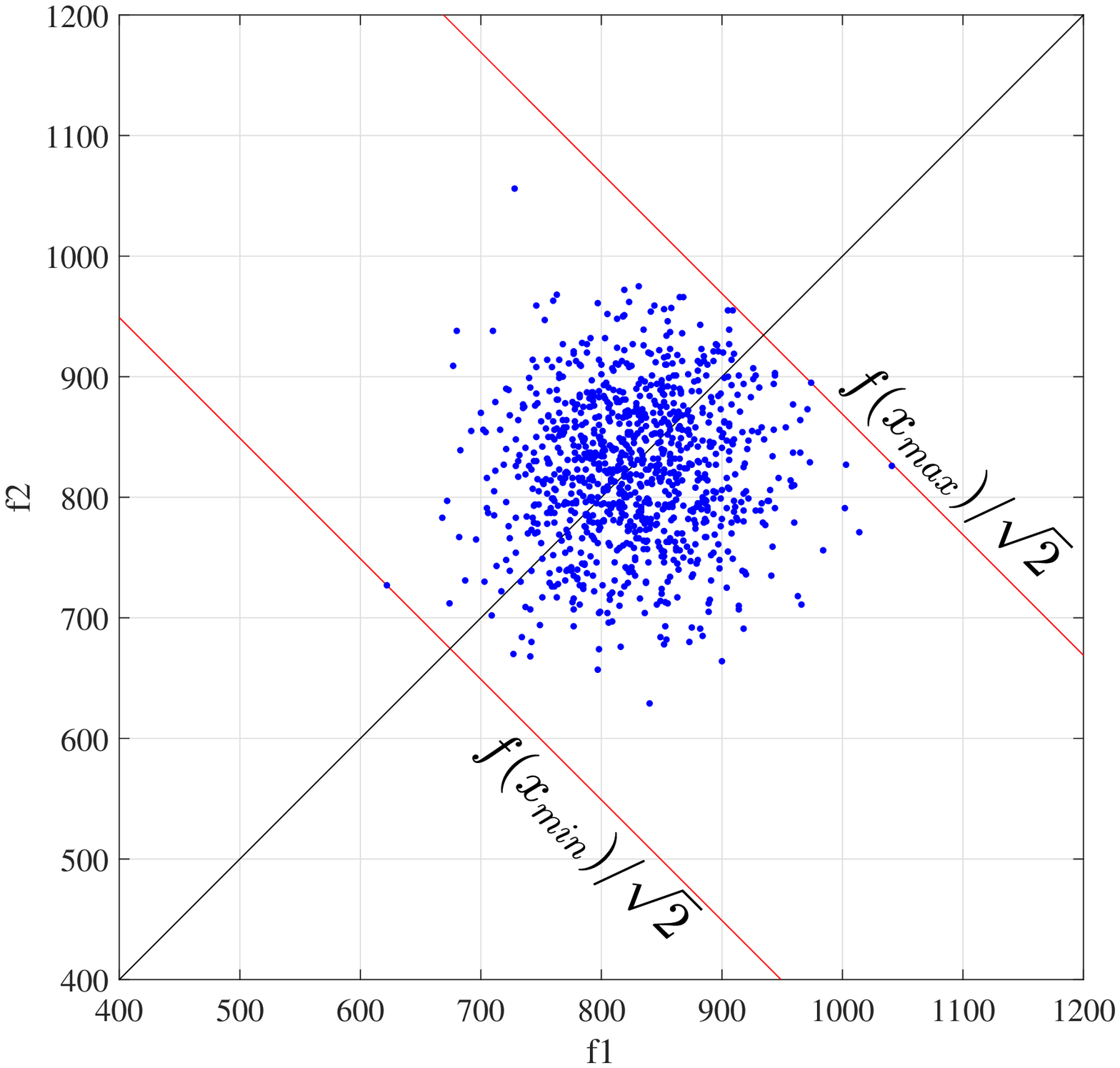}}
    \hspace{-0.005\linewidth}
  \subfigure[$\rho=0.4789$]{
    \label{fig:corr_example_6}
    \includegraphics[width=0.23\linewidth]{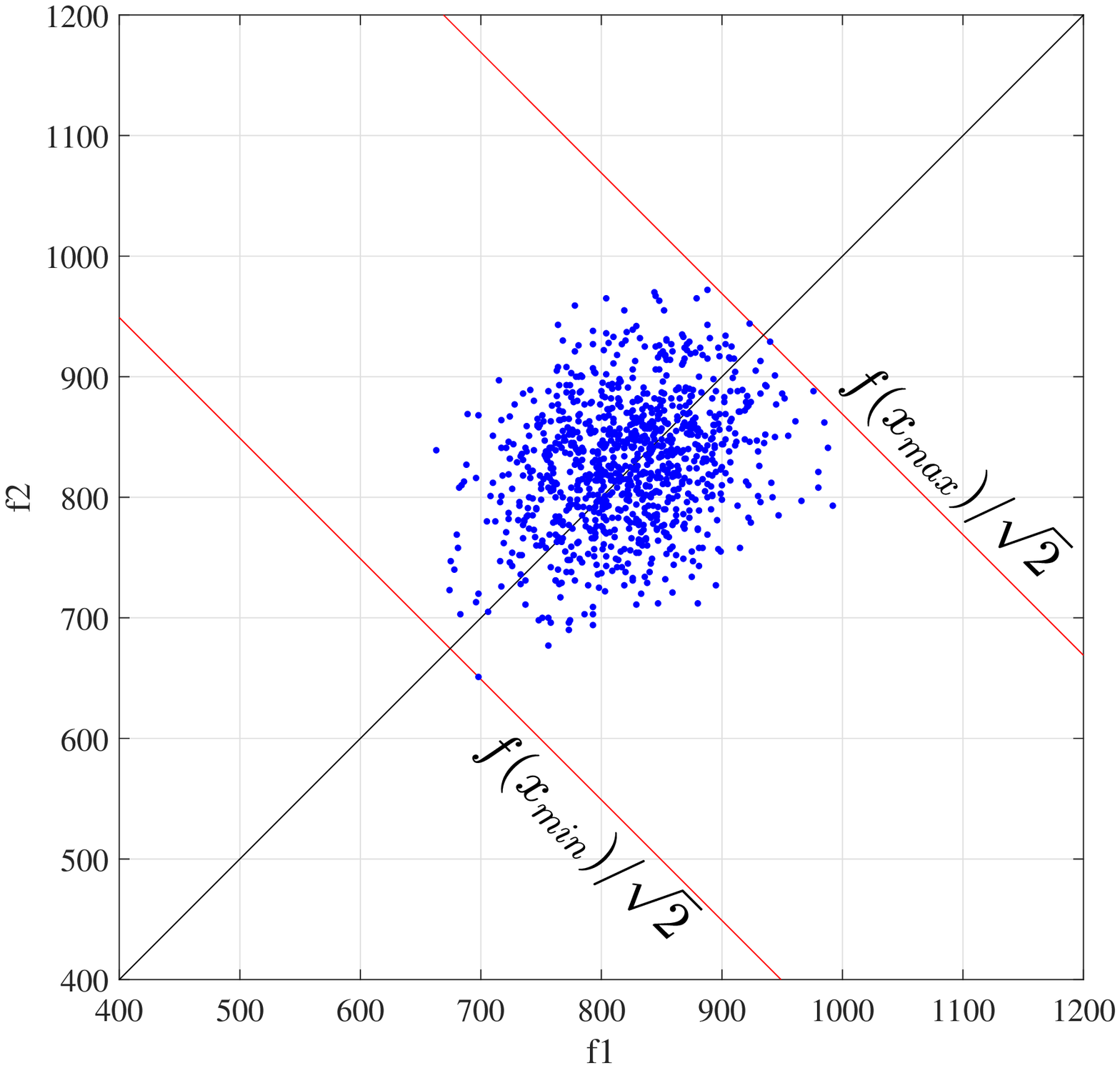}}
    \hspace{-0.005\linewidth}
  \subfigure[$\rho=0.7482$]{
    \label{fig:corr_example_7}
    \includegraphics[width=0.23\linewidth]{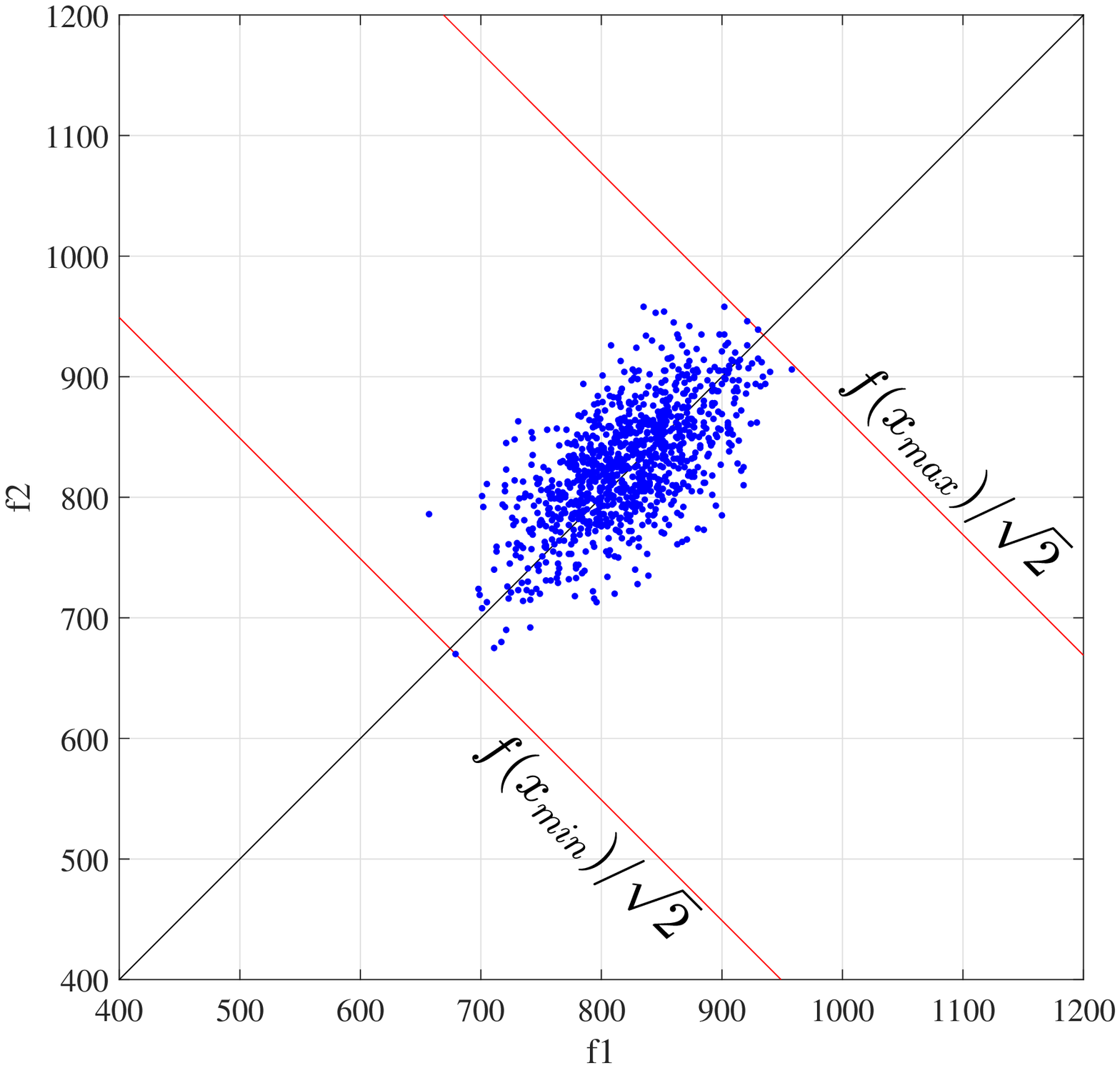}}
    \hspace{-0.005\linewidth}
  \subfigure[$\rho=0.9330$]{
    \label{fig:corr_example_8}
    \includegraphics[width=0.23\linewidth]{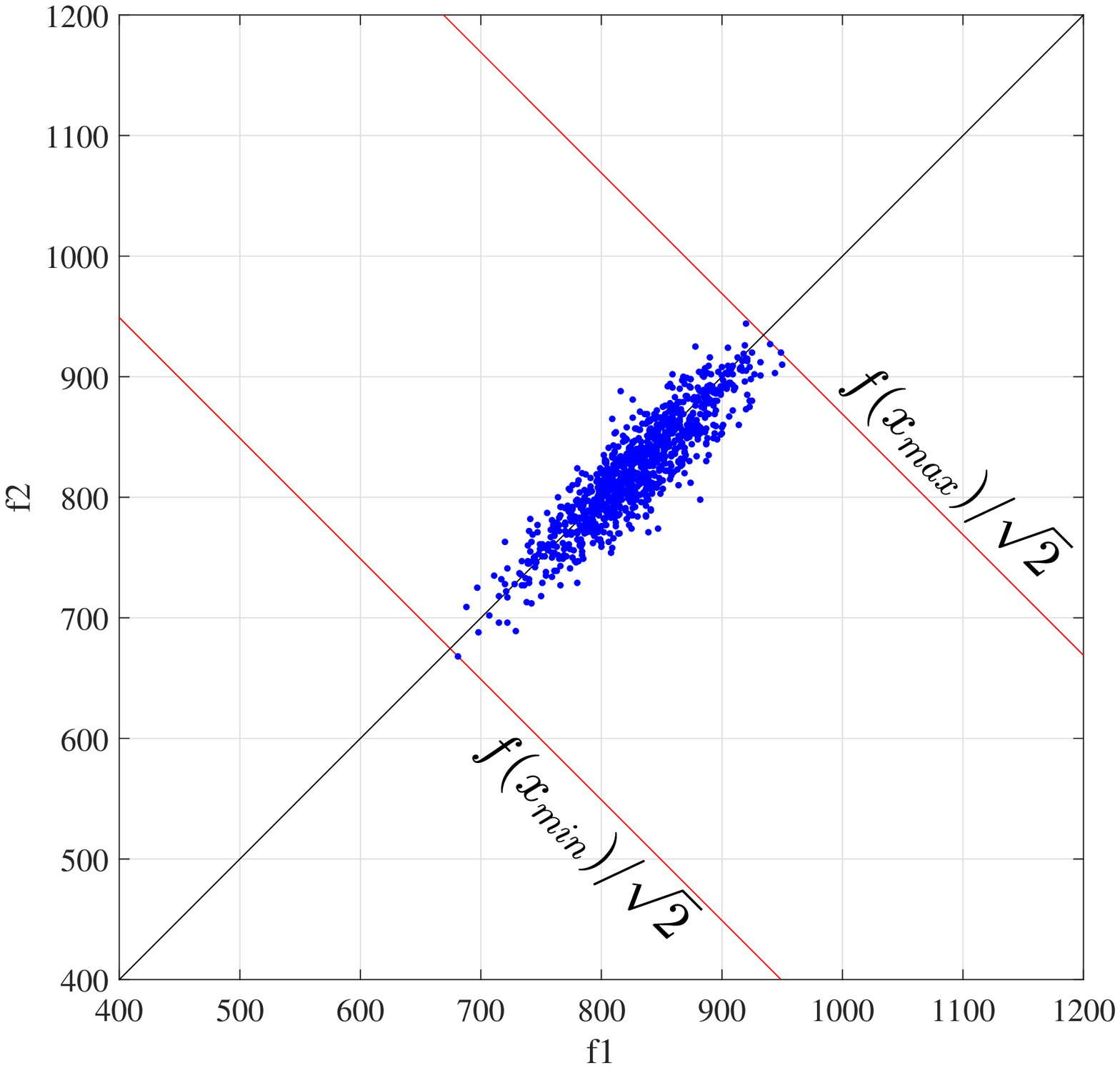}}\\
 \caption{Examples of the decomposition effect with different $\rho$ values on the TSP instance eil51, in which the same 1000 random solutions are plotted in the eight sub-figures.}\label{fig:corr_example}
\end{figure}

In summary, it is seen that with different $p$ distributions, positively correlated, negatively correlated or nearly independent sub-TSPs can be obtained after decomposition. To illustrate how the two sub-objectives behave w.r.t. $\rho$, we carried out the following experiment taking the TSP instance eil51 from the TSPLIB as an example. First the objective of eil51 is decomposed according to different $p$ distributions, and eight pairs of $(f_1, f_2)$ with different $\rho$ values evenly ranging from $-0.5657$ to $0.9330$ are selected. The details of the generation method of the eight pairs of $(f_1, f_2)$ can be found in Section~\ref{sec:nei_explore}. Then 1000 solutions of eil51 are randomly generated. The $(f_1, f_2)$ values of the 1000 solutions for each pair are shown in each subplot of Figure~\ref{fig:corr_example}.

The maximum and minimum objective value, denoted as $f(x_{\max})$ and $f(x_{\min})$, respectively, of the 1000 solutions are also shown in Figure~\ref{fig:corr_example} in red lines. From Figure~\ref{fig:corr_example}, it is seen that along the increasing of $\rho$, the solutions become more and more concentrated along the middle axis.

\subsection{Decomposition of the UBQP }\label{sec:UBQP_decomp}

The UBQP is defined as follows:
\begin{equation}\label{eq:UBQP}
\begin{split}
\mbox{maximize} ~ & ~  f(z) = z^T Q z = \sum_{i=1}^{n} \sum_{j=1}^{n} q_{i,j} z_i z_j\\
\mbox{subject to} ~ & ~  z = (z_1,\dots,z_n) \in \{0,1\}^n,
\end{split}
\end{equation}
where a solution $z = (z_1,\dots,z_n)$ is a vector of $n$ binary variables and $Q=[q_{i,j}]$ is a $n \times n$ matrix. The UBQP is $\cal NP$-hard and has been widely studied~\cite{kochenberger2014unconstrained}.

The UBQP belongs to the sum-of-the-parts COP. In the UBQP, $Q$ can be seen as the finite set $U$ with $q_{i,j}$ as the unit costs. A UBQP solution $z$ uniquely defines a subset $x=\{q_{k,l}\mid z_k=1\land z_l=1\}$ and the function value of $z$ is the summation of the members in the subset $x$. Then we can treat the UBQP function as a function of $x$ and the UBQP can be reformulated as
\begin{equation}\label{eq:UBQP2}
\begin{split}
\mbox{maximize} ~ & ~ \displaystyle f(x) = \sum_{q_{i,j}\in x}  q_{i,j},\\
\mbox{subject to} ~ & ~ x=\{q_{k,l}\mid z_k=1 \land z_l=1 \},\\
~ & ~  z = (z_1,\dots,z_n) \in \{0,1\}^n.
\end{split}
\end{equation}
Hence, the proposed objective decomposition method can be applied to the UBQP.

Different from the positiveness of the edge cost in the TSP, the element values in $Q$ is mixed by positive values, negative values and zeros. The decomposition method proposed for the TSP is thus not entirely applicable for the UBQP, but a similar idea can be used.

For each pair $(i,j)$, we propose to sample $q^{(1)}_{i,j}$ from a pre-defined probability distribution $p$ defined in the interval $(\frac{q_{i,j}}{2} - q', \frac{q_{i,j}}{2} + q')$ where $q'>0$ is a pre-defined positive constant. Similarly, we can control the correlation coefficient $\rho$ by choosing a bell-like, valley-like or uniform $p$. After the value of $q^{(1)}_{i,j}$ is generated, we let $q^{(2)}_{i,j} = q_{i,j} - q^{(1)}_{i,j}$. Eventually, we get two sub-UBQPs $f_1$ and $f_2$ based on $\{q^{(1)}_{i,j}\mid i,j\in \{1,2,\dots,n\}\}$ and $\{q^{(2)}_{i,j}\mid i,j\in \{1,2,\dots,n\}\}$ and we have $f(x) = f_1(x) + f_2(x)$ for any solution in the solution space. The correlation coefficient $\rho$ between $f_1$ and $f_2$ is
\begin{equation}\label{eq:corr_coef_ubqp}
 \rho = \frac{\mbox{cov}(\{q^{(1)}_{i,j}\},\{q^{(2)}_{i,j}\})}{\sigma(\{q^{(1)}_{i,j}\})\cdot \sigma(\{q^{(2)}_{i,j}\})}.
\end{equation}

\subsection{Decomposition of Other Similar Problems}
Decomposing other sum-of-the-parts COPs can refer to the decomposition methods of the TSP and the UBQP. The key is defining the unit set and unit cost properly. For example, for the vehicle routing problem, since it is similar to the TSP, all the possible edges in the graph of the vehicle routing problem can be seen as the finite unit set with the edge costs as the unit costs. For the quadratic assignment problem, either the cost per unit distance between facilities or the distance between sites can be seen as the unit cost. For the knapsack problem, naturally the item set can be seen as the unit set with the item values as the unit costs. Note here that, our decomposition method only decomposes the objective function. It does not decompose the constraint functions, hence it will not change the fact whether a solution meets the constraints or not. In the knapsack problem, the weights of items are not decomposed and the total weight of a knapsack problem solution will not change after the decomposition.

\section{The Proposed Multi-Objectivization Inspired Techniques}\label{sec:new_methods}

Based on the proposed objective decomposition method, we propose two new multi-objectivization inspired techniques, named as Non-Dominance Search (NDS) and Non-Dominance Exploitation (NDE), respectively. They can be used to improve the global search ability of local-search-based metaheuristics. Particularly, NDS is applicable for metaheuristics with fixed neighborhood structure, while NDE works with varied neighborhood structure.

\subsection{Non-Dominance Search (NDS)}\label{sec:NDS}

Given a neighborhood definition in the solution space, a local search process iteratively evaluates the neighborhood of the current solution and moves to a better neighboring solution. Local search usually stops at a solution that is not worse than its neighbors but not necessarily than all other solutions in the solution space, i.e.,~a local optimum. To escape from the local optimum, a possible strategy is to enlarge the neighborhood size. For example, in Variable Neighborhood Search (VNS), once the search is trapped in a local optimum, the neighborhood size is enlarged until a better solution in the enlarged neighborhood is found.

However, enlarging the neighborhood size can result in high computational complexity if all the solutions in the enlarged neighborhood are all to be evaluated. The proposed NDS can reduce the computational complexity by only selecting the neighboring solutions of the local optimum that are \emph{non-dominated} to the current local optimum with regard to $(f_1, f_2)$ and only evaluating the neighborhood of the selected neighboring solutions.

Below we first give the definition of dominance and non-dominance in the multi-objective minimization case and then present the NDS procedure.
\begin{defn}\emph{Dominance}: A vector $u=(u_1,\dots,u_m)$ is said to \emph{dominate} a vector $v=(v_1,\dots,v_m)$, if and only if $u_k\leq v_k,\ \forall k\in\{1,\dots,m\}\ \land\ \exists k\in\{1,\dots,m\}: u_k<v_k$, denoted as $u \prec v$\end{defn}

\begin{defn} \emph{Non-dominance}: If $u$ is not dominated by $v$ and $v$ is not dominated by $u$, we say that $u$ and $v$ are \emph{non-dominated} to each other, denoted as $u \nprec v$ or $v \nprec u$.
\end{defn}

\begin{figure}
  \centering
  \includegraphics[width=0.5\linewidth]{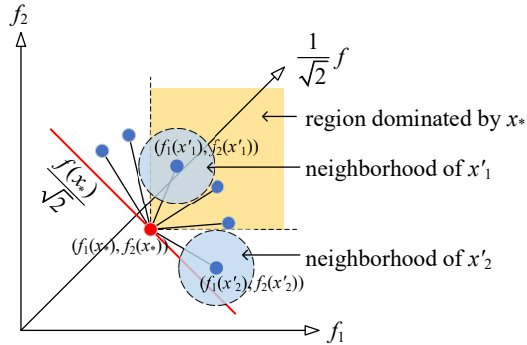}\\
  \caption{Assume the local optimum $x_*$ has six neighboring solutions. The neighborhood of $x'_2$ are more likely to break through the contour of ${f(x_*)}/{\sqrt{2}}$ than the neighborhood of $x'_1$ since $x'_2$ is non-dominated to $x_*$.}\label{fig:idea}
\end{figure}

The idea behind NDS is presented in Figure~\ref{fig:idea} in which we assume that a local optimum $x_*$ has six neighboring solutions for a minimization problem with the objective function $f(x)$. All the neighboring solutions are located above the ${f(x_*)}/{\sqrt{2}}$ contour (red line in Figure~\ref{fig:idea}) since $x_*$ is a local optimum. NDS intends to find a neighboring solution whose neighborhood can break through the contour of ${f(x_*)}/{\sqrt{2}}$. From Figure~\ref{fig:idea} we can see that the neighboring solutions that are not dominated by $x_*$ (e.g. $x'_2$) are more likely to be close to the contour of ${f(x_*)}/{\sqrt{2}}$, compared to the solutions that are dominated by $x_*$ (e.g. $x'_1$). Hence the neighborhood of $x'_2$ is more likely to contain a solution that can break through the ${f(x_*)}/{\sqrt{2}}$ contour than the neighborhood of $x'_1$. 

The detailed procedure of NDS in a minimization case is shown in Algorithm~\ref{alg:NDS}, in which the first-improvement strategy is used. The input of NDS is a local optimum $x_*$. If a solution $x'$ in the neighborhood of $x_*$ is non-dominated w.r.t. the decomposed sub-objectives (line~\ref{nds1}), the neighborhood of $x'$ is to be explored. Once a better solution is found (line~\ref{nds3}), NDS will immediately terminate and return the better solution (line~\ref{nds4}). If no better solution can be found, NDS will return the original local optimum $x_*$.

\begin{algorithm}
\small
\SetKwInput{KwInput}{Input}
\KwInput{$x_*$, $f$, $f_1$, $f_2$}
    $x_{output} \gets x_*$\;
    \For {each $x'\in$Neighborhood($x_*$)}{
        \If {$(f_1(x_*),f_2(x_*))\nprec (f_1(x'),f_2(x'))$}{ \label{nds1}
            \For {each $x''\in$Neighborhood($x'$)}{ \label{nds2}
                \If {$f(x'')<f(x_*)$}{ \label{nds3}
                    $x_{output} \gets x''$\;  \label{nds4}
                    exit;
                }
            }
        }
    }
    \KwRet{$x_{output}$}\label{lin:output}
\caption{Non-Dominance Search (NDS)}
\label{alg:NDS}
\end{algorithm}

NDS cannot be used as a standalone COP solver. Rather, it can be embedded within a metaheuristic with fixed neighborhood structure. As a case study, we combine NDS with the basic Iterated Local Search (ILS) procedure and the resultant algorithm is called ILS+NDS. The procedures of the original ILS and the proposed ILS+NDS are shown in Algorithm~\ref{alg:ILS} and Algorithm~\ref{alg:ILS+NDS}, respectively. The key difference of ILS+NDS to ILS is that the perturbation process in ILS (line~\ref{ils1} in Algorithm~\ref{alg:ILS}) is replaced by the NDS process in ILS+NDS (line~\ref{ilsnds0} in Algorithm~\ref{alg:ILS+NDS}) to obtain a better restarting point. If the NDS procedure fails, ILS+NDS will conduct a perturbation process to escape from the current local optimum (line~\ref{ilsnds1} in Algorithm~\ref{alg:ILS+NDS}).

\begin{algorithm}
\small
    Decompose $f$ into $f_1$ and $f_2$\; 
    $x_0' \gets $ random or heuristically generated solution\;
    set $x_{best} \gets x_0'$ and $j \gets 0$\;
    \While{stopping criterion is not met}{
        $x_j \gets$ LocalSearch($x_j'$)\;
        \If {$f(x_{j}) < f(x_{best})$} {
            $x_{best} \gets x_{j}$\;
        }
        $x_{j+1}' \gets$ NDS($x_j\mid f,f_1,f_2$)\;\label{ilsnds0}
        \If {$x_{j+1}'=x_j$}{\label{lin:NDS_fail}
            $x_{j+1}' \gets$ Perturbation($x_j$)\; \label{ilsnds1}
        }
        $j\gets j+1$\;
    }
    \KwRet{\mbox{the historical best solution} $x_{best}$}
\caption{ILS+NDS}
\label{alg:ILS+NDS}
\end{algorithm}

We do not claim that ILS+NDS is competitive to the state-of-the-art metaheuristics for COPs. The aim of designing ILS+NDS is to show that the proposed multi-objectivization inspired method is beneficial to metaheuristics like ILS. 

\subsection{Non-Dominance Exploitation (NDE)}\label{sec:NDE}

In the previous subsection, we proposed the NDS technique to enhance local-search-based metaheuristics. To apply NDS, the neighborhood structure in the local search method should be fixed during the search.

However, in some metaheuristics, the neighborhood structure is varying during the search. For example, in the LK local search for the TSP, a fine-grained edge exchange strategy is used at each move. The number of exchanged edges is not fixed among moves. The neighborhood structure is thus varied during the LK search.

\begin{figure}
  \centering
  \includegraphics[width=0.3\linewidth]{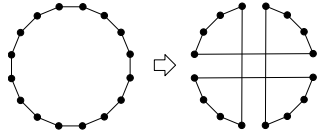}\\
  \caption{An example of the double bridge perturbation on the TSP}\label{fig:double_bridge}
\end{figure}

Though NDS is not able to be embedded within the LK, this does not mean that the proposed multi-objectivization inspired method cannot benefit the LK local search. In this section, we propose to embed the decomposition method within the Iterated Lin-Kernighan local search (ILK)~\cite{johnson1997traveling}. The proposed algorithm is called ILK with Non-Dominated Exploitation (ILK+NDE). ILK is a variant of ILS, in which a LK local search and a double bridge perturbation (please see Figure~\ref{fig:double_bridge} for a demo) are iteratively executed.

Different to NDS which finds promising neighboring solutions in the neighborhood of local optima, ILK+NDE explores promising LK local optima based on the non-dominance relationship of $(f_1, f_2)$. The detailed procedure of ILK+NDE is shown in Algorithm~\ref{alg:ILK+NDE}. In Algorithm~\ref{alg:ILK+NDE}, the original objective $f$ is first decomposed into two sub-objectives (line~\ref{nde1}). A current best solution $x_{best}$ is found by applying the LK search (lines~\ref{nde2} to~\ref{nde2.1}). At each iteration $j$, if the current solution $x_j$ is non-dominated to $x_{best}$ with regard to $(f_1, f_2)$, the region close to $x_*$ in the search space will be further exploited (line~\ref{nde3}) and $x_{j+1}$ is returned. The exploitation procedure is summarized in Algorithm~\ref{alg:futher_exploit}. In ILK+NDE, if the exploitation procedure is failed (i.e., $x_{j+1} = x_j$), a perturbation method is applied (lines~\ref{4} to~\ref{4.2}). The algorithm terminates when the stop criterion is met.

In the exploitation procedure (Algorithm~\ref{alg:futher_exploit}), at each round of the exploitation, first $k$ edges are randomly selected from $x_*$ and a penalty cost $\tilde{c}$ will be added to the selected edges (line~\ref{exploit1} in Algorithm~\ref{alg:futher_exploit}) by AddRandomPenalty($x_*$,$f$,$k$,$\tilde{c}$) (Algorithm~\ref{alg:add_rand_pen}). This will result in a new TSP instance with the objective $f'$. An LK local search is started from $x_*$ on $f'$ and returns $x'$ (line~\ref{exploit2} in Algorithm~\ref{alg:futher_exploit}). A new LK local search then applies from $x'$ on the original objective $f$ and returns $x''$ (line~\ref{exploit3}  in Algorithm~\ref{alg:futher_exploit}). If $f(x'')<f(x_*)$, then the exploitation procedure will immediately stop and output $x''$. Otherwise, a new round of random penalization will be executed on $x_*$ and $f$. If after $T$ rounds of penalization the procedure still cannot find a better $x''$ than $x_*$, Algorithm~\ref{alg:futher_exploit} terminates and returns $x_*$.


\begin{algorithm}
\small
\SetKwInput{KwInput}{Input}
\KwInput{$f$, $T$, $k$, $\tilde{c}$}
    Decompose $f$ into $f_1$ and $f_2$ such that $f(x) = f_1(x)+f_2(x)$\; \label{nde1}
    $x_0' \gets $ random or heuristically generated solution\; \label{nde2}
    $x_0 \gets$ LK($x_0'\mid f$)\; \label{nde2.1}
    $x_{best} \gets x_0$\;
    $j \gets 0$\;
    \While{stopping criterion is not met}{
        \If {$(f_1(x_{best}),f_2(x_{best}))\nprec(f_1(x_j),f_2(x_j))$}{
            $x_{j+1} \gets$ FurtherExploit($x_j\mid T,k,\tilde{c}$)\; \label{nde3}
            \If {$x_{j+1} = x_j$}{ \label{4}
                $x_{j+1}' \gets$ Perturbation($x_j$)\;  \label{4.1}
                $x_{j+1} \gets$ LK($x_{j+1}'\mid f$)\; \label{4.2}
            }
        }
        \Else {
            $x_{j+1}' \gets$ Perturbation($x_j$)\; \label{nde5}
            $x_{j+1} \gets$ LK($x_{j+1}'\mid f$)\; \label{nde6}
        }
        \If {$f(x_{j+1}) < f(x_{best})$} { \label{nde7}
            $x_{best} \gets x_{j+1}$\;  \label{nde8}
        }
        $j\gets j+1$\;
    }
    \KwRet{\mbox{the historical best solution} $x_{best}$}
\caption{ILK+NDE}
\label{alg:ILK+NDE}
\end{algorithm}




\begin{algorithm}
\small
    $x_{best} \gets x_*$\;
    $j \gets 1$\;
    \While {$j \leq T$}{
        $f' \gets$ AddRandomPenalty($x_*$,$f$,$k$,$\tilde{c}$)\; \label{exploit1}
        $x' \gets$ LK($x_*\mid f'$)\;\label{exploit2}
        $x'' \gets$ LK($x'\mid f$)\;\label{exploit3}
        \If {$f(x'')<f(x_{best})$}{
            $x_{best} \gets x''$\;
           exit\;
        }
        $j \gets j+1$\;
    }
    \KwRet{$x_{best}$}\label{lin:output_ENS_LK}
\caption{FurtherExploit($x_*\mid T,k,\tilde{c}$)}
\label{alg:futher_exploit}
\end{algorithm}

\begin{algorithm}
\small
    $\tilde{\cal E} \gets$ randomly select $k$ edges from $x_*$\;
    \For {each edge $(i,j)$ in the TSP $f$}{
        \If {edge $(i,j) \in \tilde{\cal E}$}{
            $c'_{i,j} \gets c_{i,j} + \tilde{c}$
        }
        \Else{
            $c'_{i,j} \gets c_{i,j}$
        }
    }
    \KwRet{$f'$: the TSP based on $\{c'\}$}
\caption{AddRandomPenalty($x_*,f,k,\tilde{c}$)}
\label{alg:add_rand_pen}
\end{algorithm}

\subsection{Discussions}

The similarity between NDS and NDE is that they both use the non-dominance relationship introduced by the decomposed sub-objectives $(f_1, f_2)$ to judge whether a solution is ``promising'' (i.e., worth further exploitation). In NDS, the neighboring solutions of a local optimum are checked, while in NDE, the local optima encountered during the search are judged based on $(f_1, f_2)$. One may argue that a reasonable and easy way to judge the potential of the neighboring solutions is to set a threshold and exclude solutions that are with objective function values worse than the threshold. However, since the objective functions of different problem instances have different value ranges, it is not easy to find a general method to properly set the threshold for different problem instances. Using the sub-objectives $(f_1, f_2)$ as the judging criterion is relatively less subjective since the decomposition is conducted in a stochastic way. In addition, the proposed objective decomposition method can be easily applied to different problem instances.

\section{Experimental Studies and Results}\label{sec:epm}

In this section, we first investigate the neighborhood non-dominance hypothesis, then conduct systematic experiments to test the performance of ILS+NDS and ILK+NDE. In the end, we combine the proposed NDS with Iterated Tabu Search (ITS)~\cite{palubeckis2006iterated} and test the performance of the resulting algorithm. All the compared algorithms are implemented by the authors using C++.  

\subsection{The Neighborhood Non-dominance Hypothesis}\label{sec:nei_explore}

\begin{figure}
  \centering
  \includegraphics[width=0.3\linewidth]{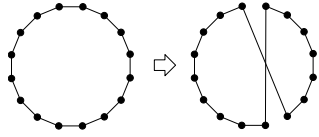}\\
  \caption{An example of the 2-Opt neighborhood in the TSP}\label{fig:2opt_neighbor}
\end{figure}

In the first experiment, we verify the hypothesis of NDS, i.e., the neighborhood of the non-dominated neighbors of a local optimum is more likely to contain a better solution. We select five TSP instances from the TSPLIB~\cite{reinelt1991tsplib} and five UBQP instances form the OR-Library~\cite{beasley1990or-library}. For the TSP instances, we use the 2-Opt neighborhood structure, i.e., a neighboring solution is obtained by replacing two edges of the current solution by another two edges, as illustrated in Figure~\ref{fig:2opt_neighbor}. For an $n$-city TSP, the size of the 2-Opt neighborhood is $n(n-3)/2$~\cite{michiels2007theoretical}. For the UBQP instances, we use the 1-bit-flip neighborhood structure, i.e., a neighboring solution is obtained by flipping a bit of the current solution. For an $n$-bit UBQP, the size of the 1-bit-flip neighborhood is $n$. Features of the selected TSP and UBQP instances are shown in Table~\ref{tbl:inst}.

\begin{table}
\caption{The selected TSP and UBQP instances} 
\centering
\label{tbl:inst}
\resizebox{\linewidth}{!}{
\begin{tabular}{l | c c c c c}
  \hline
  TSP instance & eil51 & st70 & pr76 & rat99 & rd100 \\
  \hline
  Problem size & 51 & 70 & 76 & 99 & 100 \\
  \hline
  2-Opt neighborhood size & 1224 & 2345 & 2774 & 4752 & 4850 \\
  \hline
  \hline
  UBQP instance& bqp1000.1 & bqp2500.1 & p3000.1 & p4000.1 & p5000.1 \\
  \hline
  Problem size & 1000 & 2500 & 3000 & 4000 & 5000 \\
  \hline
  1-bit-flip neighborhood size & 1000 & 2500 & 3000 & 4000 & 5000 \\
  \hline
\end{tabular}
}
\end{table}

\begin{figure}
  \centering
  \subfigure[$a=-10$]{
    \label{fig:real_p_fun_a-10} 
    \includegraphics[width=0.3\linewidth]{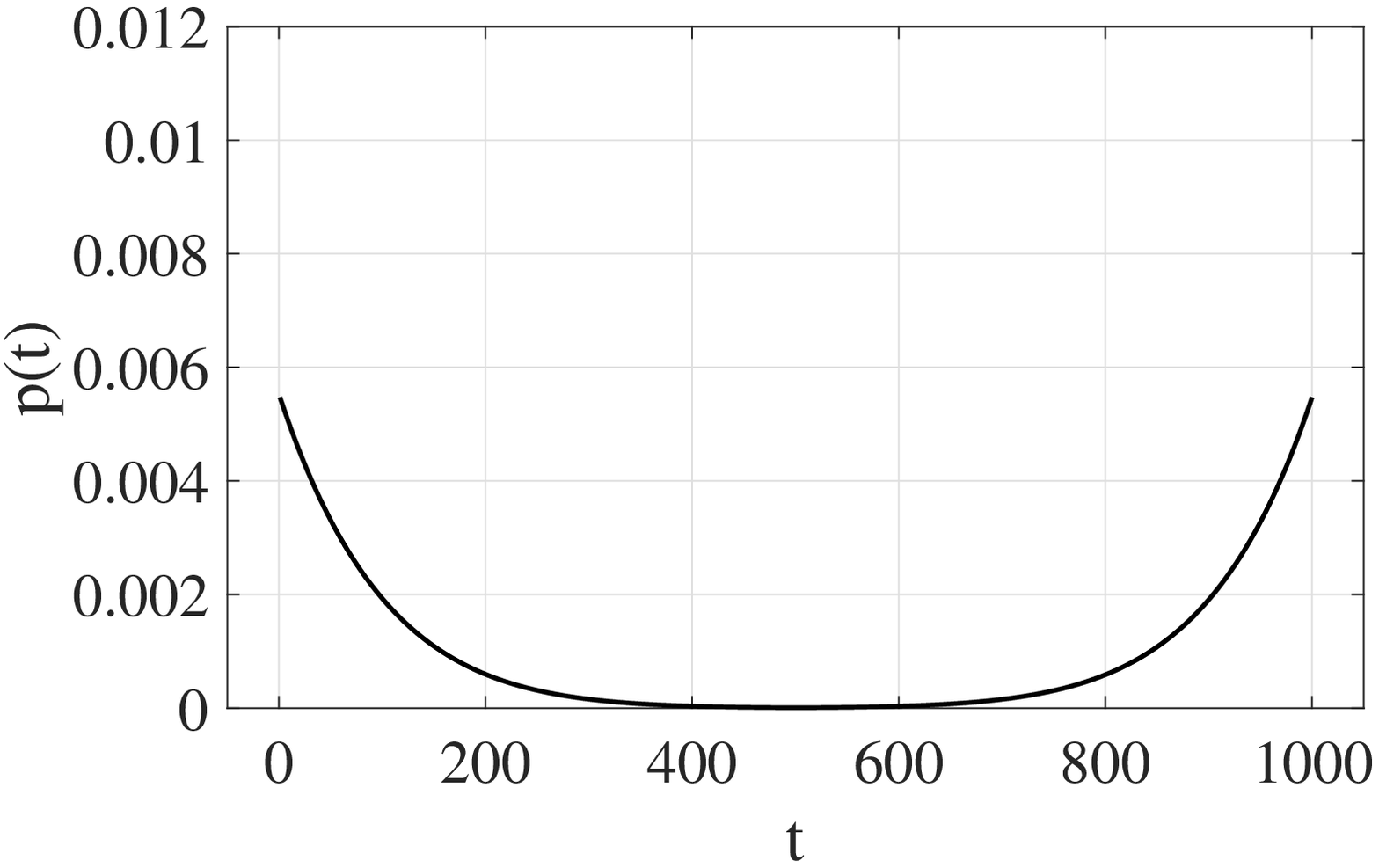}}
    \hspace{0.0\linewidth}
  \subfigure[$a=-5$]{
    \label{fig:real_p_fun_a-5} 
    \includegraphics[width=0.3\linewidth]{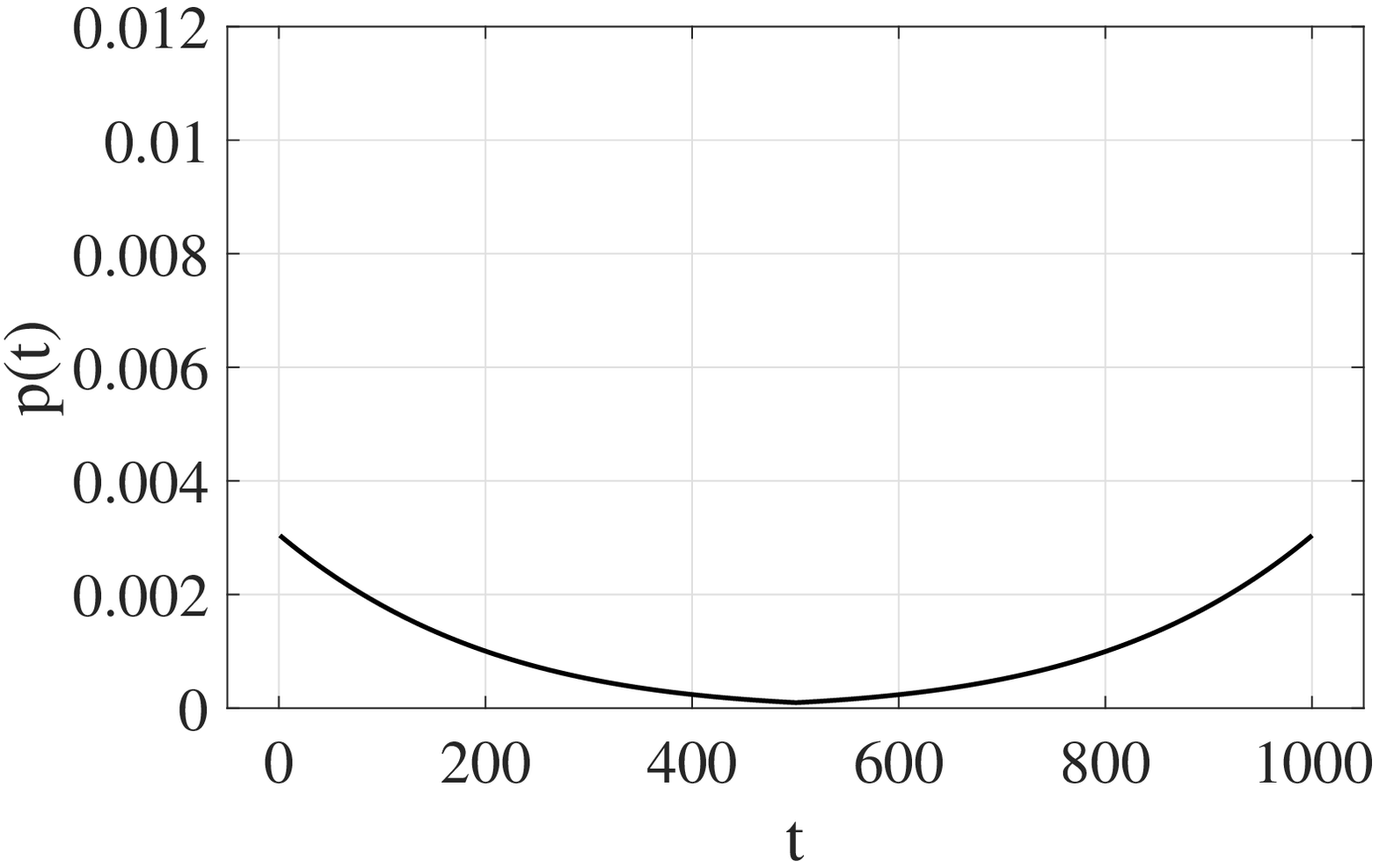}}
    \hspace{0.0\linewidth}
  \subfigure[$a=0$]{
    \label{fig:real_p_fun_a0}
    \includegraphics[width=0.3\linewidth]{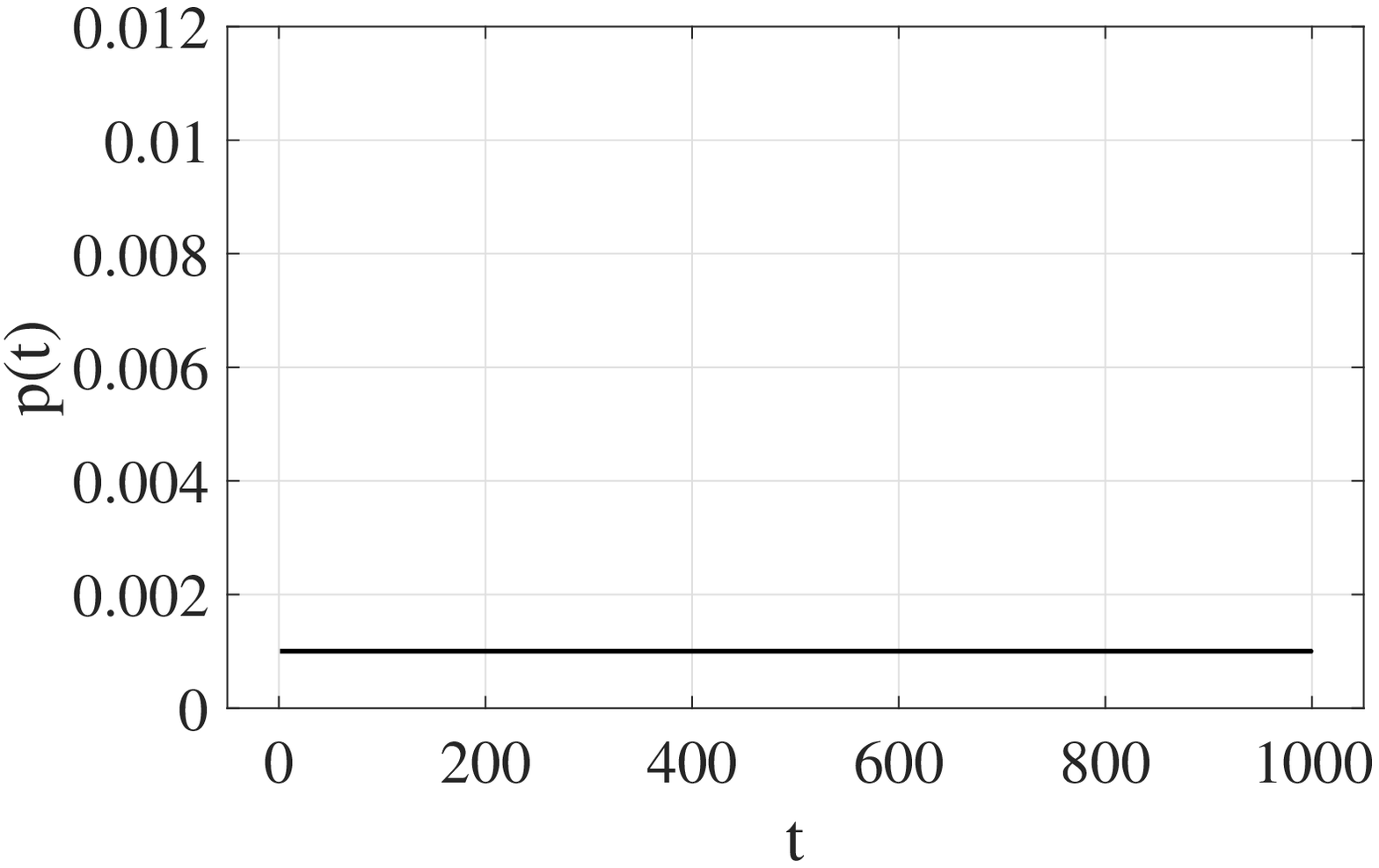}}
    \hspace{0.0\linewidth}
  \subfigure[$a=5$]{
    \label{fig:real_p_fun_a5} 
    \includegraphics[width=0.3\linewidth]{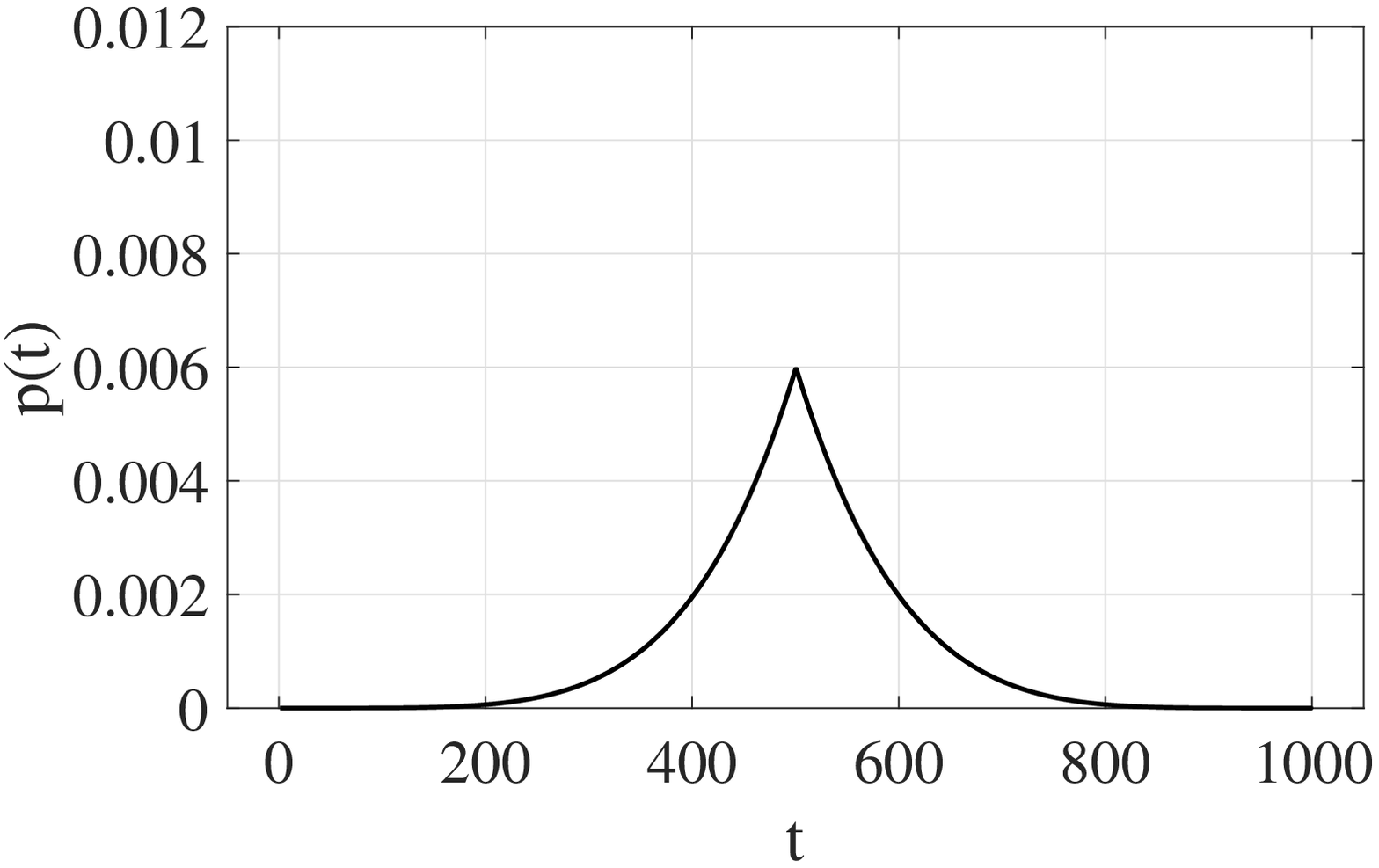}}
    \hspace{0.0\linewidth}
  \subfigure[$a=10$]{
    \label{fig:real_p_fun_a10}
    \includegraphics[width=0.3\linewidth]{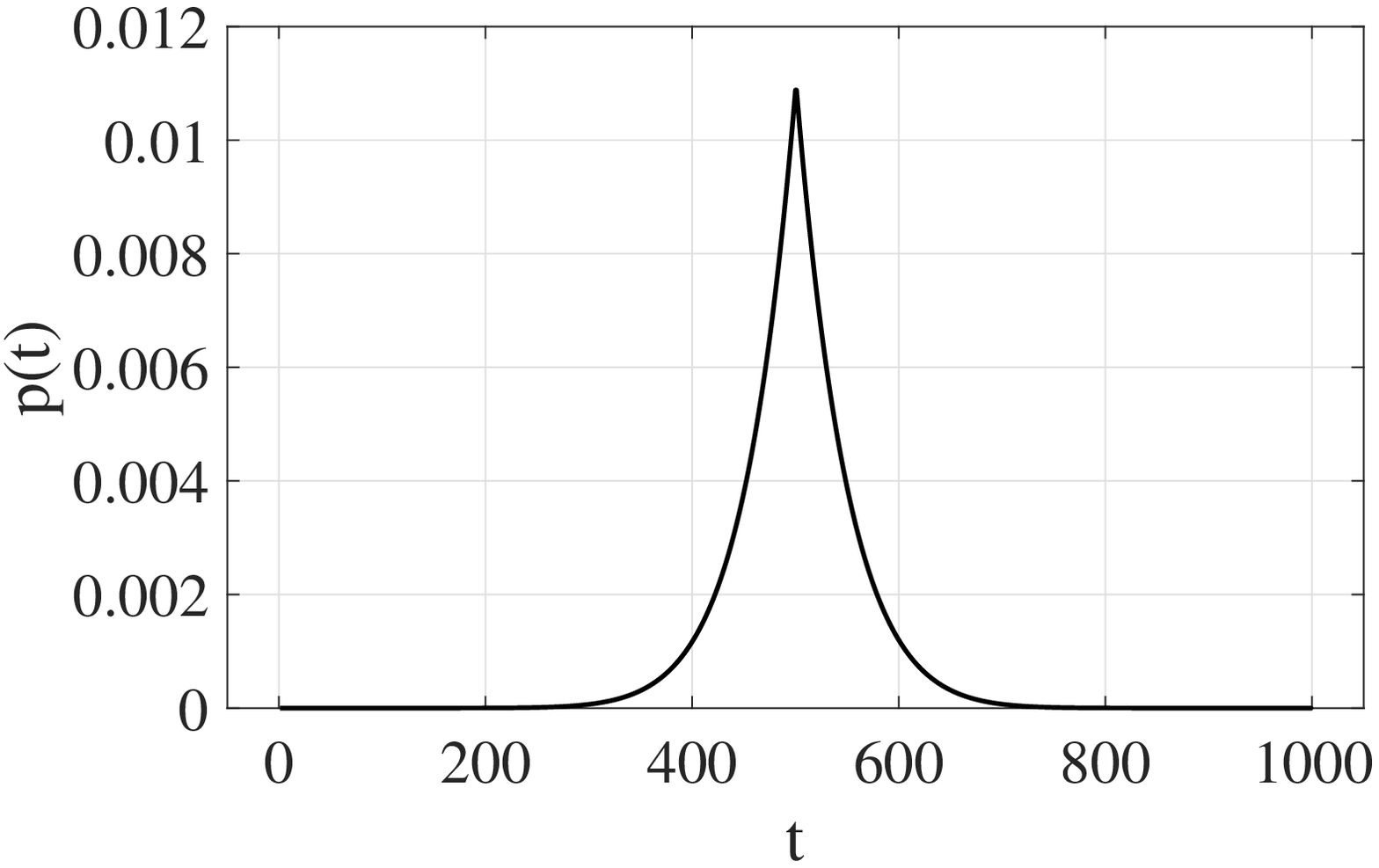}}
 \caption{Examples of probability distribution $p(t)$ when $c_{i,j}=1000$.}\label{fig:real_p} 
\end{figure}

To decompose the objective functions of the TSP instances, we first define a function $p'(t)$ in the interval $(0,c_{i,j})$ for each edge $(i,j)$:
\begin{equation}\label{eq:p_prime}
  p'(t)=\left\{
  \begin{array}{rl}
    (\frac{c_{i,j}}{2} - \mbox{sign}(a)(\frac{c_{i,j}}{2}-t))^{|a|} & \ \mbox{if} \ 0<t\leq\frac{c_{i,j}}{2},\\
    (\frac{c_{i,j}}{2} + \mbox{sign}(a)(\frac{c_{i,j}}{2}-t))^{|a|} & \ \mbox{if} \ \frac{c_{i,j}}{2}<t<c_{i,j},\\
  \end{array}
  \right.
\end{equation}
where $a$ is a pre-defined parameter and sign($\cdot$) is the signum function. From the definition of $p'(t)$ we can see that the signum of $a$ controls whether $p'(t)$ is of a shape of ``bell'', ``valley'' or ``line'' and $|a|$ controls how steep of the ``bell'' or ``valley''. Then the probability distribution function $p(t)$ for each edge $(i,j)$ is defined by
\begin{equation}\label{eq:p}
  p(t)=\frac{p'(t)}{\int_0^{c_{i,j}}p'(t)dt}.
\end{equation}
For example, if $c_{i,j}=1000$, Figure~\ref{fig:real_p} shows the probability distribution function $p(t)$ when $a = -10$, $-5$, $0$, $5$ and $10$, respectively.

On each TSP instance, we test different $a$ values ranging from $-15$ to $15$ and calculate the $\rho$ values of the generated $(f_1, f_2)$ pairs. Then, eight pairs of $(f_1, f_2)$ with the $\rho$ value ranging from about $-0.5$ to about $0.9$ are selected for the following experiments. The $\rho$ values of the selected eight pairs of $(f_1, f_2)$ on each TSP instance are listed in Table~\ref{tbl:nei_type_TSP}. For the decomposition of the UBQP instances, the probability distribution function $p(t)$ is defined in the interval $(\frac{q_{i,j}}{2} - q', \frac{q_{i,j}}{2} + q')$ (see Section~\ref{sec:UBQP_decomp}). We set $q'=100$ and use the similar method to generate $p(t)$ and decompose each UBQP instance into eight sub-objective pairs. Specially, if $q_{i,j}=0$ in the original UBQP $f$, then we directly let $q_{i,j}^{(1)}=0$ in the sub-objective $f_1$ and $q_{i,j}^{(2)}=0$ in $f_2$. The sub-objective pairs of the UBQP instances are listed in Table~\ref{tbl:nei_type_UBQP}.

On each TSP or UBQP instance, we collect \text{10,000} locally optimal solutions by executing \text{10,000} rounds of local search from different randomly generated solutions. The optimization objective of local search is the original objective $f$. For the TSP instances, the 2-Opt local search is applied and for the UBQP instances, the 1-bit-flip local search is applied. Both local search methods use the first-improvement strategy. Each round of local search starts from a random solution and ends in a local optimum. Then the neighborhood and the neighborhood's neighborhood of these \text{10,000} local optima are evaluated exhaustively, based on the original objective function $f$ and the eight sub-objective function pairs.


Assume that $x'$ is a neighboring solution of a local optimum $x_*$, if $x'$ satisfies that $\exists x''\in\mbox{Neighborhood}(x')$, $f(x'')<f(x_*)$ (in a maximization case, $f(x'')>f(x_*)$), then we denote that $x'$ is a \emph{promising neighboring solution} (P) of the local optimum $x_*$. Otherwise, $x'$ is a \emph{non-promising neighboring solution} (NP) of $x_*$. On the other hand, given a sub-objective pair $(f_1, f_2)$, if $(f_1(x'),f_2(x'))$ is dominated by $(f_1(x_*),f_2(x_*))$, then we denote that $x'$ is a \emph{dominated neighboring solution} (D) of $x_*$. Otherwise, $x'$ is a \emph{non-dominated neighboring solution} (ND) of $x_*$.

In our experiment, for each local optimum, we count the proportion of P (NP) neighboring solutions and the proportion of D (ND) neighboring solutions based on each pair of $(f_1, f_2)$. Then, we average the counting results of the \text{10,000} local optima. Table~\ref{tbl:nei_type_TSP} and Table~\ref{tbl:nei_type_UBQP} list the average proportion of each type of neighboring solution on the TSP instances and the UBQP instances, respectively. In Table~\ref{tbl:nei_type_TSP} and Table~\ref{tbl:nei_type_UBQP}, we also list the average proportion of the cross type of neighboring solutions, e.g., NP\&D indicates the neighboring solutions that are both non-promising and dominated. The last two columns in Table~\ref{tbl:nei_type_TSP} and Table~\ref{tbl:nei_type_UBQP} give the proportion of promising solutions in all the dominated neighboring solution $\left(\frac{\mbox{P\&D}}{\mbox{D}}\right)$ and the proportion of promising solutions in all of the non-dominated neighboring solutions $\left(\frac{\mbox{P\&ND}}{\mbox{ND}}\right)$.

\begin{table}
\caption{Local Optimum Neighborhood Investigate Results on 5 TSP Instances}
\centering
\label{tbl:nei_type_TSP}
\resizebox{\linewidth}{!}{
\begin{tabular}{l|l |l | c c | c c | c c c c | c c }
\hline
\multirow{2}{*}{Instance} & \multirow{2}{*}{\begin{minipage}{70pt}Sub-objective pairs\end{minipage}} & \multirow{2}{*}{\begin{minipage}{65pt}Correlation coefficient\end{minipage}} & \multicolumn{4}{c|}{Neighboring solution type} & \multicolumn{4}{c|}{Cross neighboring solution type} & \multicolumn{2}{c}{Relative ratio} \\
\cline{4-13}
& & & NP & P & D & ND & NP\&D & NP\&ND & P\&D & P\&ND & $\frac{\mbox{P\&D}}{\mbox{D}}$ & $\frac{\mbox{P\&ND}}{\mbox{ND}}$\\
\hline
\multirow{8}{*}{eil51} & $(f_1,f_2)_1$ & $\rho$  = $-$0.5657 & \multirow{8}{*}{98.86\%} & \multirow{8}{*}{1.14\%} & 66.06\% & 33.94\% & 65.72\% & 33.14\% & 0.34\% & 0.81\% & 0.51\% & 2.37\%\\
& $(f_1,f_2)_2$ & $\rho$  = $-$0.3586 & & & 82.85\% & 17.15\% & 82.42\% & 16.43\% & 0.42\% & 0.72\% & 0.51\% & 4.19\%\\
& $(f_1,f_2)_3$ & $\rho$  = $-$0.2271 & & & 88.20\% & 11.80\% & 87.75\% & 11.11\% & 0.46\% & 0.69\% & 0.52\% & 5.82\%\\
& $(f_1,f_2)_4$ & $\rho$  = $-$0.1087 & & & 91.14\% & 8.86\% & 90.59\% & 8.27\% & 0.55\% & 0.59\% & 0.60\% & 6.71\%\\
& $(f_1,f_2)_5$ & $\rho$  = 0.3122 & & & 97.53\% & 2.47\% & 96.78\% & 2.08\% & 0.75\% & 0.40\% & 0.76\% & 16.07\%\\
& $(f_1,f_2)_6$ & $\rho$  = 0.4789 & & & 98.12\% & 1.88\% & 97.30\% & 1.56\% & 0.81\% & 0.33\% & 0.83\% & 17.43\%\\
& $(f_1,f_2)_7$ & $\rho$  = 0.7482 & & & 99.33\% & 0.67\% & 98.35\% & 0.50\% & 0.98\% & 0.16\% & 0.98\% & 24.58\%\\
& $(f_1,f_2)_8$ & $\rho$  = 0.9339 & & & 99.67\% & 0.33\% & 98.60\% & 0.26\% & 1.07\% & 0.07\% & 1.07\% & 21.73\%\\
\hline
\multirow{8}{*}{st70} & $(f_1,f_2)_1$ & $\rho$  = $-$0.5847 & \multirow{8}{*}{99.21\%} & \multirow{8}{*}{0.79\%} & 71.96\% & 28.04\% & 71.75\% & 27.46\% & 0.22\% & 0.57\% & 0.30\% & 2.05\%\\
& $(f_1,f_2)_2$ & $\rho$  = $-$0.3876 & & & 86.72\% & 13.28\% & 86.45\% & 12.76\% & 0.27\% & 0.52\% & 0.31\% & 3.92\%\\
& $(f_1,f_2)_3$ & $\rho$  = $-$0.2535 & & & 91.21\% & 8.79\% & 90.94\% & 8.27\% & 0.28\% & 0.51\% & 0.30\% & 5.83\%\\
& $(f_1,f_2)_4$ & $\rho$  = $-$0.1130 & & & 93.99\% & 6.01\% & 93.69\% & 5.52\% & 0.30\% & 0.49\% & 0.32\% & 8.15\%\\
& $(f_1,f_2)_5$ & $\rho$  = 0.3392 & & & 97.58\% & 2.42\% & 97.12\% & 2.09\% & 0.46\% & 0.33\% & 0.47\% & 13.76\%\\
& $(f_1,f_2)_6$ & $\rho$  = 0.4846 & & & 98.28\% & 1.72\% & 97.81\% & 1.40\% & 0.47\% & 0.32\% & 0.48\% & 18.39\%\\
& $(f_1,f_2)_7$ & $\rho$  = 0.7441 & & & 99.29\% & 0.71\% & 98.67\% & 0.54\% & 0.62\% & 0.17\% & 0.63\% & 23.48\%\\
& $(f_1,f_2)_8$ & $\rho$  = 0.9334 & & & 99.64\% & 0.36\% & 98.93\% & 0.28\% & 0.71\% & 0.08\% & 0.71\% & 22.17\%\\
\hline
\multirow{8}{*}{pr76} & $(f_1,f_2)_1$ & $\rho$  = $-$0.5462 & \multirow{8}{*}{99.24\%} & \multirow{8}{*}{0.76\%} & 68.27\% & 31.73\% & 68.09\% & 31.15\% & 0.19\% & 0.58\% & 0.27\% & 1.83\%\\
& $(f_1,f_2)_2$ & $\rho$  = $-$0.3412 & & & 84.96\% & 15.04\% & 84.72\% & 14.52\% & 0.24\% & 0.52\% & 0.29\% & 3.46\%\\
& $(f_1,f_2)_3$ & $\rho$  = $-$0.1811 & & & 90.25\% & 9.75\% & 89.94\% & 9.29\% & 0.31\% & 0.46\% & 0.34\% & 4.68\%\\
& $(f_1,f_2)_4$ & $\rho$  = $-$0.0223 & & & 94.57\% & 5.43\% & 94.23\% & 5.00\% & 0.34\% & 0.43\% & 0.36\% & 7.89\%\\
& $(f_1,f_2)_5$ & $\rho$  = 0.3535 & & & 98.00\% & 2.00\% & 97.54\% & 1.69\% & 0.45\% & 0.31\% & 0.46\% & 15.55\%\\
& $(f_1,f_2)_6$ & $\rho$  = 0.5442 & & & 98.31\% & 1.69\% & 97.83\% & 1.40\% & 0.48\% & 0.29\% & 0.48\% & 17.11\%\\
& $(f_1,f_2)_7$ & $\rho$  = 0.7753 & & & 99.12\% & 0.88\% & 98.54\% & 0.70\% & 0.59\% & 0.18\% & 0.59\% & 20.37\%\\
& $(f_1,f_2)_8$ & $\rho$  = 0.9371 & & & 99.61\% & 0.39\% & 98.93\% & 0.30\% & 0.67\% & 0.09\% & 0.68\% & 23.10\%\\
\hline
\multirow{8}{*}{rat99} & $(f_1,f_2)_1$ & $\rho$  = $-$0.5066 & \multirow{8}{*}{99.39\%} & \multirow{8}{*}{0.61\%} & 73.14\% & 26.86\% & 72.96\% & 26.43\% & 0.17\% & 0.44\% & 0.23\% & 1.63\%\\
& $(f_1,f_2)_2$ & $\rho$  = $-$0.2872 & & & 88.94\% & 11.06\% & 88.73\% & 10.66\% & 0.21\% & 0.40\% & 0.23\% & 3.63\%\\
& $(f_1,f_2)_3$ & $\rho$  = $-$0.1518 & & & 93.15\% & 6.85\% & 92.93\% & 6.46\% & 0.22\% & 0.39\% & 0.24\% & 5.67\%\\
& $(f_1,f_2)_4$ & $\rho$  = 0.0350 & & & 95.67\% & 4.33\% & 95.39\% & 4.00\% & 0.28\% & 0.33\% & 0.29\% & 7.63\%\\
& $(f_1,f_2)_5$ & $\rho$  = 0.3819 & & & 98.60\% & 1.40\% & 98.24\% & 1.15\% & 0.36\% & 0.25\% & 0.37\% & 17.54\%\\
& $(f_1,f_2)_6$ & $\rho$  = 0.5717 & & & 99.04\% & 0.96\% & 98.64\% & 0.75\% & 0.40\% & 0.21\% & 0.40\% & 22.11\%\\
& $(f_1,f_2)_7$ & $\rho$  = 0.7940 & & & 99.56\% & 0.44\% & 99.06\% & 0.33\% & 0.49\% & 0.12\% & 0.50\% & 26.06\%\\
& $(f_1,f_2)_8$ & $\rho$  = 0.9476 & & & 99.90\% & 0.10\% & 99.33\% & 0.06\% & 0.57\% & 0.04\% & 0.57\% & 37.97\%\\
\hline
\multirow{8}{*}{rd100} & $(f_1,f_2)_1$ & $\rho$  = $-$0.5940 & \multirow{8}{*}{99.46\%} & \multirow{8}{*}{0.54\%} & 72.94\% & 27.06\% & 72.81\% & 26.65\% & 0.14\% & 0.41\% & 0.19\% & 1.51\%\\
& $(f_1,f_2)_2$ & $\rho$  = $-$0.4017 & & & 88.79\% & 11.21\% & 88.65\% & 10.80\% & 0.14\% & 0.40\% & 0.16\% & 3.59\%\\
& $(f_1,f_2)_3$ & $\rho$  = $-$0.2770 & & & 92.30\% & 7.70\% & 92.13\% & 7.33\% & 0.17\% & 0.37\% & 0.19\% & 4.82\%\\
& $(f_1,f_2)_4$ & $\rho$  = $-$0.0821 & & & 95.34\% & 4.66\% & 95.14\% & 4.31\% & 0.20\% & 0.35\% & 0.21\% & 7.43\%\\
& $(f_1,f_2)_5$ & $\rho$  = 0.3106 & & & 98.15\% & 1.85\% & 97.88\% & 1.57\% & 0.27\% & 0.27\% & 0.28\% & 14.80\%\\
& $(f_1,f_2)_6$ & $\rho$  = 0.4873 & & & 98.64\% & 1.36\% & 98.33\% & 1.13\% & 0.31\% & 0.23\% & 0.32\% & 17.01\%\\
& $(f_1,f_2)_7$ & $\rho$  = 0.7526 & & & 99.31\% & 0.69\% & 98.91\% & 0.54\% & 0.39\% & 0.15\% & 0.40\% & 21.71\%\\
& $(f_1,f_2)_8$ & $\rho$  = 0.9374 & & & 99.82\% & 0.18\% & 99.34\% & 0.12\% & 0.48\% & 0.06\% & 0.48\% & 35.42\%\\
\hline
\multicolumn{13}{c}{\begin{minipage}{700pt}*P: promising neighboring solution. NP: non-promising neighboring solution. D: dominated neighboring solution. ND: non-dominated neighboring solution.\end{minipage}}
\end{tabular}
}
\end{table}

\begin{table}
\caption{Local Optimum Neighborhood Investigate Results on 5 UBQP Instances}
\centering
\label{tbl:nei_type_UBQP}
\resizebox{\linewidth}{!}{
\begin{tabular}{l|l |l | c c | c c | c c c c | c c }
\hline
\multirow{2}{*}{Instance} & \multirow{2}{*}{\begin{minipage}{70pt}Sub-objective pairs\end{minipage}} & \multirow{2}{*}{\begin{minipage}{65pt}Correlation coefficient\end{minipage}} & \multicolumn{4}{c|}{Neighboring solution type} & \multicolumn{4}{c|}{Cross neighboring solution type} & \multicolumn{2}{c}{Relative ratio} \\
\cline{4-13}
& & & NP & P & D & ND & NP\&D & NP\&ND & P\&D & P\&ND & $\frac{\mbox{P\&D}}{\mbox{D}}$ & $\frac{\mbox{P\&ND}}{\mbox{ND}}$\\
\hline
\multirow{8}{*}{bqp1000.1} & $(f_1,f_2)_1$ & $\rho$  = $-$0.4162 & \multirow{8}{*}{99.25\%} & \multirow{8}{*}{0.75\%} & 48.11\% & 51.89\% & 48.08\% & 51.17\% & 0.02\% & 0.73\% & 0.05\% & 1.40\%\\
& $(f_1,f_2)_2$ & $\rho$  = $-$0.2055 & & & 55.08\% & 44.92\% & 55.05\% & 44.20\% & 0.03\% & 0.72\% & 0.05\% & 1.60\%\\
& $(f_1,f_2)_3$ & $\rho$  = $-$0.0112 & & & 63.90\% & 36.10\% & 63.86\% & 35.39\% & 0.04\% & 0.71\% & 0.06\% & 1.97\%\\
& $(f_1,f_2)_4$ & $\rho$  = 0.1716 & & & 69.87\% & 30.13\% & 69.82\% & 29.43\% & 0.05\% & 0.70\% & 0.07\% & 2.33\%\\
& $(f_1,f_2)_5$ & $\rho$  = 0.3194 & & & 75.87\% & 24.13\% & 75.82\% & 23.43\% & 0.05\% & 0.70\% & 0.07\% & 2.89\%\\
& $(f_1,f_2)_6$ & $\rho$  = 0.5270 & & & 81.52\% & 18.48\% & 81.45\% & 17.80\% & 0.07\% & 0.68\% & 0.08\% & 3.69\%\\
& $(f_1,f_2)_7$ & $\rho$  = 0.7417 & & & 88.95\% & 11.05\% & 88.86\% & 10.39\% & 0.09\% & 0.66\% & 0.10\% & 5.97\%\\
& $(f_1,f_2)_8$ & $\rho$  = 0.9231 & & & 95.62\% & 4.38\% & 95.45\% & 3.80\% & 0.18\% & 0.57\% & 0.19\% & 13.06\%\\
\hline
\multirow{8}{*}{bqp2500.1} & $(f_1,f_2)_1$ & $\rho$  = $-$0.4128 & \multirow{8}{*}{99.64\%} & \multirow{8}{*}{0.36\%} & 48.41\% & 51.59\% & 48.40\% & 51.23\% & 0.01\% & 0.36\% & 0.02\% & 0.69\%\\
& $(f_1,f_2)_2$ & $\rho$  = $-$0.2008 & & & 57.17\% & 42.83\% & 57.16\% & 42.48\% & 0.01\% & 0.35\% & 0.02\% & 0.83\%\\
& $(f_1,f_2)_3$ & $\rho$  = $-$0.0073 & & & 64.30\% & 35.70\% & 64.29\% & 35.35\% & 0.01\% & 0.35\% & 0.02\% & 0.99\%\\
& $(f_1,f_2)_4$ & $\rho$  = 0.1726 & & & 70.49\% & 29.51\% & 70.47\% & 29.16\% & 0.01\% & 0.35\% & 0.02\% & 1.18\%\\
& $(f_1,f_2)_5$ & $\rho$  = 0.3225 & & & 75.34\% & 24.66\% & 75.32\% & 24.31\% & 0.02\% & 0.35\% & 0.02\% & 1.41\%\\
& $(f_1,f_2)_6$ & $\rho$  = 0.5289 & & & 82.52\% & 17.48\% & 82.49\% & 17.14\% & 0.02\% & 0.34\% & 0.03\% & 1.96\%\\
& $(f_1,f_2)_7$ & $\rho$  = 0.7451 & & & 88.87\% & 11.13\% & 88.84\% & 10.79\% & 0.03\% & 0.33\% & 0.04\% & 2.99\%\\
& $(f_1,f_2)_8$ & $\rho$  = 0.9241 & & & 95.82\% & 4.18\% & 95.76\% & 3.88\% & 0.06\% & 0.31\% & 0.06\% & 7.32\%\\
\hline
\multirow{8}{*}{p3000.1} & $(f_1,f_2)_1$ & $\rho$  = $-$0.4087 & \multirow{8}{*}{99.90\%} & \multirow{8}{*}{0.10\%} & 48.30\% & 51.70\% & 48.30\% & 51.60\% & 0.00\% & 0.10\% & 0.00\% & 0.19\%\\
& $(f_1,f_2)_2$ & $\rho$  = $-$0.1957 & & & 56.05\% & 43.95\% & 56.05\% & 43.85\% & 0.00\% & 0.10\% & 0.00\% & 0.23\%\\
& $(f_1,f_2)_3$ & $\rho$  = $-$0.0024 & & & 64.48\% & 35.52\% & 64.48\% & 35.42\% & 0.00\% & 0.10\% & 0.00\% & 0.28\%\\
& $(f_1,f_2)_4$ & $\rho$  = 0.1797 & & & 70.02\% & 29.98\% & 70.02\% & 29.88\% & 0.00\% & 0.10\% & 0.00\% & 0.33\%\\
& $(f_1,f_2)_5$ & $\rho$  = 0.3273 & & & 75.43\% & 24.57\% & 75.43\% & 24.47\% & 0.00\% & 0.10\% & 0.00\% & 0.41\%\\
& $(f_1,f_2)_6$ & $\rho$  = 0.5328 & & & 82.47\% & 17.53\% & 82.47\% & 17.43\% & 0.00\% & 0.10\% & 0.00\% & 0.56\%\\
& $(f_1,f_2)_7$ & $\rho$  = 0.7466 & & & 89.54\% & 10.46\% & 89.53\% & 10.36\% & 0.00\% & 0.10\% & 0.00\% & 0.94\%\\
& $(f_1,f_2)_8$ & $\rho$  = 0.9248 & & & 96.14\% & 3.86\% & 96.13\% & 3.77\% & 0.01\% & 0.09\% & 0.01\% & 2.45\%\\
\hline
\multirow{8}{*}{p4000.1} & $(f_1,f_2)_1$ & $\rho$  = $-$0.4087 & \multirow{8}{*}{99.92\%} & \multirow{8}{*}{0.08\%} & 48.39\% & 51.61\% & 48.39\% & 51.53\% & 0.00\% & 0.08\% & 0.00\% & 0.15\%\\
& $(f_1,f_2)_2$ & $\rho$  = $-$0.1955 & & & 58.16\% & 41.84\% & 58.16\% & 41.76\% & 0.00\% & 0.08\% & 0.00\% & 0.19\%\\
& $(f_1,f_2)_3$ & $\rho$  = $-$0.0027 & & & 63.73\% & 36.27\% & 63.73\% & 36.19\% & 0.00\% & 0.08\% & 0.00\% & 0.22\%\\
& $(f_1,f_2)_4$ & $\rho$  = 0.1793 & & & 70.35\% & 29.65\% & 70.35\% & 29.57\% & 0.00\% & 0.08\% & 0.00\% & 0.27\%\\
& $(f_1,f_2)_5$ & $\rho$  = 0.3264 & & & 75.81\% & 24.19\% & 75.81\% & 24.11\% & 0.00\% & 0.08\% & 0.00\% & 0.33\%\\
& $(f_1,f_2)_6$ & $\rho$  = 0.5335 & & & 82.88\% & 17.12\% & 82.88\% & 17.04\% & 0.00\% & 0.08\% & 0.00\% & 0.46\%\\
& $(f_1,f_2)_7$ & $\rho$  = 0.7466 & & & 89.93\% & 10.07\% & 89.93\% & 9.99\% & 0.00\% & 0.08\% & 0.00\% & 0.77\%\\
& $(f_1,f_2)_8$ & $\rho$  = 0.9248 & & & 96.56\% & 3.44\% & 96.56\% & 3.36\% & 0.01\% & 0.08\% & 0.01\% & 2.18\%\\
\hline
\multirow{8}{*}{p5000.1} & $(f_1,f_2)_1$ & $\rho$  = $-$0.4085 & \multirow{8}{*}{99.94\%} & \multirow{8}{*}{0.06\%} & 48.65\% & 51.35\% & 48.65\% & 51.29\% & 0.00\% & 0.06\% & 0.00\% & 0.12\%\\
& $(f_1,f_2)_2$ & $\rho$  = $-$0.1949 & & & 57.69\% & 42.31\% & 57.69\% & 42.24\% & 0.00\% & 0.06\% & 0.00\% & 0.15\%\\
& $(f_1,f_2)_3$ & $\rho$  = $-$0.0025 & & & 64.50\% & 35.50\% & 64.50\% & 35.44\% & 0.00\% & 0.06\% & 0.00\% & 0.18\%\\
& $(f_1,f_2)_4$ & $\rho$  = 0.1801 & & & 70.28\% & 29.72\% & 70.28\% & 29.65\% & 0.00\% & 0.06\% & 0.00\% & 0.21\%\\
& $(f_1,f_2)_5$ & $\rho$  = 0.3267 & & & 75.33\% & 24.67\% & 75.33\% & 24.61\% & 0.00\% & 0.06\% & 0.00\% & 0.26\%\\
& $(f_1,f_2)_6$ & $\rho$  = 0.5330 & & & 82.50\% & 17.50\% & 82.50\% & 17.43\% & 0.00\% & 0.06\% & 0.00\% & 0.36\%\\
& $(f_1,f_2)_7$ & $\rho$  = 0.7467 & & & 89.48\% & 10.52\% & 89.47\% & 10.46\% & 0.00\% & 0.06\% & 0.00\% & 0.60\%\\
& $(f_1,f_2)_8$ & $\rho$  = 0.9247 & & & 96.46\% & 3.54\% & 96.45\% & 3.48\% & 0.00\% & 0.06\% & 0.00\% & 1.72\%\\
\hline
\multicolumn{13}{c}{\begin{minipage}{700pt}*P: promising neighboring solution. NP: non-promising neighboring solution. D: dominated neighboring solution. ND: non-dominated neighboring solution.\end{minipage}}
\end{tabular}
}
\end{table}

Based on the results in Table~\ref{tbl:nei_type_TSP} and Table~\ref{tbl:nei_type_UBQP}, we have the following observations.

First, by comparing column ``NP'' and column ``P'' in Table~\ref{tbl:nei_type_TSP} and Table~\ref{tbl:nei_type_UBQP} we can see that on all the TSP/UBQP instances, in all the neighboring solutions of a local optimum, the proportion of the promising neighboring solutions is quite low. The proportion decreases further as the problem size increases. This indicates that it is quite hard to find a promising neighboring solution.

Second, by observing column ``D'' and column ``ND'' in Table~\ref{tbl:nei_type_TSP} and Table~\ref{tbl:nei_type_UBQP} we can see that in most cases, the proportion of the non-dominated neighboring solutions is significantly lower than that of the dominated neighboring solutions, except when $\rho$ is very small (e.g. when $\rho=-0.4162$ for the UBQP instance bqp1000.1 in Table~\ref{tbl:nei_type_UBQP}). In addition, by increasing $\rho$, the proportion of the non-dominated neighboring solutions further decreases. This is intuitive since when $\rho\to1$, $(f_1(x), f_2(x))\approx(f(x)/2, f(x)/2)$ for any solution in the solution space (as shown in Figure~\ref{fig:corr_example}) and $(f(x')/2, f(x')/2) \prec (f(x*)/2, f(x*)/2)$ if $x_*$ is a local optimum and $x'$ is one of its neighboring solutions.

Third, by observing columns ``NP\&D'', ``NP\&ND'', ``P\&D'' and ``P\&ND'' in Table~\ref{tbl:nei_type_TSP} and Table~\ref{tbl:nei_type_UBQP} we can see that in most cases, the proportion of the both non-promising and dominated neighboring solutions (NP\&D) is the highest among the four cross types of neighboring solutions. The proportion of the intersection of the promising and dominated neighboring solutions (P\&D) is very small in most cases. This means that if a neighboring solution is dominated by the original local optimum, then it is very likely that this solution is non-promising.

Fourth, in the last two columns we list the ratios ``$\frac{\mbox{P\&D}}{\mbox{D}}$'' and ``$\frac{\mbox{P\&ND}}{\mbox{ND}}$''. The ratios reflect the probability to find a promising neighboring solution in the dominated neighboring solutions and in the non-dominated neighboring solutions, respectively. We can see that the chance to find a promising neighboring solution in the non-dominated neighboring solutions is significantly higher than that in the dominated neighboring solution. It is also higher than the probability to find a promising solution in all the neighboring solutions (column ``P''). This supports the neighborhood non-dominance hypothesis that the non-dominated neighboring solutions of a local optimum are more likely to contain a neighboring solution that improves the local optimum.

\begin{figure}
  \centering
  \includegraphics[width=0.4\linewidth]{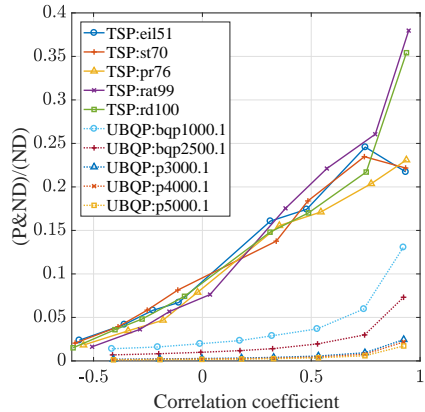}\\
  \caption{$\frac{\mbox{P\&ND}}{\mbox{ND}}$ versus correlation coefficient $\rho$ in Table~\ref{tbl:nei_type_TSP} and Table~\ref{tbl:nei_type_UBQP}.}\label{fig:P_ND_ND}
\end{figure}

Fifth, from Table~\ref{tbl:nei_type_TSP} and Table~\ref{tbl:nei_type_UBQP}, we see that the correlation between the sub-objectives have a significant influence on the ratio $\frac{\mbox{P\&ND}}{\mbox{ND}}$ (see Figure~\ref{fig:P_ND_ND}). On most test instances, we observed that the ratio increases along the increasing of $\rho$. However, the correlation should not be too large, since when $\rho=1$ the bi-objective problem with the objectives $(f_1, f_2)$ will degenerate into a single-objective problem (because $f_1=f_2$) and there will be no non-dominated neighboring solution. On the small size TSP instances eil51 and st70, we observed the decrease of the ratio $\frac{\mbox{P\&ND}}{\mbox{ND}}$ when $\rho$ increases from about $0.7$ to about $0.9$.

Particularly, in Table~\ref{tbl:nei_type_UBQP}, an interesting phenomenon is that on the UBQP instances the proportion of the intersection of the promising and dominated neighboring solutions (P\&D) is extremely low, especially on large UBQP instances. For example, on the UBQP instance p5000.1, the P\&D proportion is 0 for all the eight sub-objective pairs.

\begin{figure}
  \centering
  \subfigure[$\rho=-0.5657$]{
    \label{fig:nei_decomp_example_eil51_1}
    \includegraphics[width=0.23\linewidth]{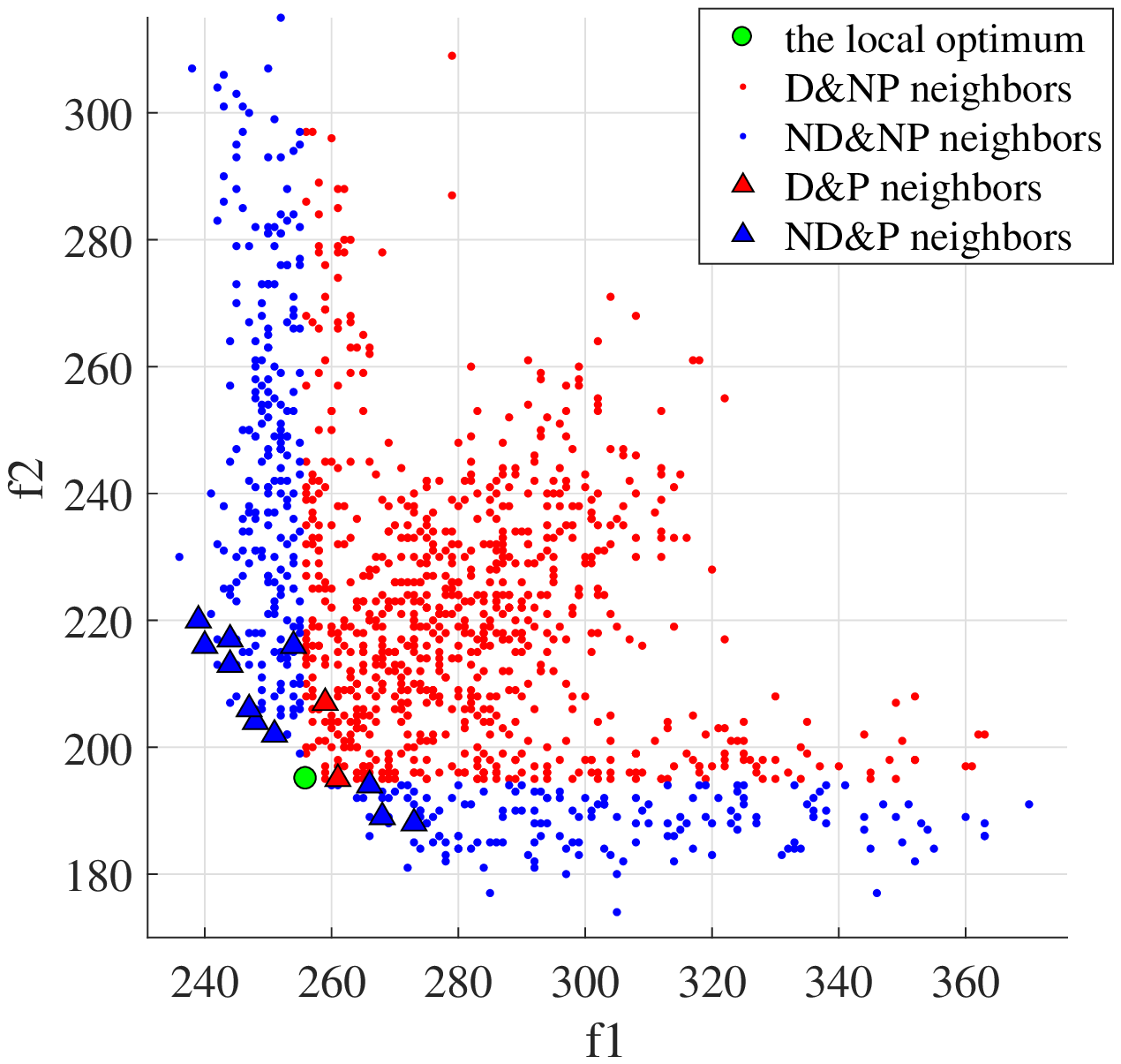}}
    \hspace{-0.005\linewidth}
  \subfigure[$\rho=-0.3586$]{
    \label{fig:nei_decomp_example_eil51_2}
    \includegraphics[width=0.23\linewidth]{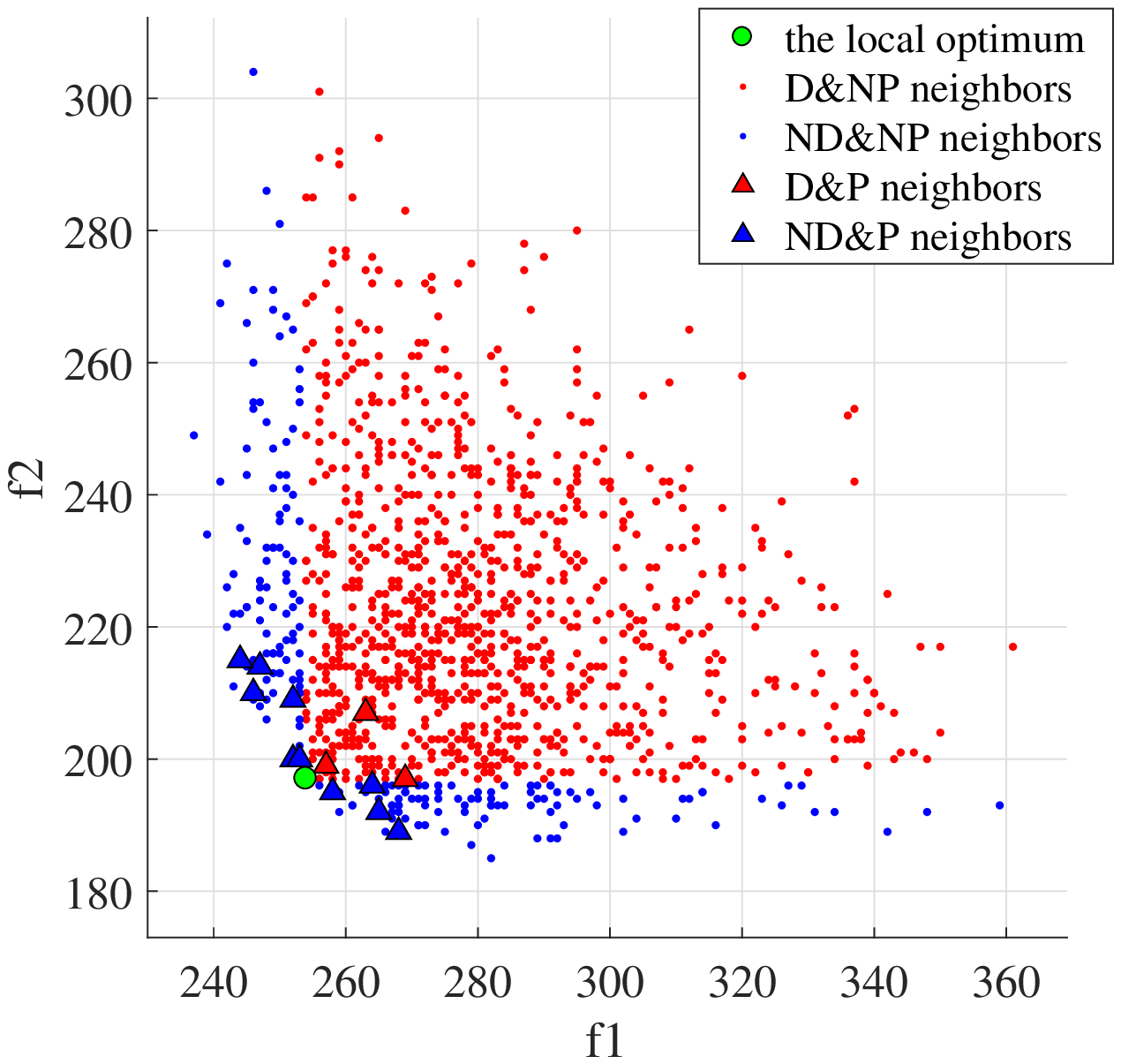}}
    \hspace{-0.005\linewidth}
  \subfigure[$\rho=-0.2271$]{
    \label{fig:nei_decomp_example_eil51_3}
    \includegraphics[width=0.23\linewidth]{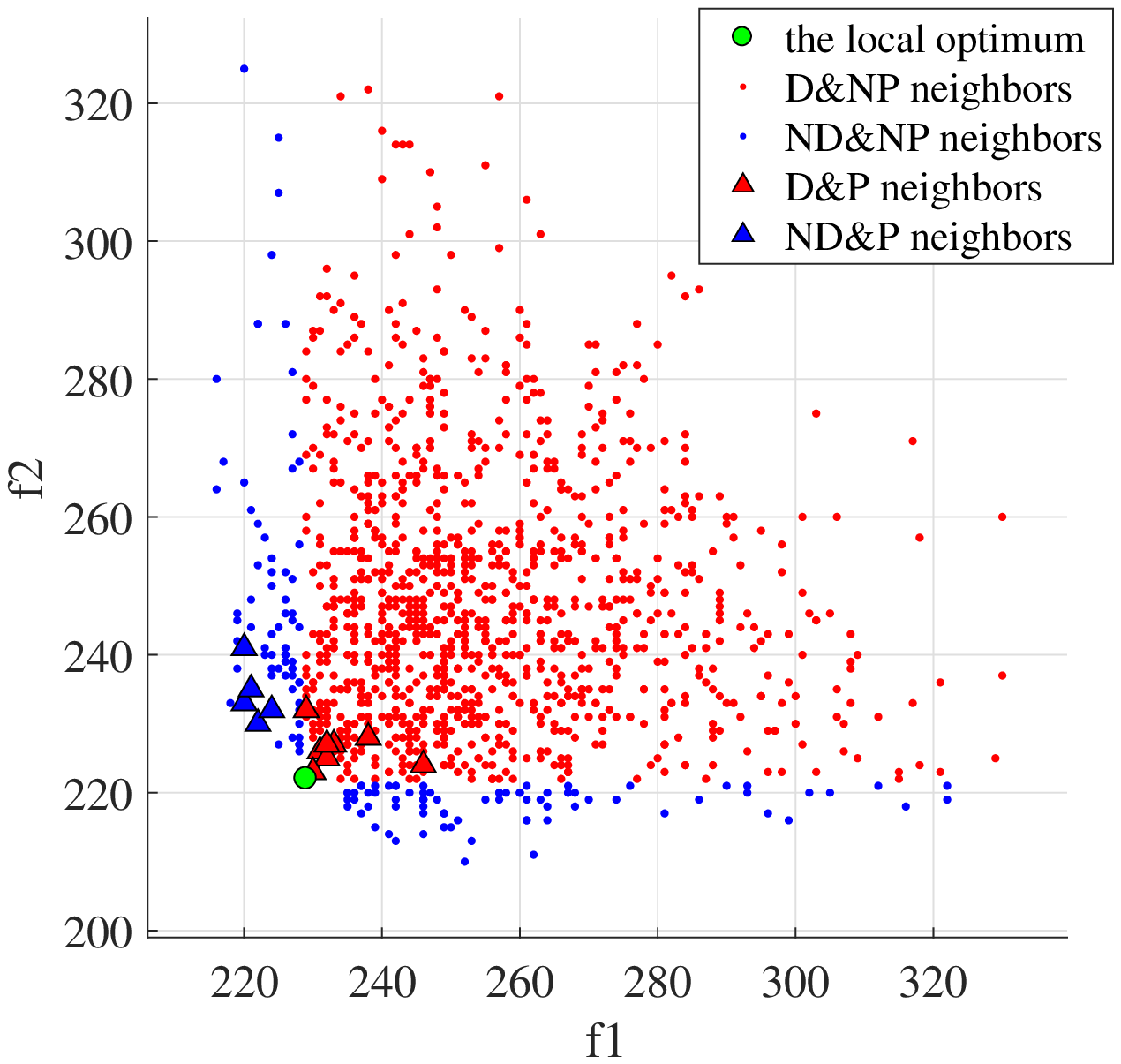}}
    \hspace{-0.005\linewidth}
  \subfigure[$\rho=-0.1087$]{
    \label{fig:nei_decomp_example_eil51_4}
    \includegraphics[width=0.23\linewidth]{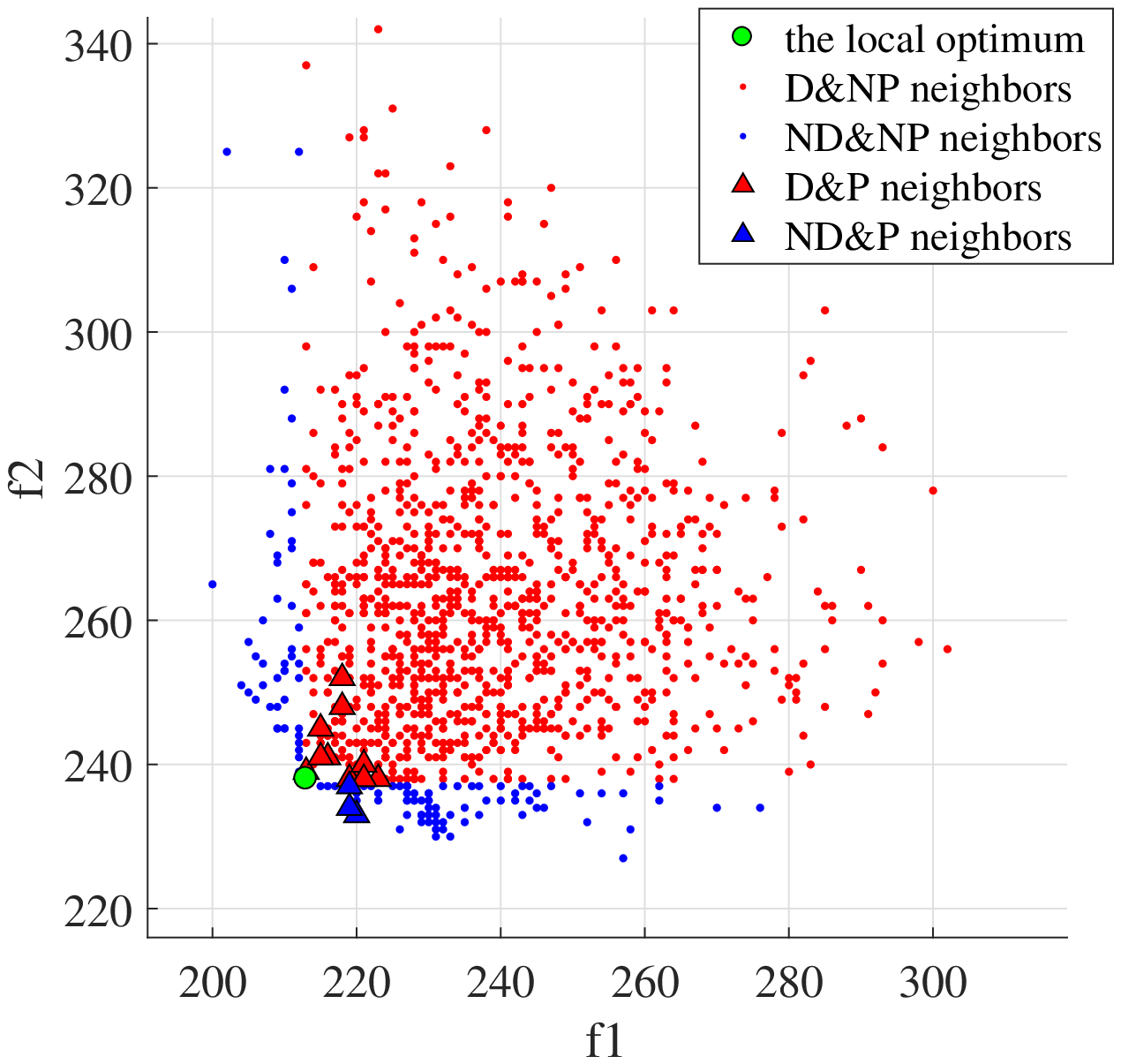}}\\
    \vspace{-0.005\linewidth}
  \subfigure[$\rho=0.3122$]{
    \label{fig:nei_decomp_example_eil51_5}
    \includegraphics[width=0.23\linewidth]{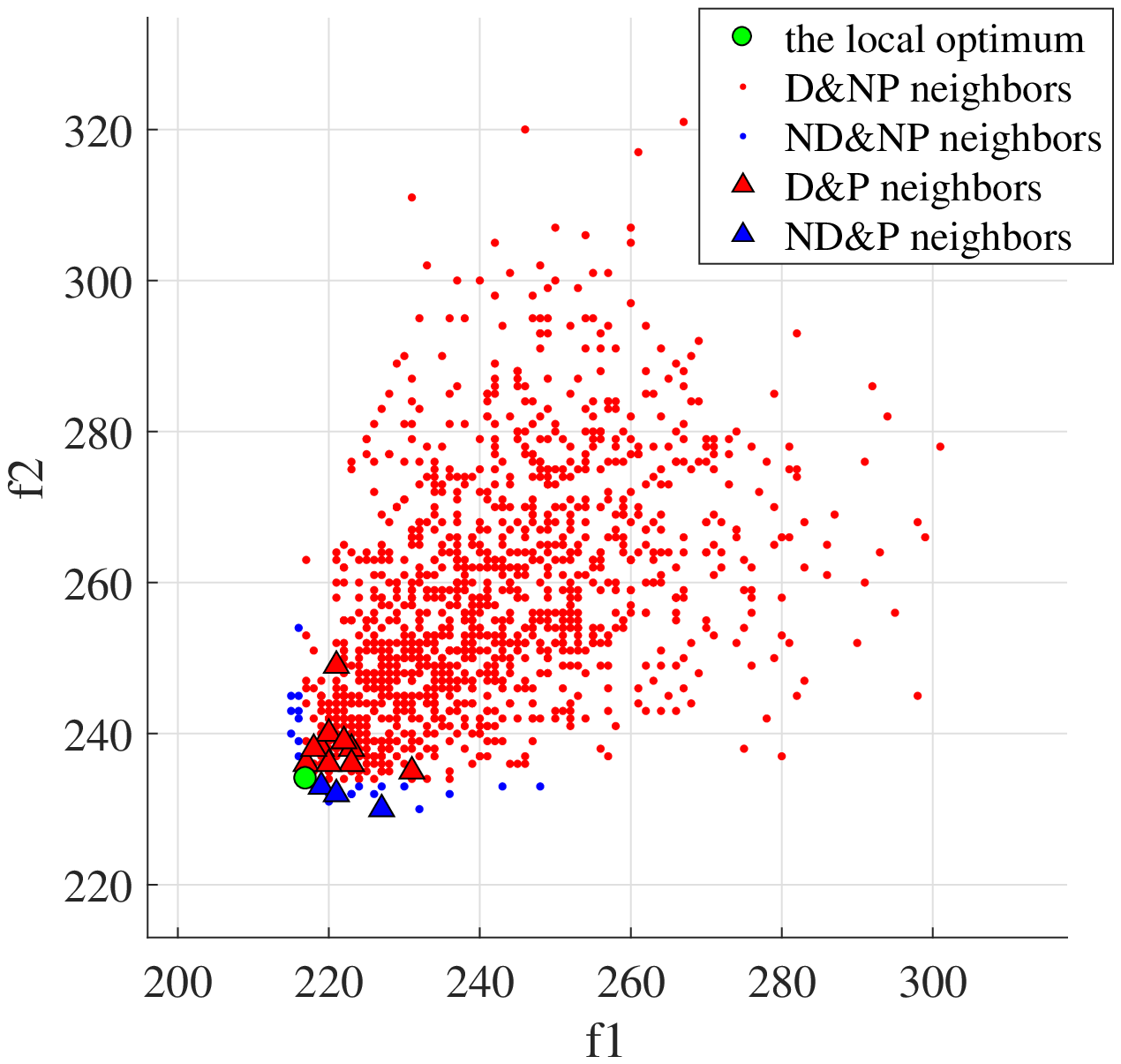}}
    \hspace{-0.005\linewidth}
  \subfigure[$\rho=0.4789$]{
    \label{fig:nei_decomp_example_eil51_6}
    \includegraphics[width=0.23\linewidth]{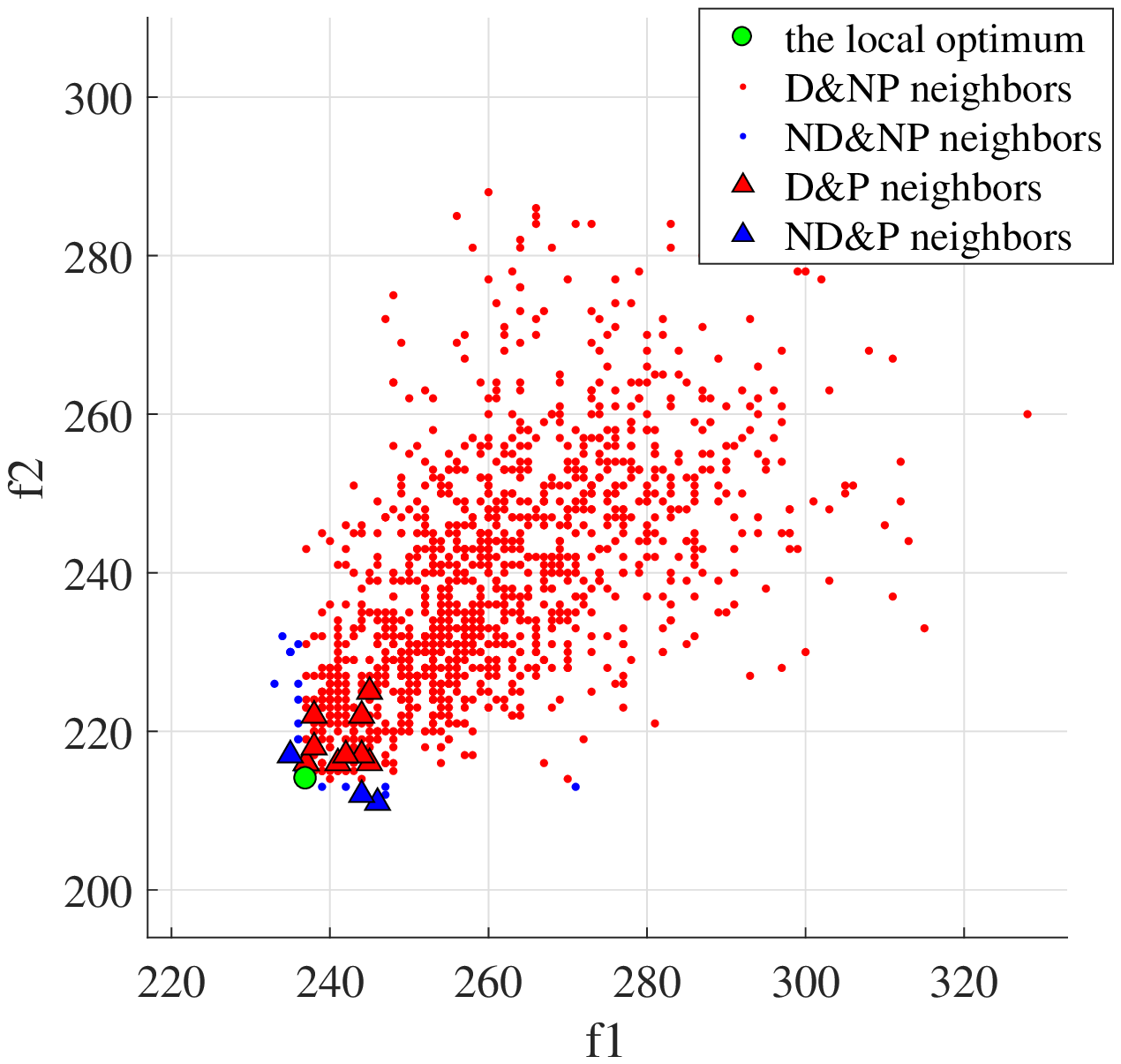}}
    \hspace{-0.005\linewidth}
  \subfigure[$\rho=0.7482$]{
    \label{fig:nei_decomp_example_eil51_7}
    \includegraphics[width=0.23\linewidth]{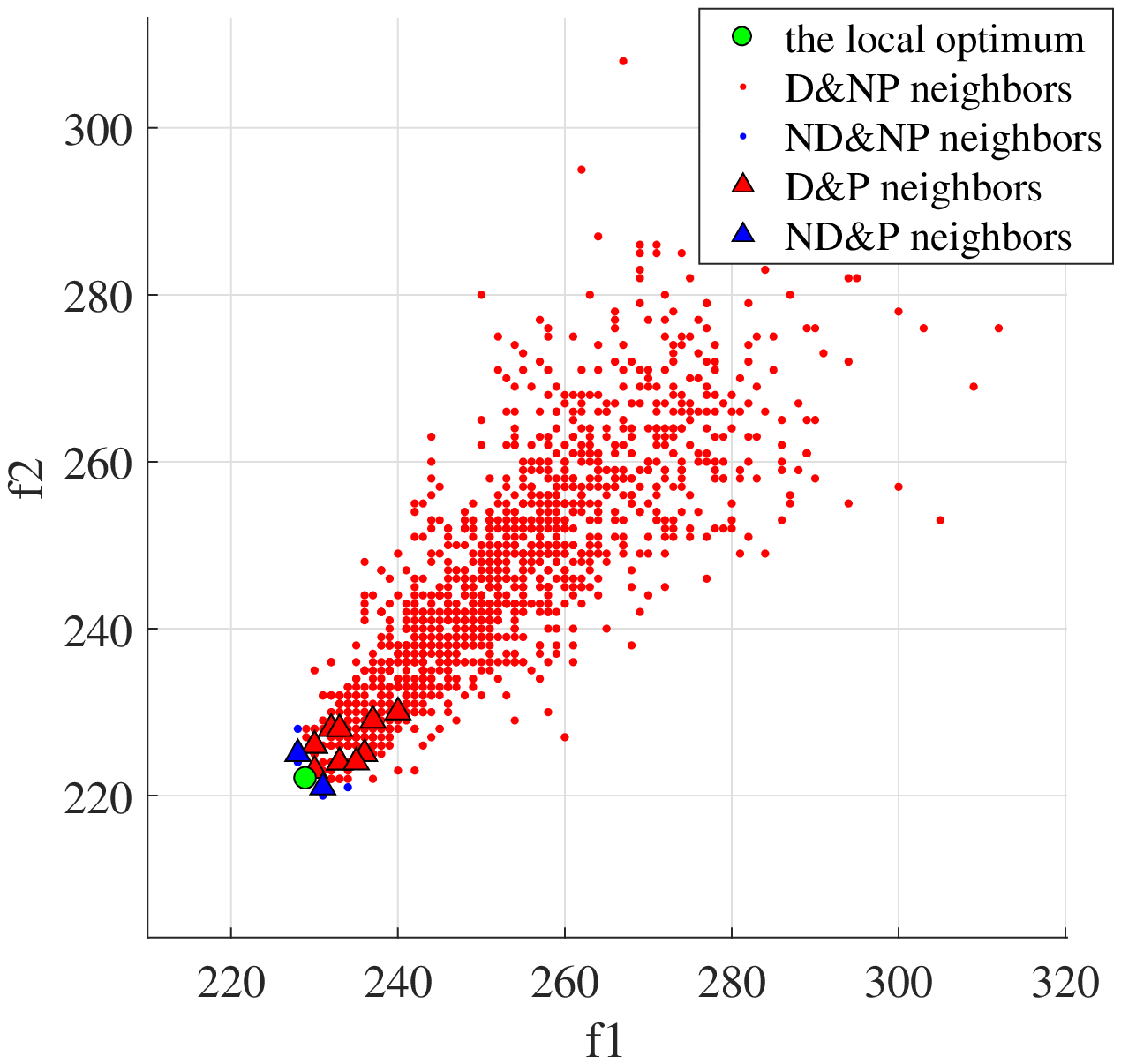}}
    \hspace{-0.005\linewidth}
  \subfigure[$\rho=0.9330$]{
    \label{fig:nei_decomp_example_eil51_8}
    \includegraphics[width=0.23\linewidth]{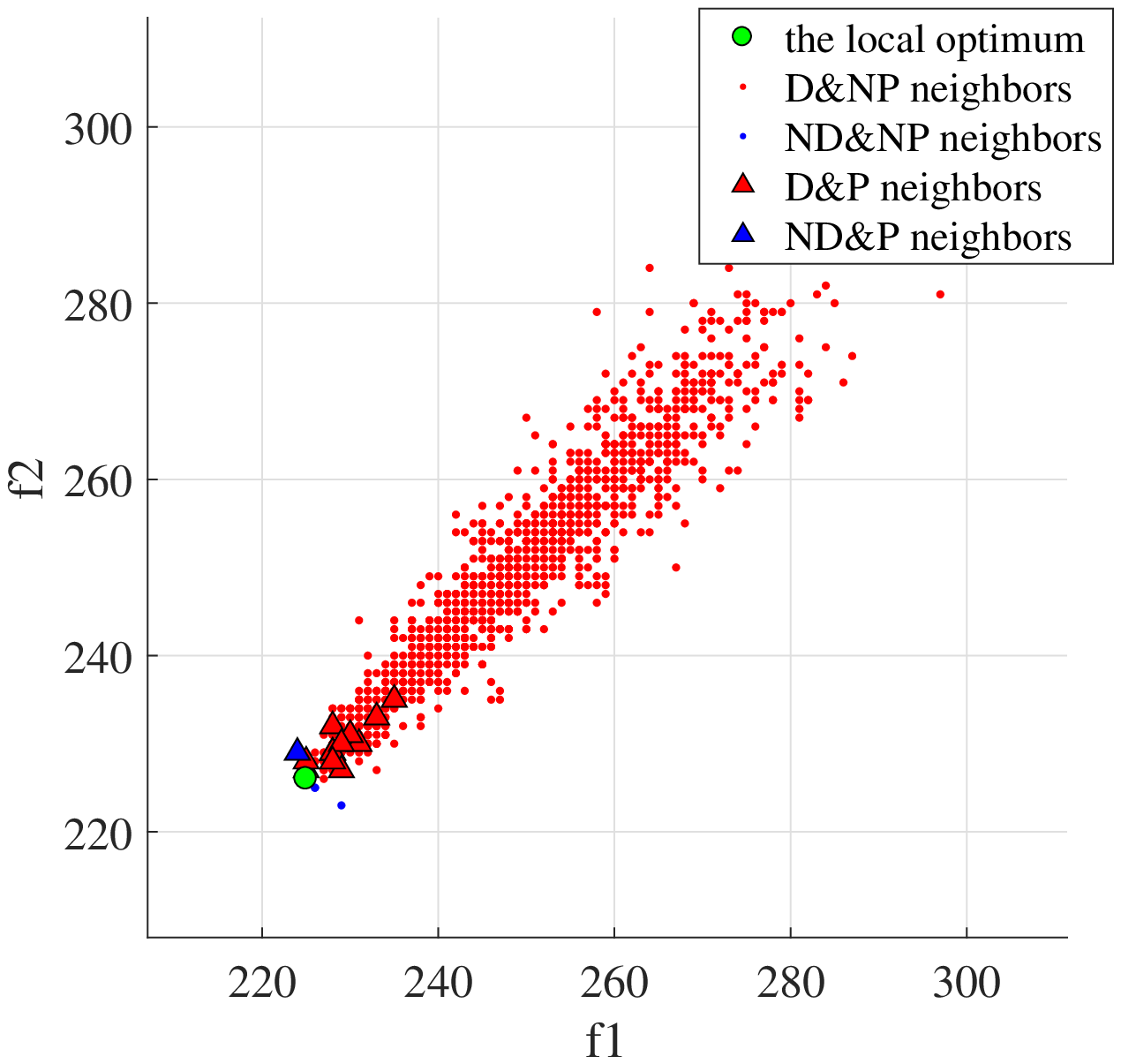}}\\
 \caption{The layout of different types of neighboring solutions of a 2-Opt local optimum of the TSP instance eil51. The neighboring solutions are plotted on 8 sub-objective pairs with different correlation coefficients.}\label{fig:nei_decomp_example_eil51}
\end{figure}

\begin{figure}
  \centering
  \subfigure[$\rho=-0.4162$]{
    \label{fig:nei_decomp_example_bqp1000_1_1}
    \includegraphics[width=0.23\linewidth]{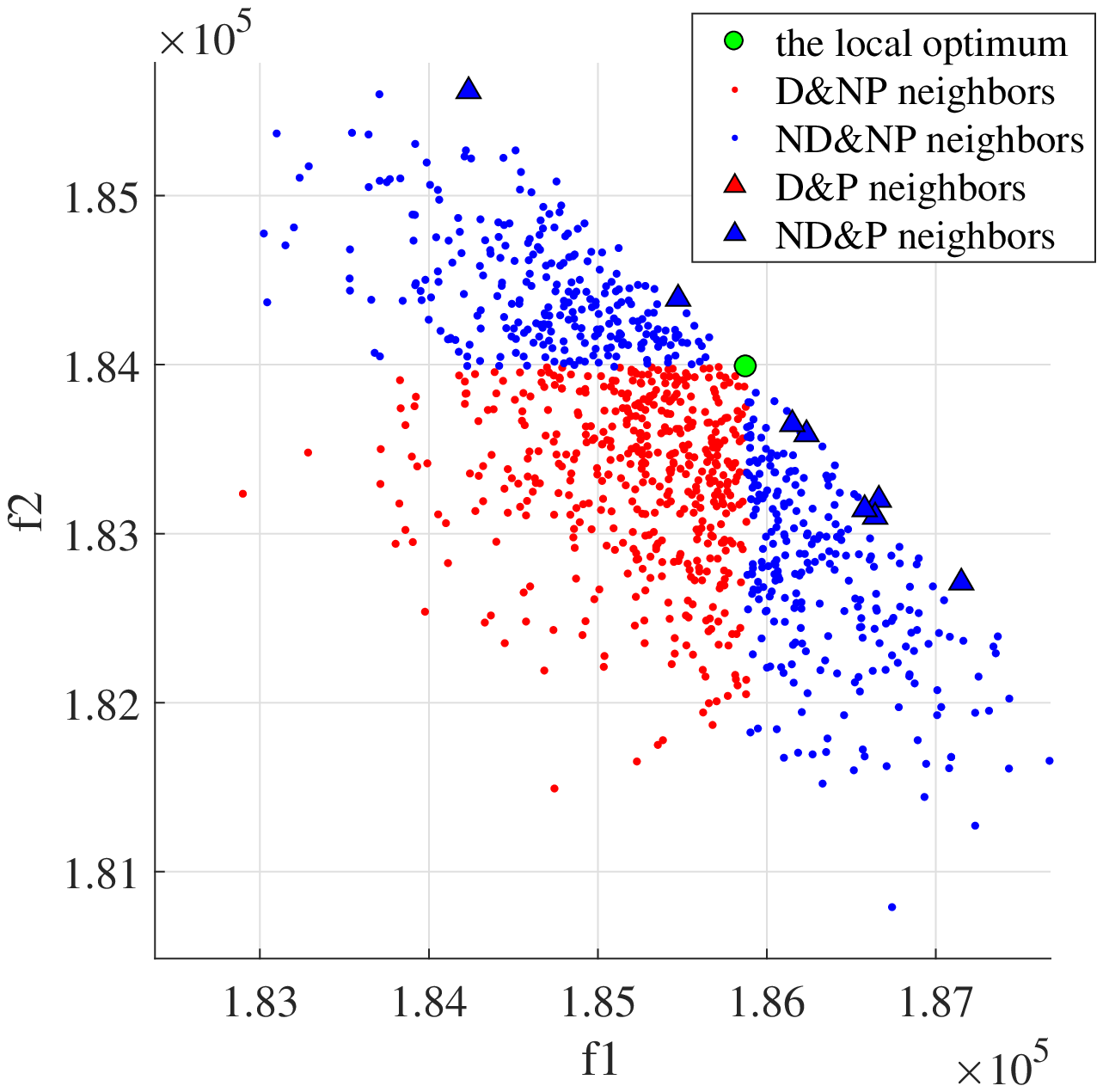}}
    \hspace{-0.005\linewidth}
  \subfigure[$\rho=-0.2055$]{
    \label{fig:nei_decomp_example_bqp1000_1_2}
    \includegraphics[width=0.23\linewidth]{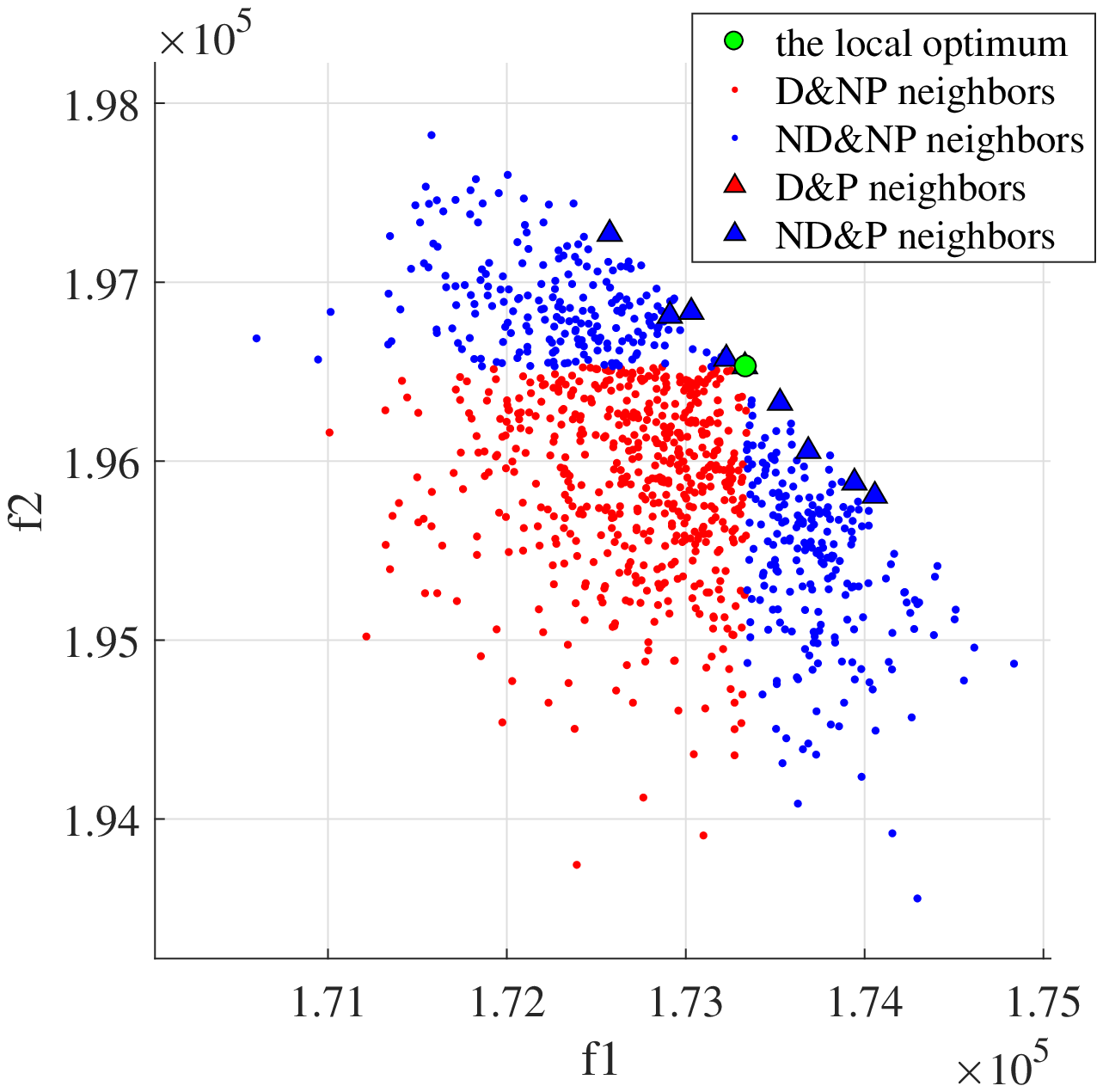}}
    \hspace{-0.005\linewidth}
  \subfigure[$\rho=-0.0112$]{
    \label{fig:nei_decomp_example_bqp1000_1_3}
    \includegraphics[width=0.23\linewidth]{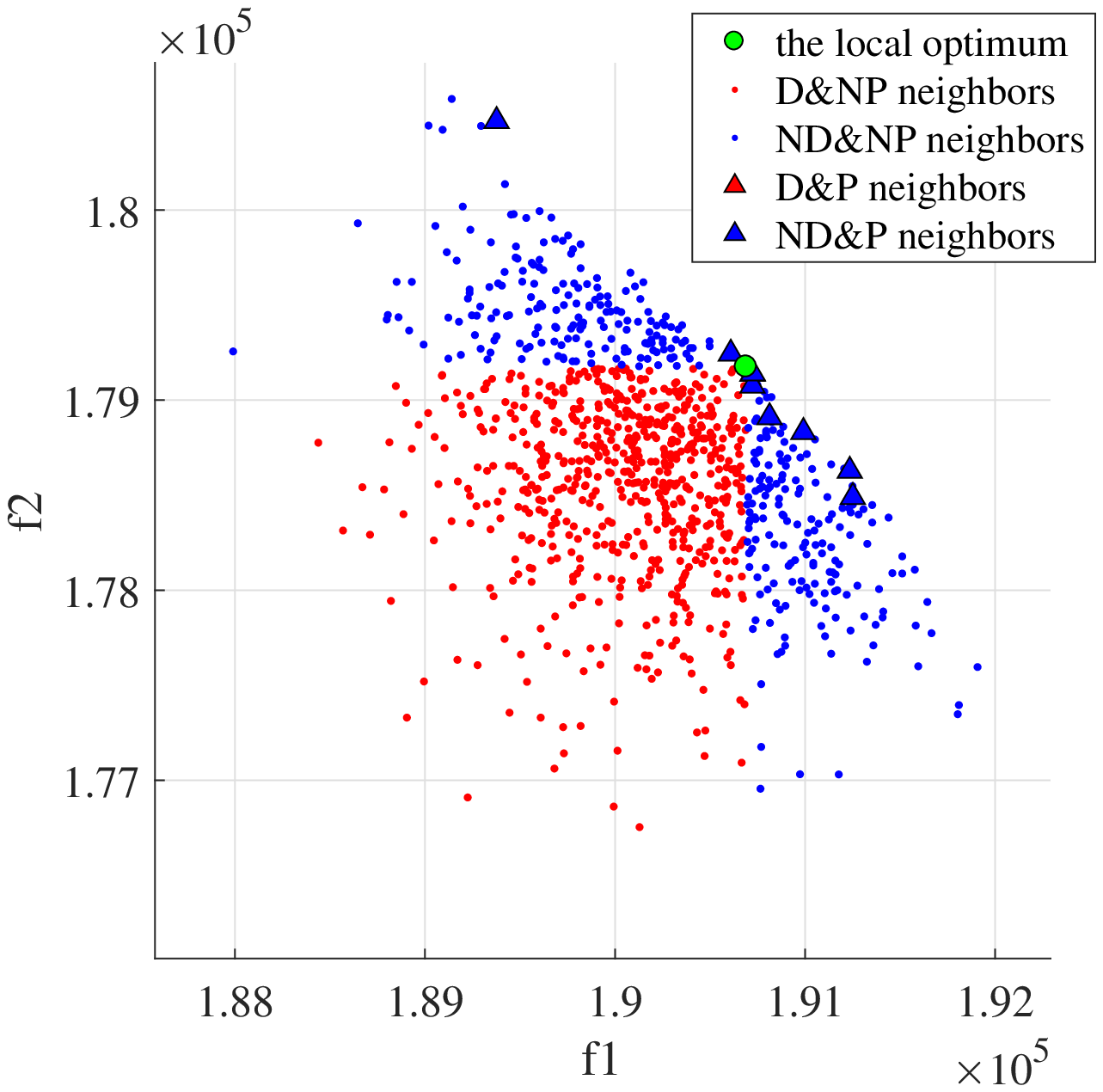}}
    \hspace{-0.005\linewidth}
  \subfigure[$\rho=0.1716$]{
    \label{fig:nei_decomp_example_bqp1000_1_4}
    \includegraphics[width=0.23\linewidth]{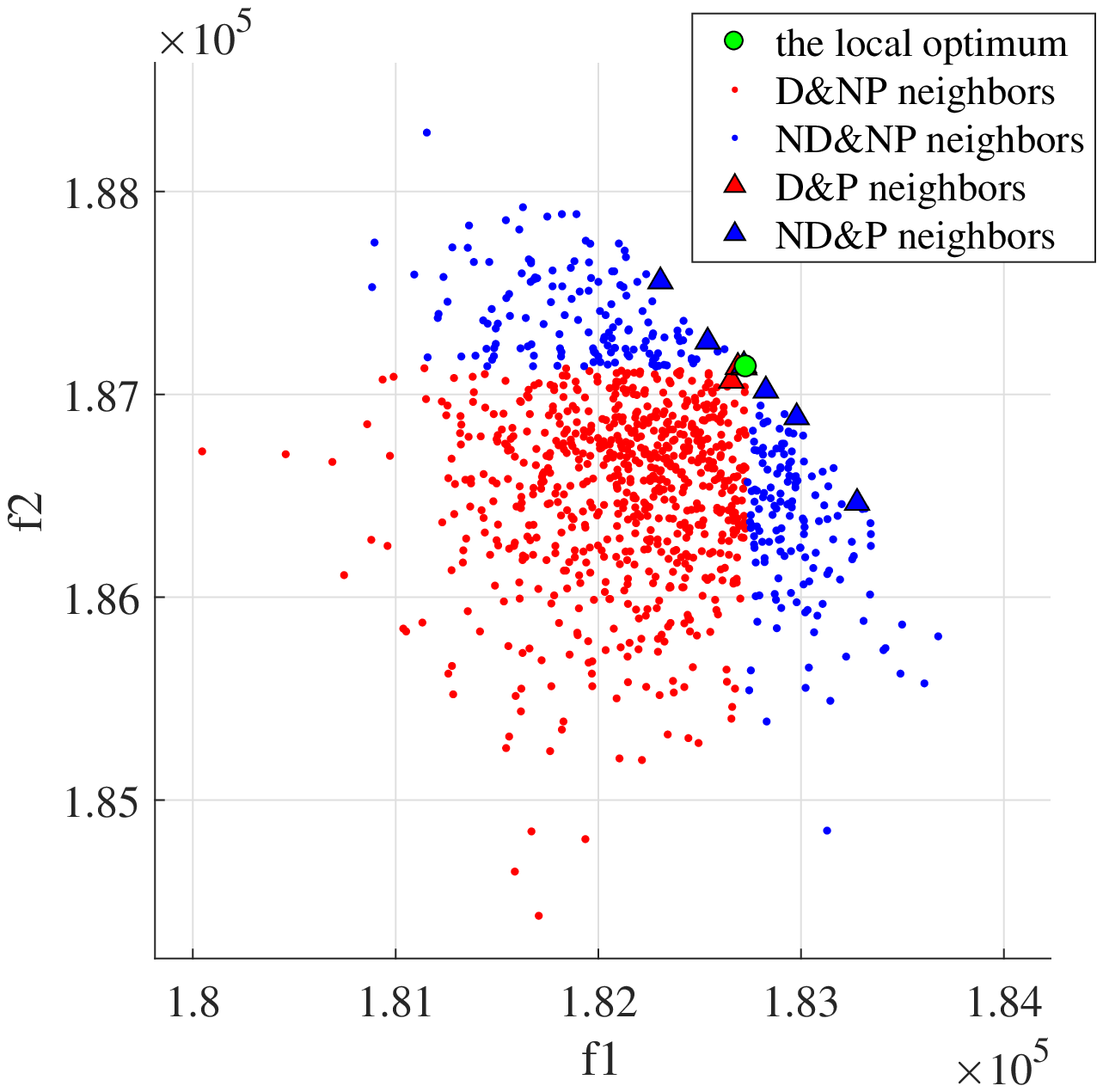}}\\
    \vspace{-0.005\linewidth}
  \subfigure[$\rho=0.3194$]{
    \label{fig:nei_decomp_example_bqp1000_1_5}
    \includegraphics[width=0.23\linewidth]{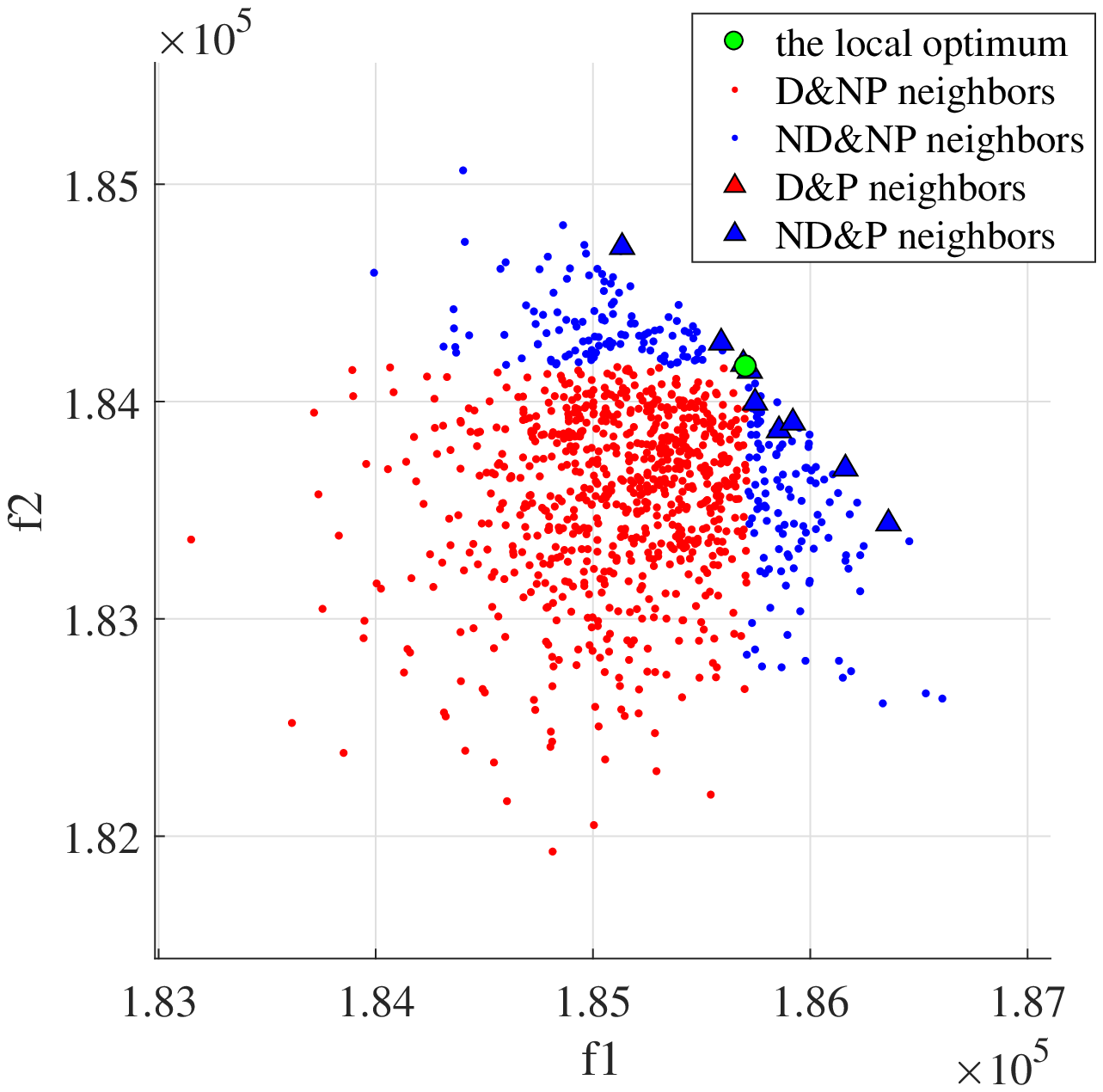}}
    \hspace{-0.005\linewidth}
  \subfigure[$\rho=0.5270$]{
    \label{fig:nei_decomp_example_bqp1000_1_6}
    \includegraphics[width=0.23\linewidth]{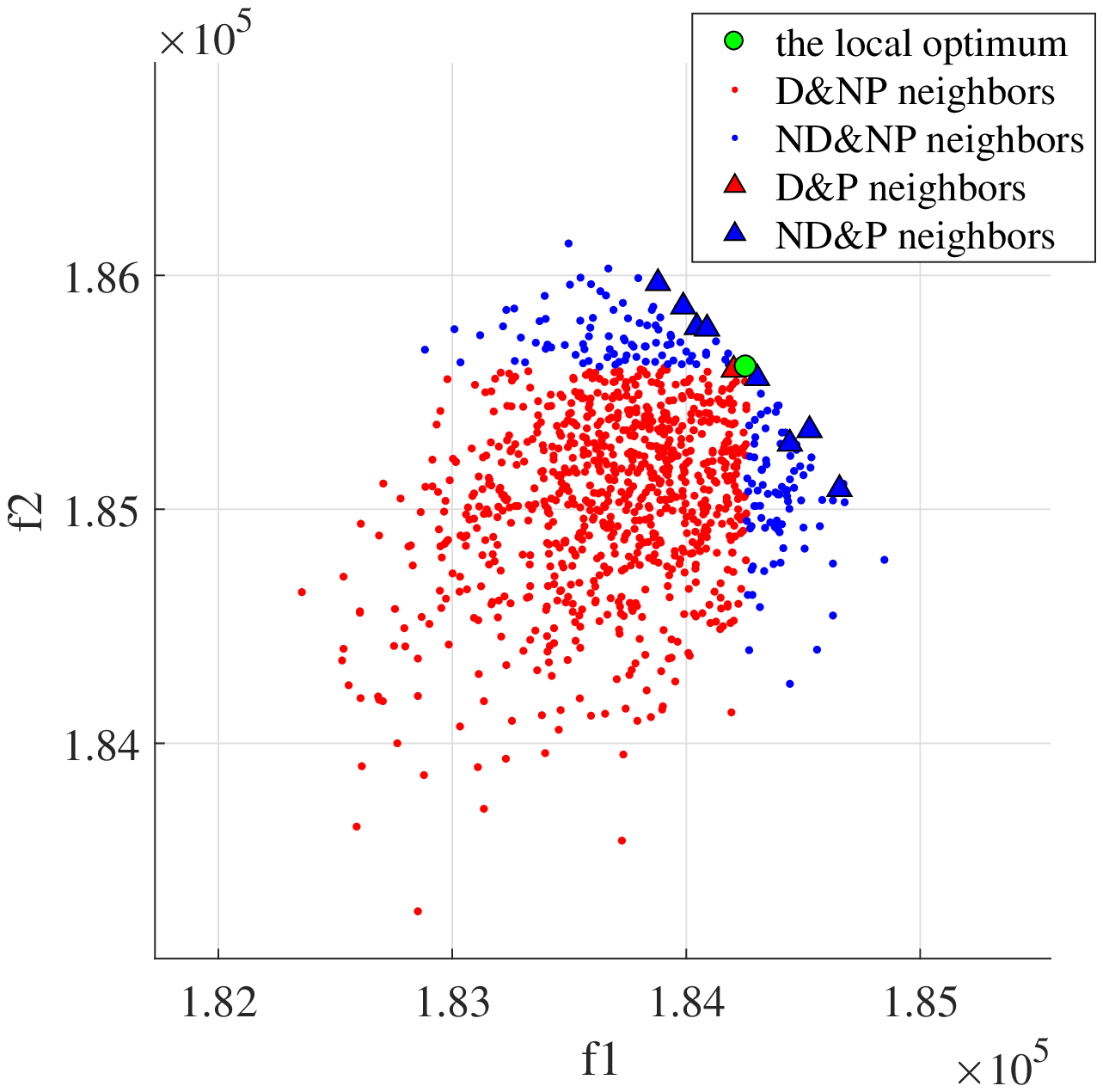}}
    \hspace{-0.005\linewidth}
  \subfigure[$\rho=0.7417$]{
    \label{fig:nei_decomp_example_bqp1000_1_7}
    \includegraphics[width=0.23\linewidth]{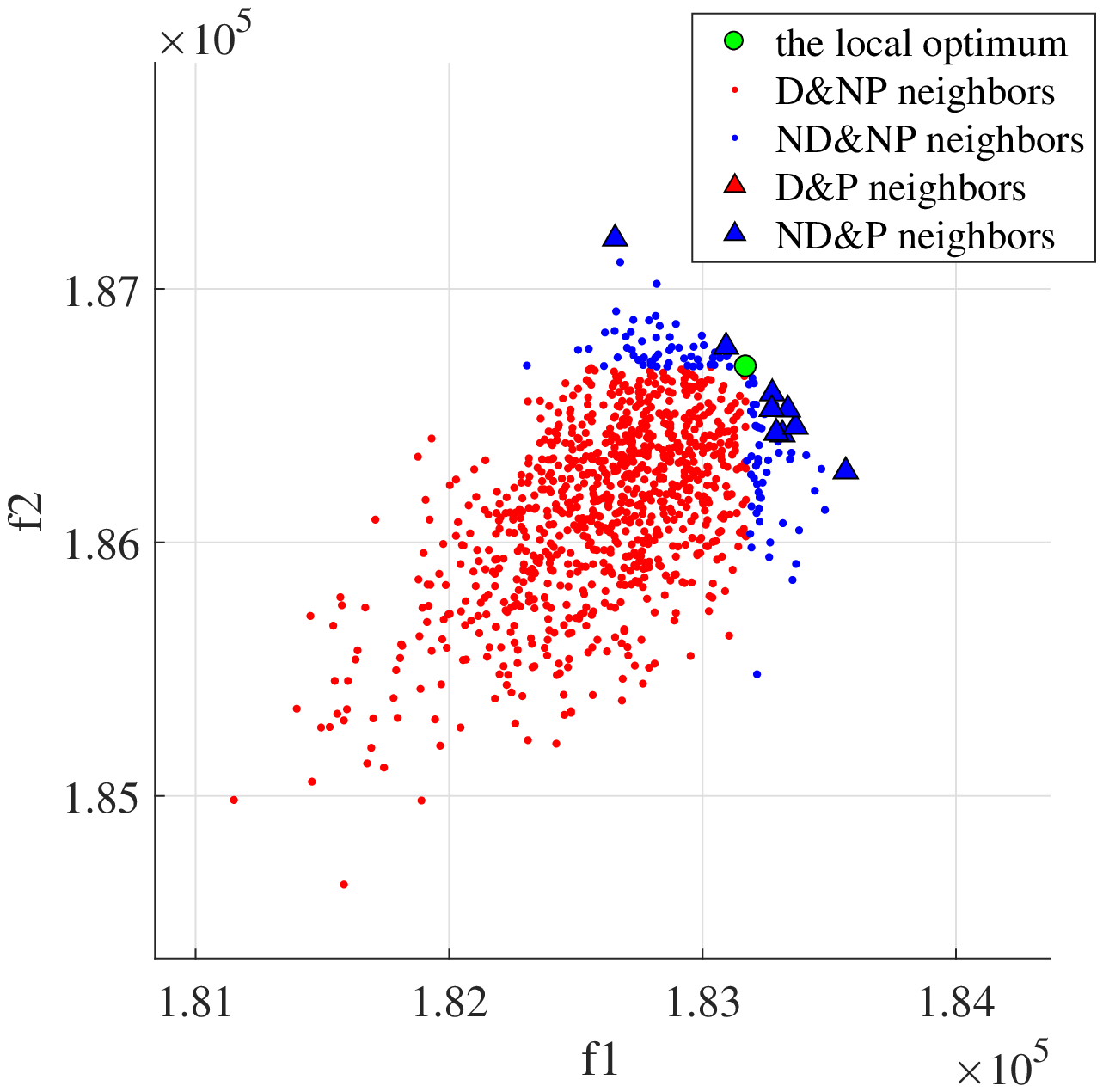}}
    \hspace{-0.005\linewidth}
  \subfigure[$\rho=0.9231$]{
    \label{fig:nei_decomp_example_bqp1000_1_8}
    \includegraphics[width=0.23\linewidth]{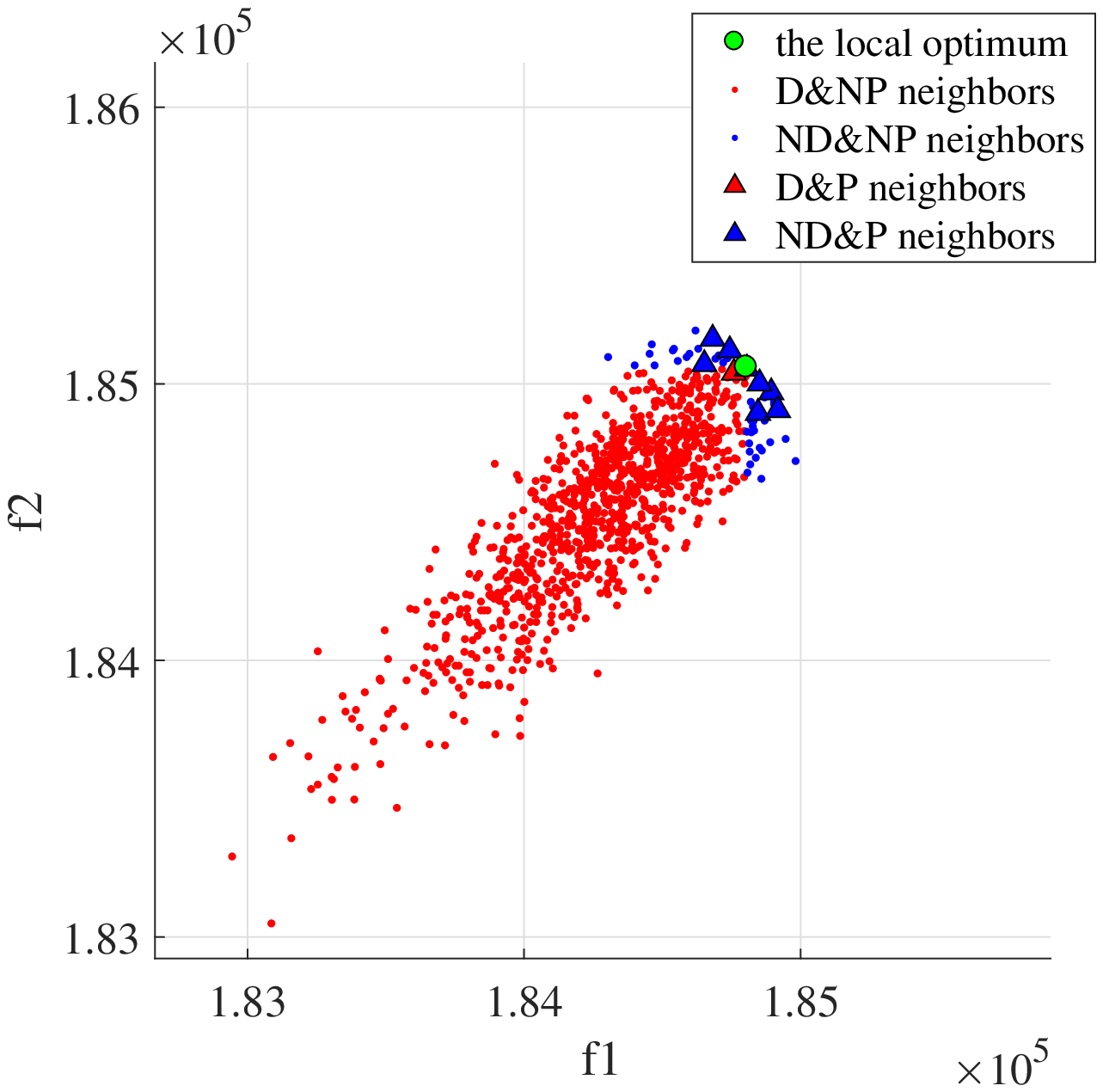}}\\
 \caption{The layout of different types of neighboring solutions of a 1-bit-flip local optimum of the UBQP instance bqp1000.1. The neighboring solutions are plotted on 8 sub-objective pairs with different correlation coefficients.}\label{fig:nei_decomp_example_bqp1000_1}
\end{figure}

To better illustrate the results of the neighborhood investigate, Figure~\ref{fig:nei_decomp_example_eil51} and Figure~\ref{fig:nei_decomp_example_bqp1000_1} show all the neighboring solutions of example local optima in the TSP instance eil51 and in the UBQP instance bqp1000.1 respectively. In Figure~\ref{fig:nei_decomp_example_eil51} and Figure~\ref{fig:nei_decomp_example_bqp1000_1}, eight pairs of sub-objectives with different correlation coefficients are shown. Note here that the UBQP is a maximization problem hence the dominance definition in the UBQP is opposite to that in the TSP. In Figure~\ref{fig:nei_decomp_example_eil51} and Figure~\ref{fig:nei_decomp_example_bqp1000_1} the local optimum is marked by green dots, the dominated neighboring solutions are in red color while the non-dominated ones are in blue color; the promising neighboring solutions are marked by triangles while the non-promising ones are marked by dots. Hence, in Figure~\ref{fig:nei_decomp_example_eil51} and Figure~\ref{fig:nei_decomp_example_bqp1000_1} a blue triangle means a solution is both non-dominated and promising. From Figure~\ref{fig:nei_decomp_example_eil51} and  Figure~\ref{fig:nei_decomp_example_bqp1000_1} we can see that most of the promising neighboring solutions are non-dominated to the original local optimum.

Based on the neighborhood exploration results in Table~\ref{tbl:nei_type_TSP} and Table~\ref{tbl:nei_type_UBQP}, we can estimate the expected number of function evaluations until finding a promising neighboring solution (i.e., improving the current local optimum) in NDS, which can be calculated by
\begin{equation}\label{eq:EFE_NDS}
  E(\mbox{\#FE}_{NDS}) = \frac{\mbox{ND}\times N+\mbox{D}}{\mbox{P\&ND}},
\end{equation}
where $N$ is the neighborhood size. Eq.~\ref{eq:EFE_NDS} can also be expressed as
\begin{equation}\label{eq:EFE_NDS_2}
  E(\mbox{\#FE}_{NDS}) = \frac{N+\frac{\mbox{D}}{\mbox{ND}}}{\frac{\mbox{P\&ND}}{\mbox{ND}}}.
\end{equation}
From Eq.~\ref{eq:EFE_NDS_2} we can deduce that a too large correlation coefficient $\rho$ is unwanted, because the proportion of non-dominated neighboring solutions ``ND'' decreases with the increasing of $\rho$ and when $\mbox{ND}\to0$, $\frac{\mbox{D}}{\mbox{ND}}\to+\infty$. In this case, even if all the non-dominated neighboring solutions are promising neighboring solutions, i.e., $\frac{\mbox{P\&ND}}{\mbox{ND}}=1$, the expected function evaluation number will be very large.

Without the filtering of NDS, finding a promising neighboring solution requires searching the neighborhood's neighborhood exhaustively. Hence, without NDS, the expected number of function evaluations until finding a promising neighboring solution is
\begin{equation}\label{eq:EFE_ENS}
  E(\mbox{\#FE})=\frac{N^2}{\mbox{P}}.
\end{equation}

\begin{figure}
  \subfigure[TSP: eil51]{
    \label{fig:EFE_eil51}
    \includegraphics[height=0.350\linewidth]{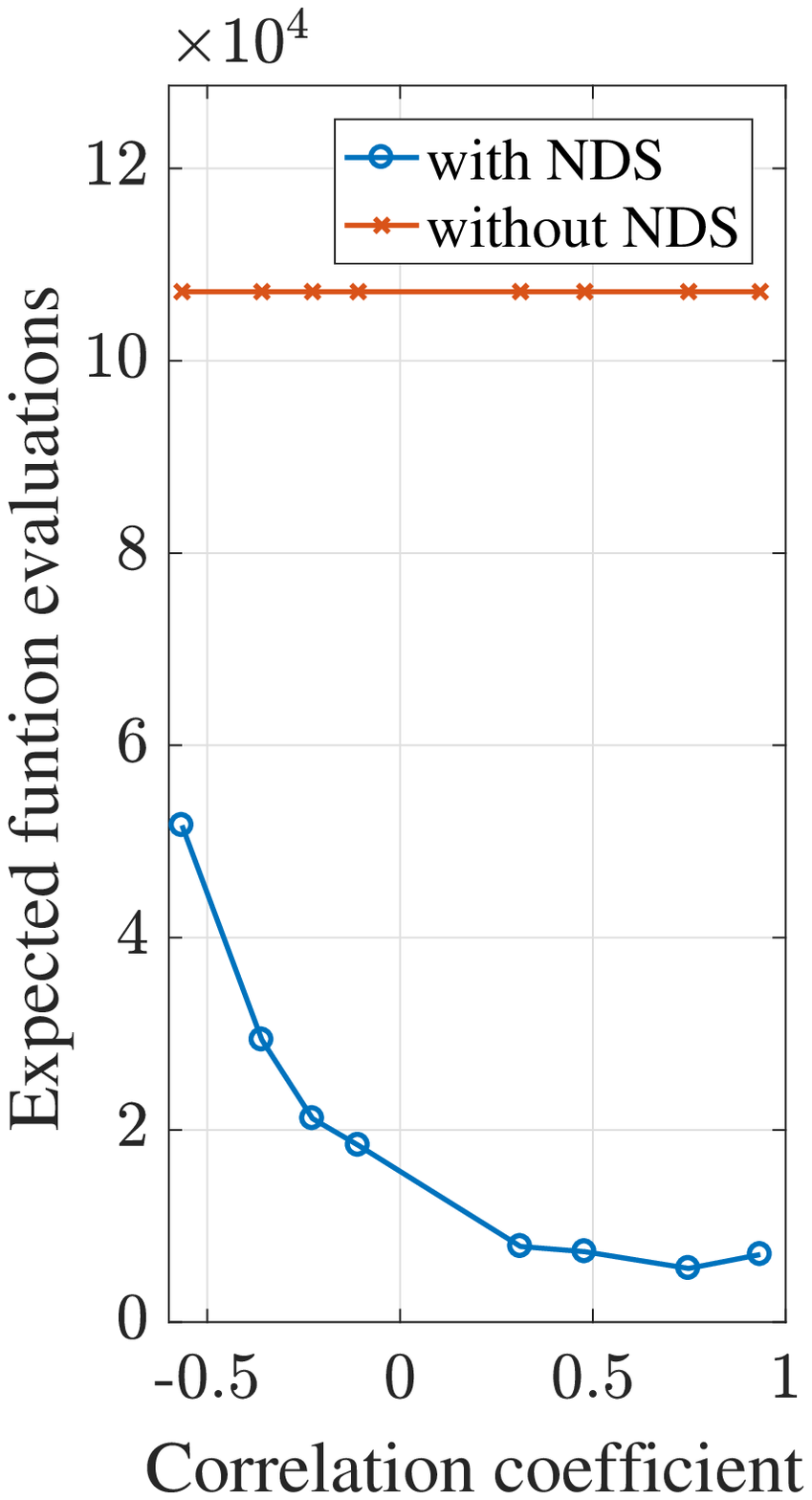}}
    \hspace{-0.01\linewidth}
  \subfigure[st70]{
    \label{fig:EFE_st70}
    \includegraphics[height=0.350\linewidth]{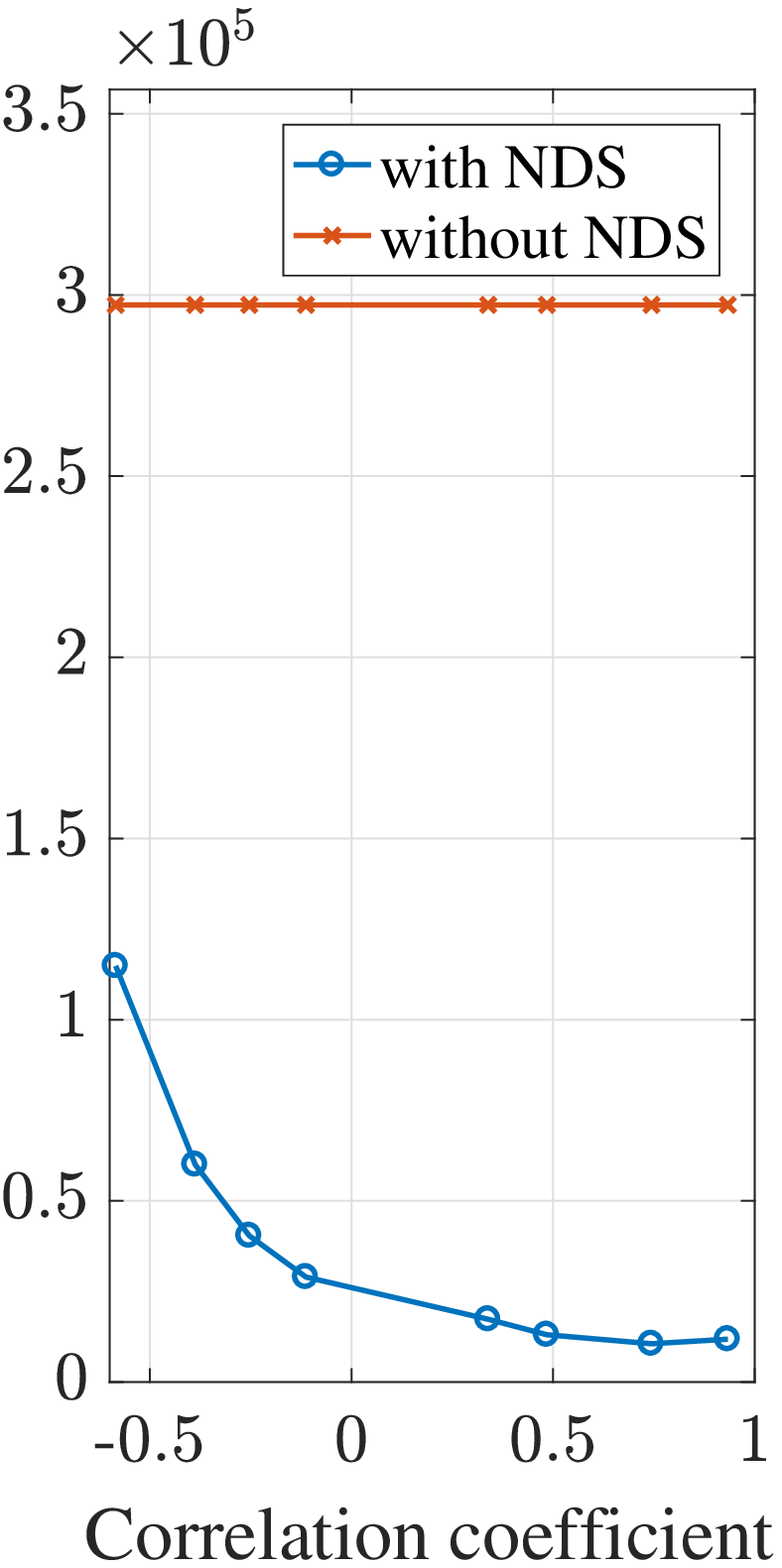}}
    \hspace{-0.01\linewidth}
  \subfigure[pr76]{
    \label{fig:EFE_pr76}
    \includegraphics[height=0.350\linewidth]{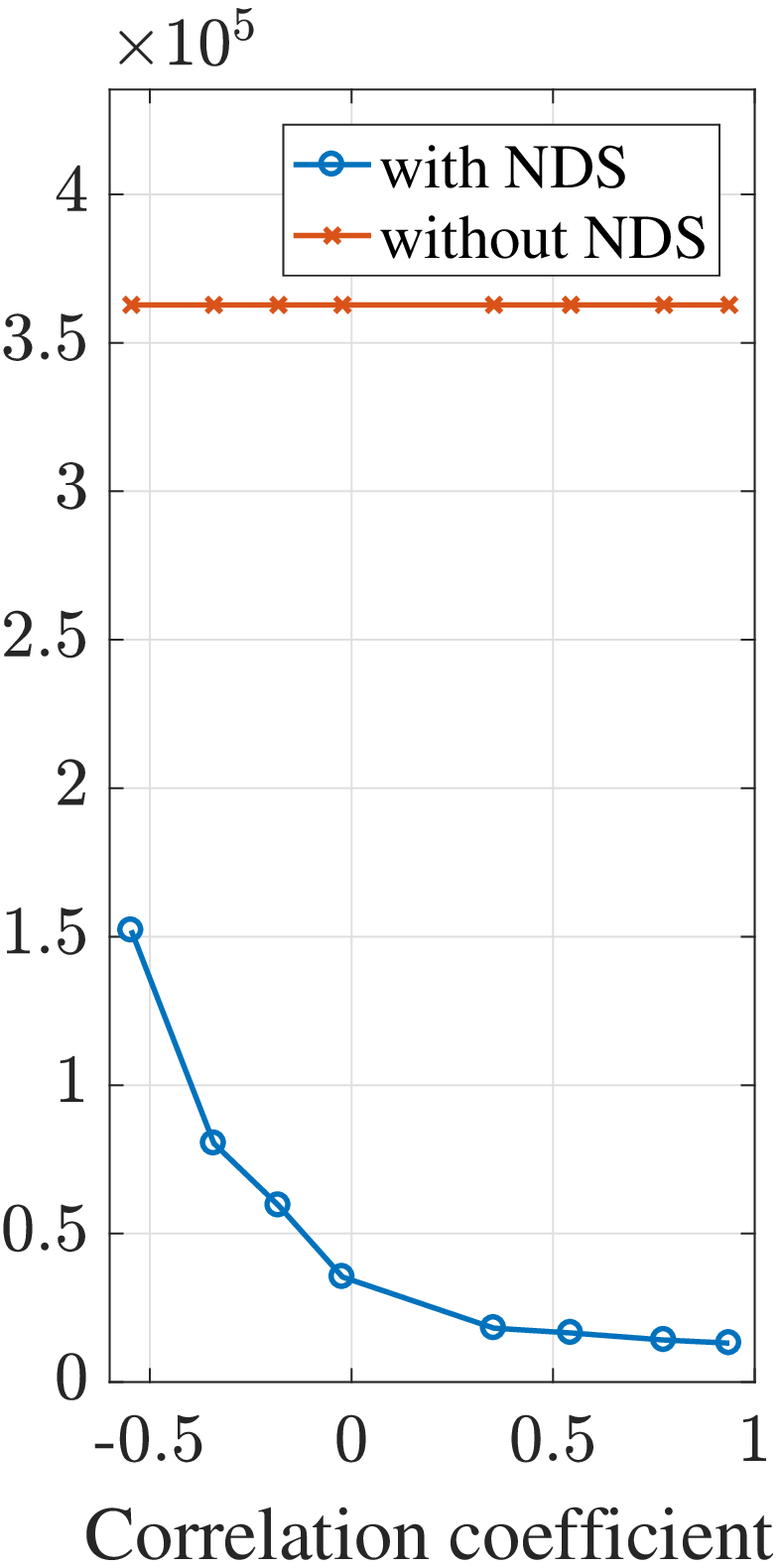}}
    \hspace{-0.01\linewidth}
  \subfigure[rat99]{
    \label{fig:EFE_rat99}
    \includegraphics[height=0.350\linewidth]{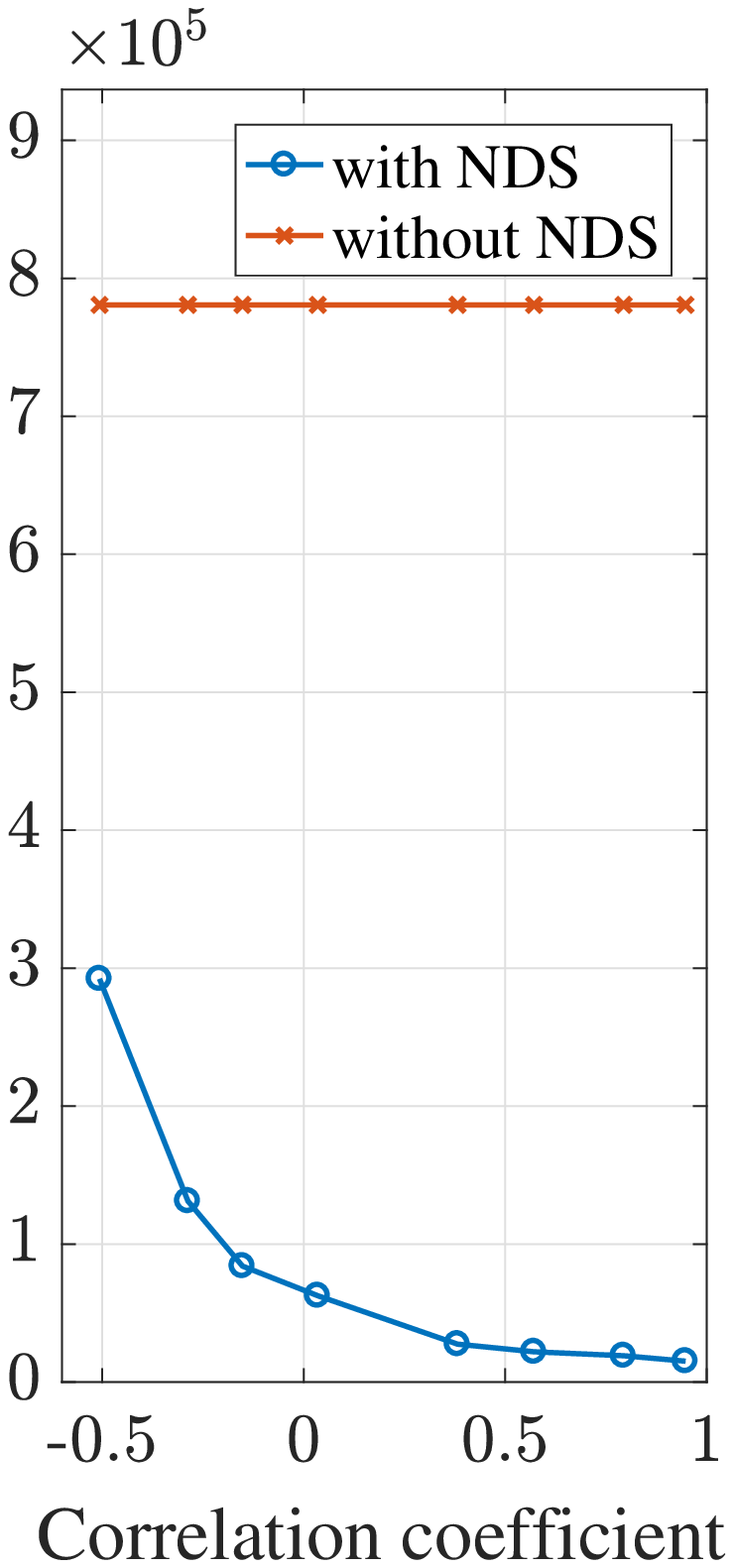}}
    \hspace{-0.01\linewidth}
  \subfigure[rd100]{
    \label{fig:EFE_rd100}
    \includegraphics[height=0.350\linewidth]{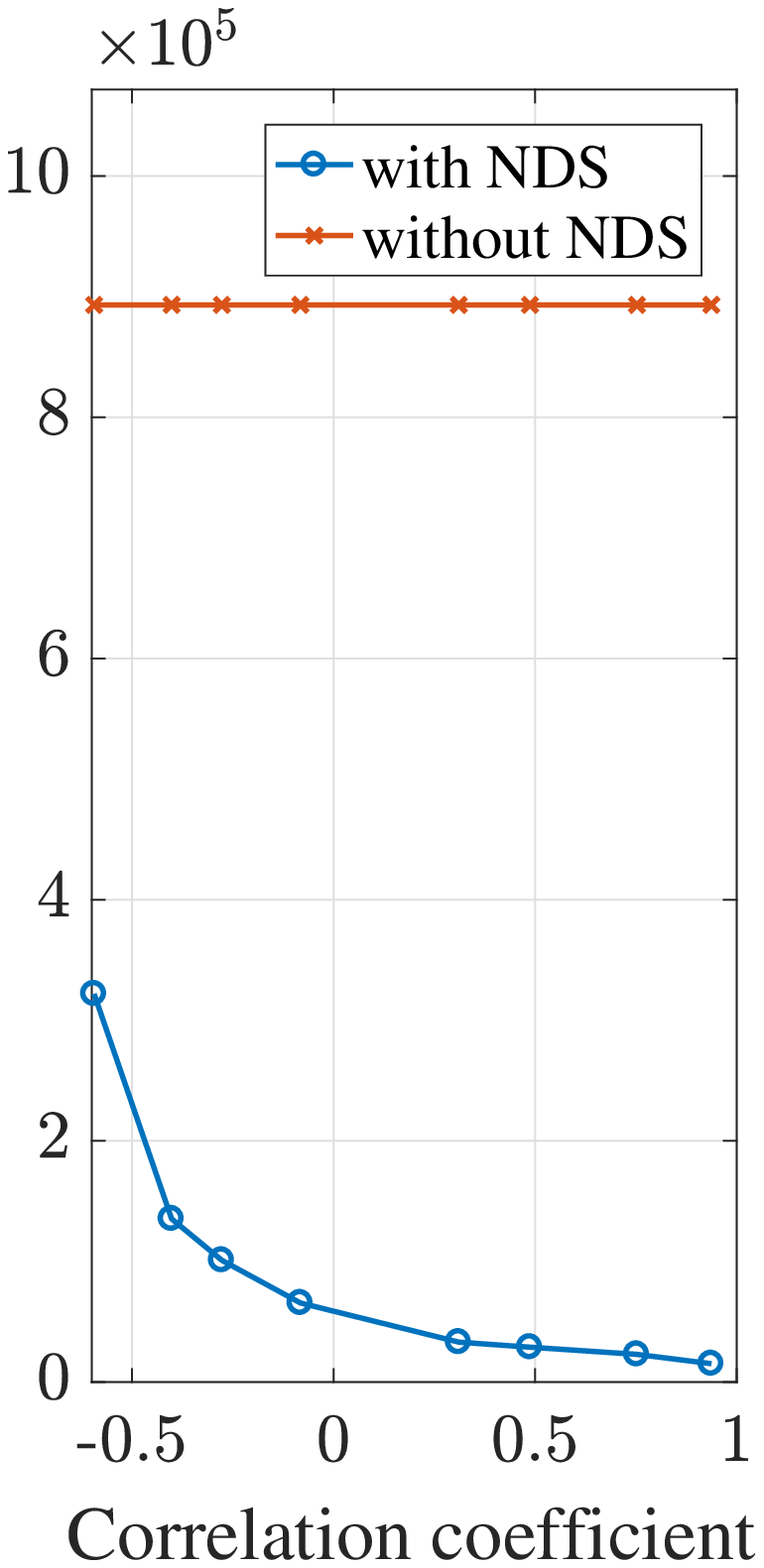}}\\
  \subfigure[\tiny{UBQP:bqp1000.1}]{
    \label{fig:EFE_bqp1000_1}
    \includegraphics[height=0.350\linewidth]{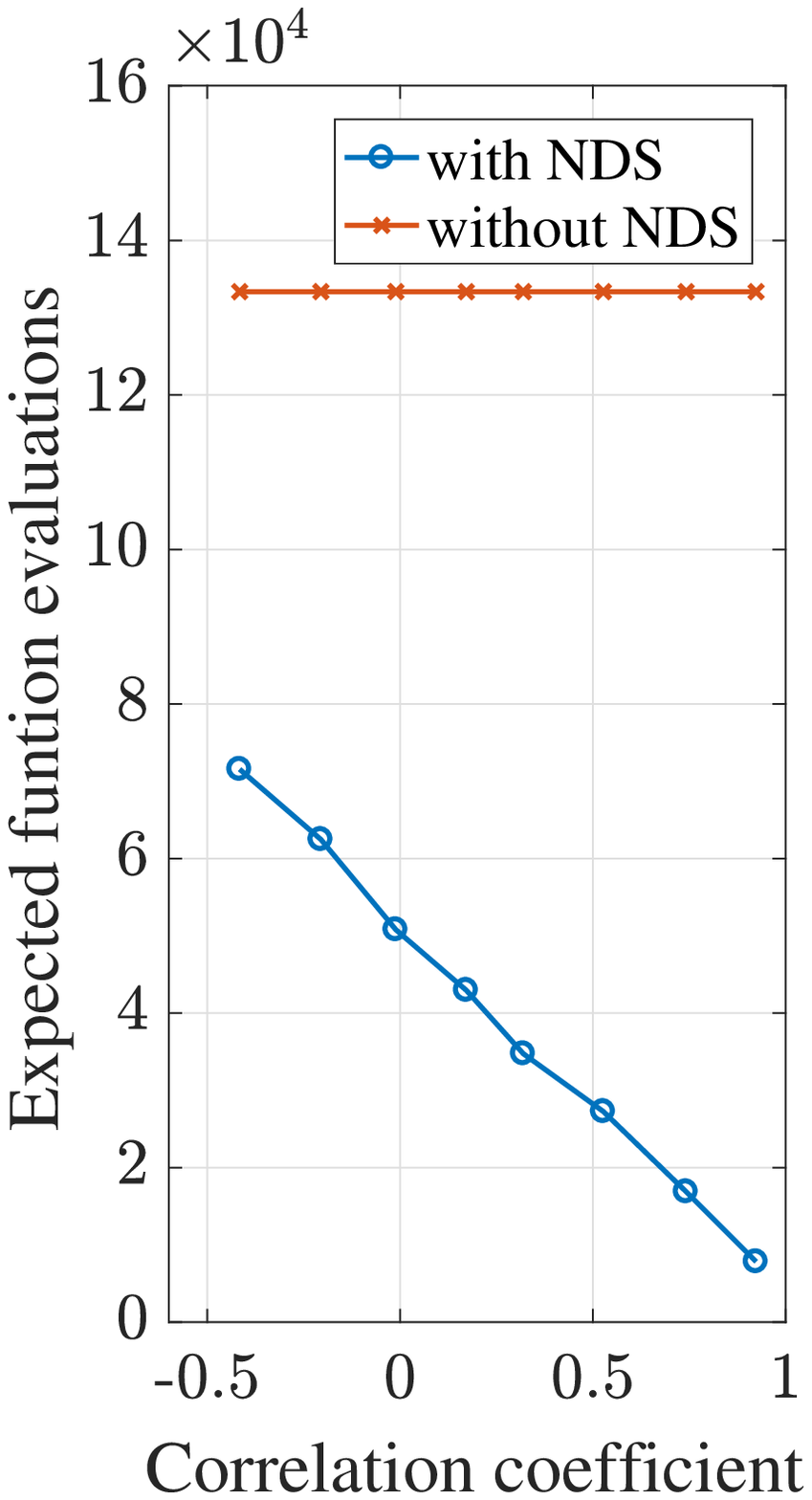}}
    \hspace{-0.01\linewidth}
  \subfigure[\tiny{bqp2500.1}]{
    \label{fig:EFE_bqp2500_1}
    \includegraphics[height=0.350\linewidth]{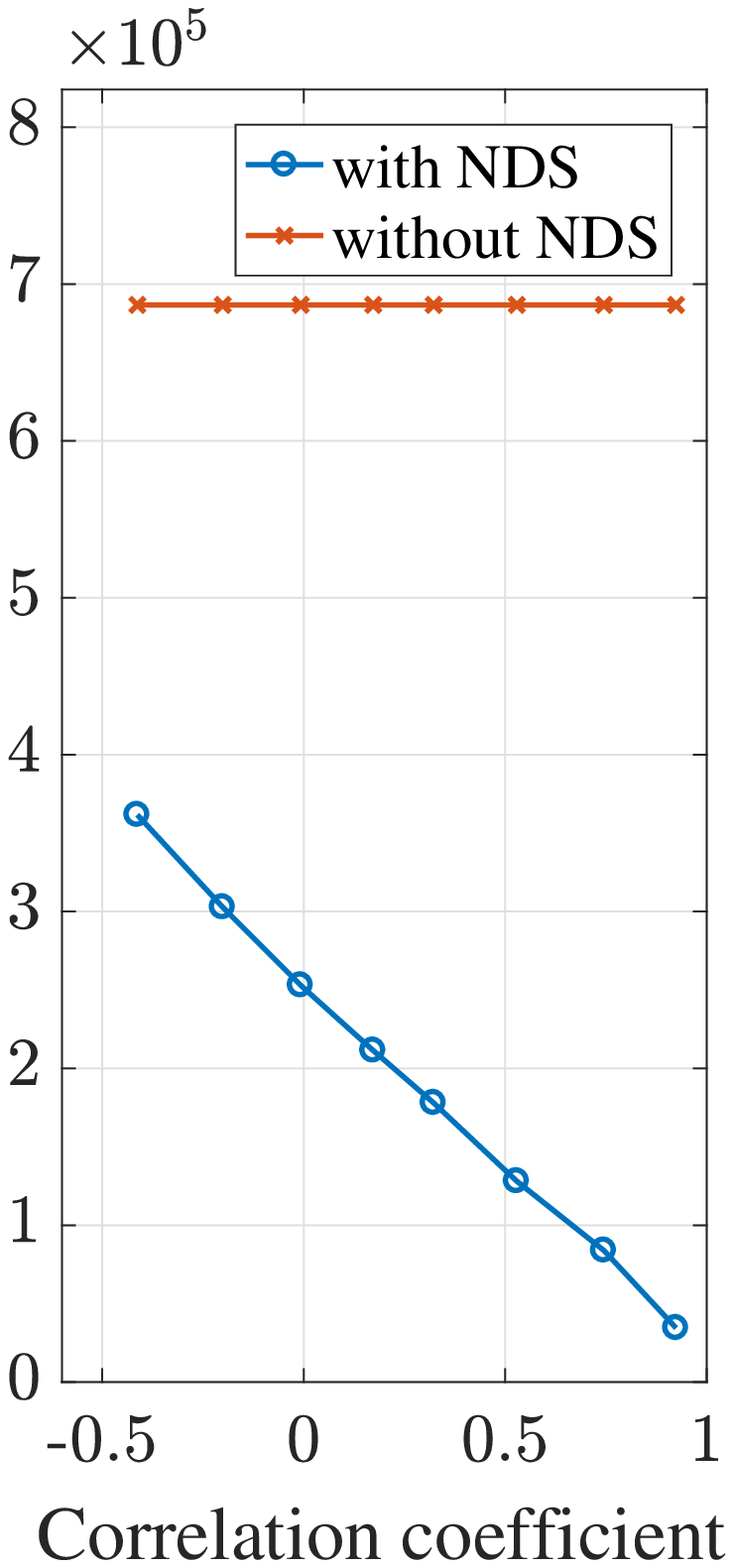}}
    \hspace{-0.01\linewidth}
  \subfigure[\tiny{p3000.1}]{
    \label{fig:EFE_p3000_1}
    \includegraphics[height=0.350\linewidth]{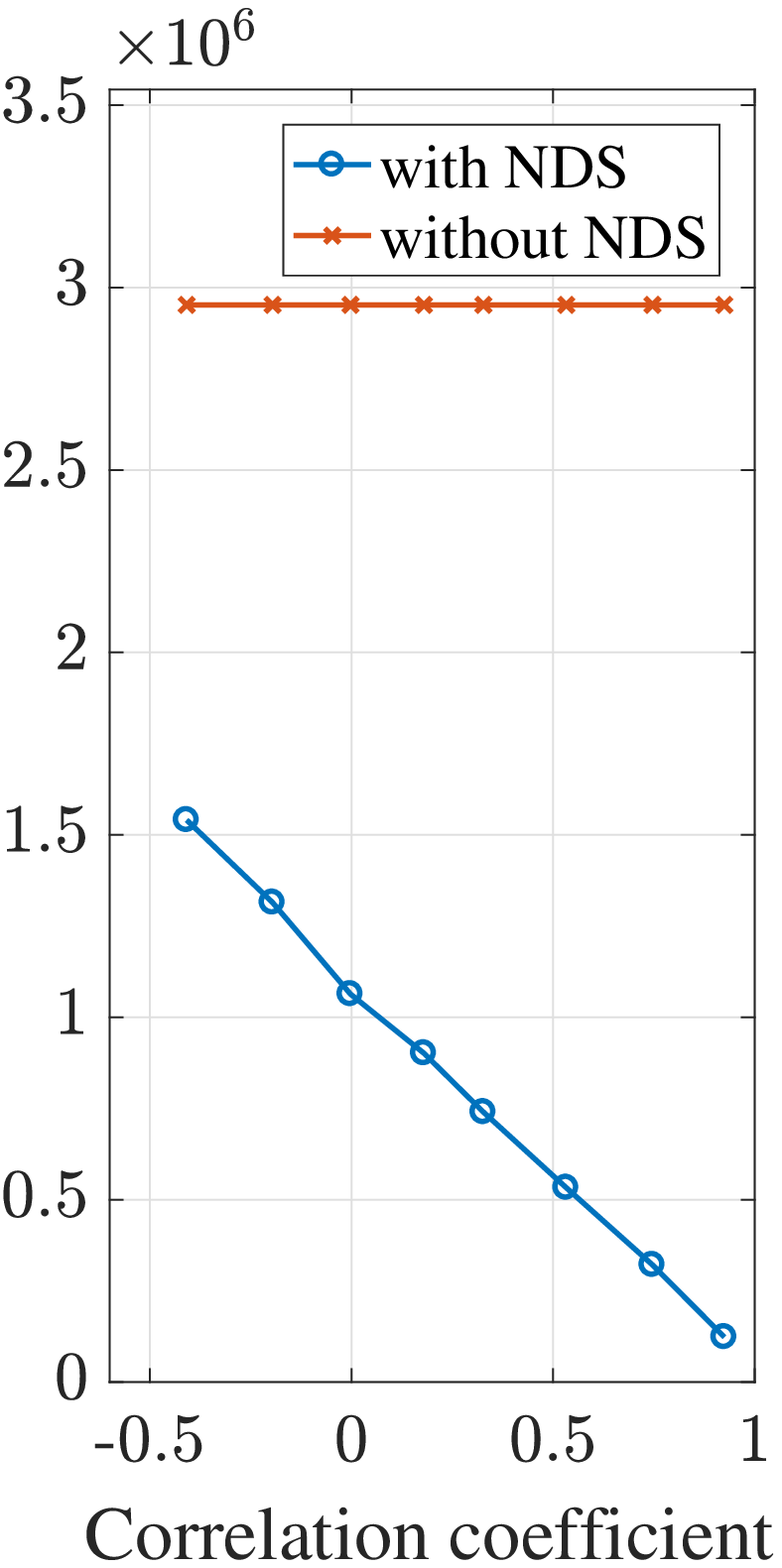}}
    \hspace{-0.01\linewidth}
  \subfigure[\tiny{p4000.1}]{
    \label{fig:EFE_p4000_1}
    \includegraphics[height=0.350\linewidth]{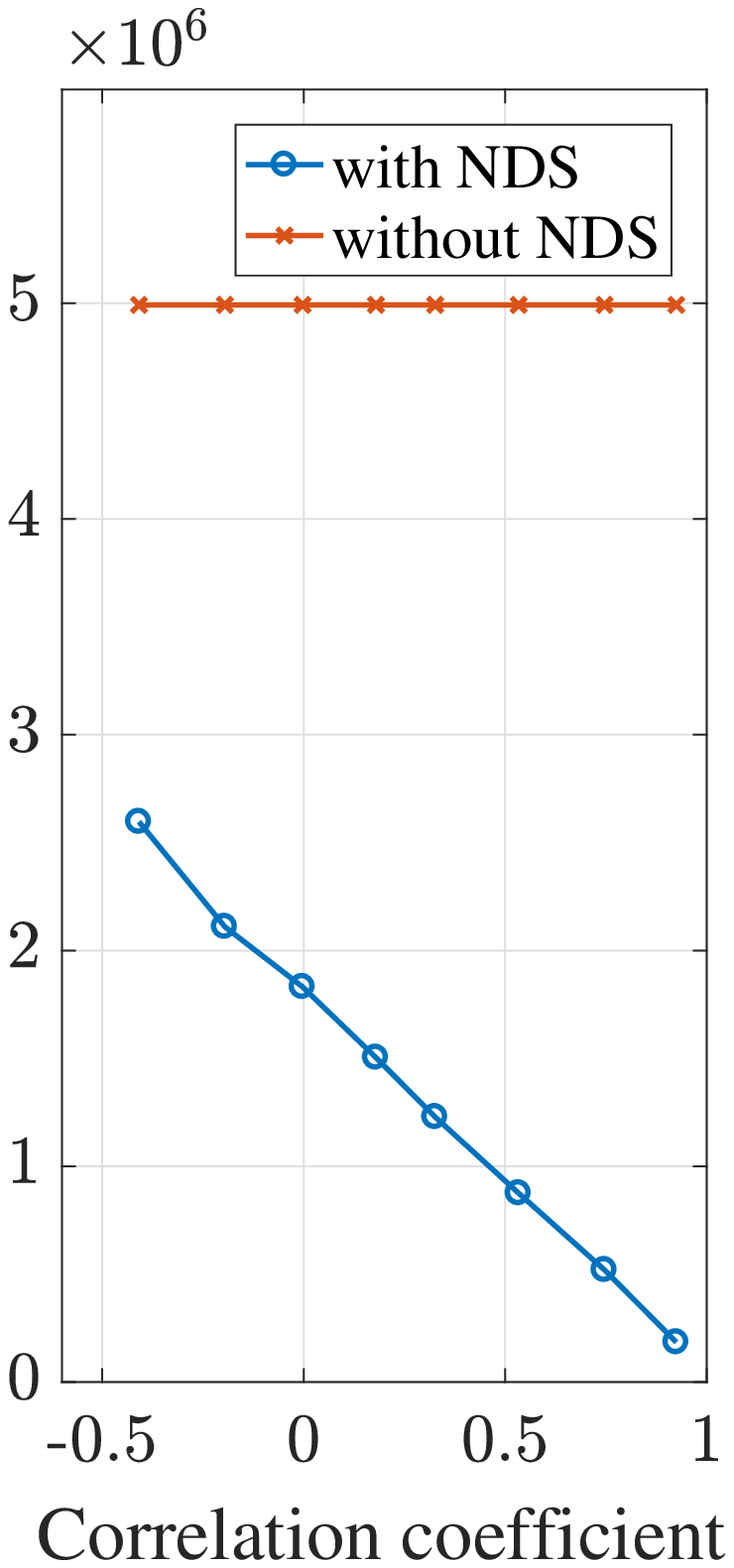}}
    \hspace{-0.01\linewidth}
  \subfigure[\tiny{p5000.1}]{
    \label{fig:EFE_p5000_1}
    \includegraphics[height=0.350\linewidth]{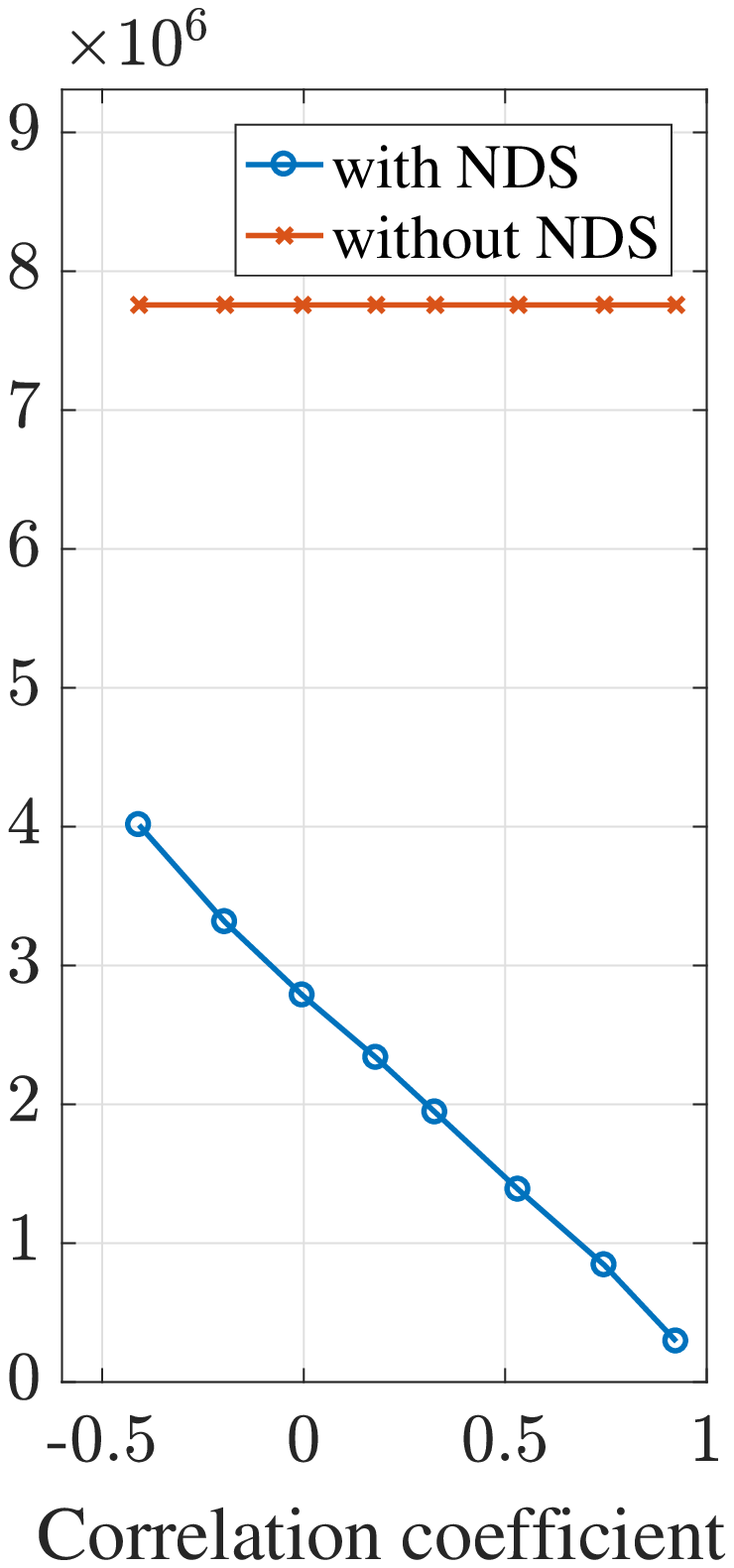}}\\
 \caption{The expected number of function evaluations until finding a promising neighboring solution}\label{fig:EFE}
\end{figure}

Figure~\ref{fig:EFE} shows the estimation of expected number of function evaluations until finding a promising neighboring solution with and without NDS on the \text{10,000} local optima. From Figure~\ref{fig:EFE} we can see that the expected number of function evaluations with NDS is significantly lower than that without NDS, which implies that NDS may be a desirable method to help an algorithm escape from local optima. In the previous analysis, we state that a too large value of $\rho$ will cause the expected function evaluation number increases, however, in Figure~\ref{fig:EFE}, on most instances the inflection point does not appear. The reason is that the neighborhood size $N$ is significantly larger than $\frac{\mbox{D}}{\mbox{ND}}$ in most settings of $\rho$, as shown in Table~\ref{tbl:nei_type_TSP} and Table~\ref{tbl:nei_type_UBQP}. For example, on the UBQP instance p5000.1, the 1-bit-flip neighborhood size is $5000$ and when $\rho=0.9247$, $\frac{\mbox{D}}{\mbox{ND}} = 96.46\% / 3.54\% =  27.2486$. Based on Eq.~\ref{eq:EFE_NDS_2}, when $N\gg\frac{\mbox{D}}{\mbox{ND}}$, $E(\mbox{\#FE}_{NDS}) \approx \frac{\mbox{$N$}}{\frac{\mbox{P\&ND}}{\mbox{ND}}}$, which means that the expected function evaluation number is mostly determined by the value of $\frac{\mbox{P\&ND}}{\mbox{ND}}$. From Table~\ref{tbl:nei_type_UBQP} we can see that on the UBQP instance p5000.1, when $\rho=0.9247$, $\frac{\mbox{P\&ND}}{\mbox{ND}}=1.72\%$, which is the largest. Hence the expected function evaluation number is the lowest when $\rho=0.9247$ on p5000.1. From Figure~\ref{fig:EFE} we can also see that, on two TSP instances, eli51 and st70, the expected function evaluation slightly increases when $\rho$ increases from about $0.7$ to about $0.9$. Considering that on these two instances the neighborhood size is not very large while the ratio $\frac{\mbox{D}}{\mbox{ND}}$ is relatively large, this phenomenon is consistent with the aforementioned explanations.

In a word, the experimental results in this sub-section support the motivation of NDS, i.e., the neighborhood of the non-dominated neighbors of a local optimum is more likely to contain a better solution.

\subsection{The Performance of ILS+NDS}\label{sec:epm_NDS}

In this subsection, we compare ILS+NDS against the original ILS, and an ILS variant called ILS with Exhaustive Neighborhood Search (ILS+ENS). ILS+ENS is similar to ILS+NDS, except that in ILS+ENS the NDS procedure is replaced by the ENS procedure. The ENS procedure can be seen as a NDS procedure without the guidance of the sub-objectives $(f_1, f_2)$, which is shown in Algorithm~\ref{alg:ENS}. By comparing ILS+NDS against ILS+ENS, we should know whether the sub-objectives can truly improve the performance of ILS.

\begin{algorithm}
\small
\SetKwInput{KwInput}{Input}
\KwInput{$x_*$, $f$}
    $x_{output} \gets x_*$\;
    \For {each $x'\in$Neighborhood($x_*$)}{
        \For {each $x''\in$Neighborhood($x'$)}{
            \If {$f(x'')<f(x_*)$}{
                $x_{output} \gets x''$\;
                exit\; 
            }
        }
    }
    \KwRet{$x_{output}$}\label{lin:output_ENS}
\caption{Exhaustive Neighborhood Search (ENS)}
\label{alg:ENS}
\end{algorithm}

The test instances in Table~\ref{tbl:inst} are also used as benchmark. For the TSP instances, the 2-Opt neighborhood and the double bridge perturbation are applied in the implementations of ILS, ILS+ENS and ILS+NDS. For the UBQP instances, the 1-bit-flip neighborhood and a random flip perturbation strategy are applied. In the random flip perturbation, 25\% of the total bits in the current solution are randomly selected and flipped. This perturbation strength is based on the experimental studies in \cite{glover2010diversification}. On each instance, the implementation of ILS+NDS uses the same eight sub-objective pairs in Table~\ref{tbl:nei_type_TSP} and Table~\ref{tbl:nei_type_UBQP}. Hence, on each test instance, we have ten test algorithms: ILS, ILS+ENS, ILS+NDS with $(f_1, f_2)_1$, ILS+NDS with $(f_1, f_2)_2$, $\dots$, ILS+NDS with $(f_1, f_2)_8$. Each algorithm is executed 50 times from different random initial solutions and stops when the globally optimal function value is reached or after $10^{10}$ function evaluations. The globally optimal function values of the UBQP instances are available from~\cite{glover2010diversification}. The globally optimal function values of the TSP instances are available from TSPLIB~\cite{reinelt1991tsplib}. The code is implemented in GNU C++ with O2 optimizing compilation. The computing platform has two 6-core 2.00GHz Intel Xeon E5-2620 CPUs (24 logical processors) under Ubuntu OS.

\begin{figure}
  \subfigure[TSP: eil51]{
    \label{fig:excess_eil51}
    \includegraphics[height=0.50\linewidth]{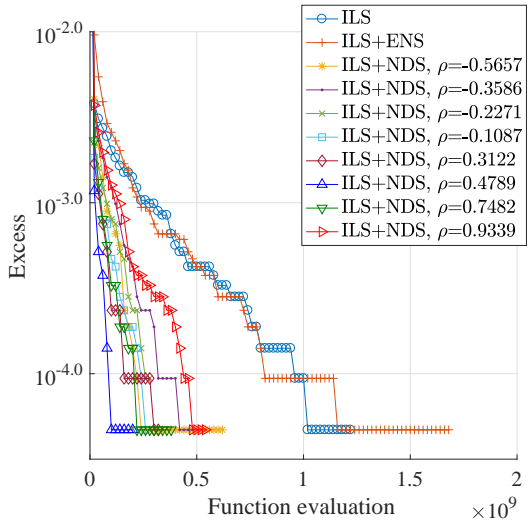}}
    \hspace{-0.01\linewidth}
  \subfigure[TSP: st70]{
    \label{fig:excess_st70}
    \includegraphics[height=0.50\linewidth]{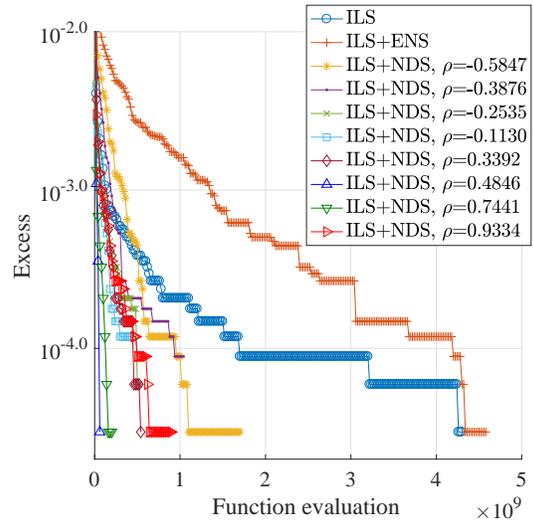}}\\
    \hspace{-0.01\linewidth}
  \subfigure[TSP: pr76]{
    \label{fig:excess_pr76}
    \includegraphics[height=0.50\linewidth]{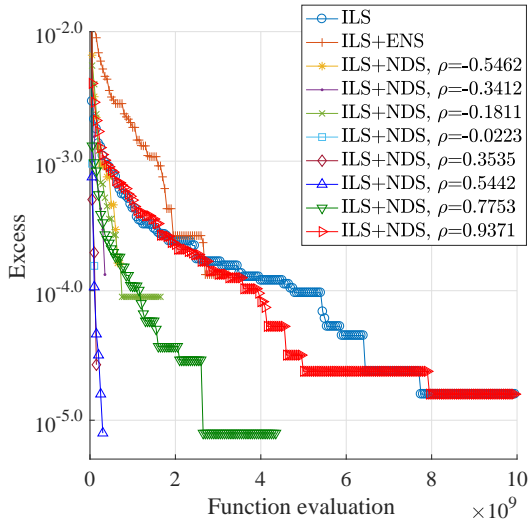}}
    \hspace{-0.01\linewidth}
  \subfigure[TSP: rat99]{
    \label{fig:excess_rat99}
    \includegraphics[height=0.50\linewidth]{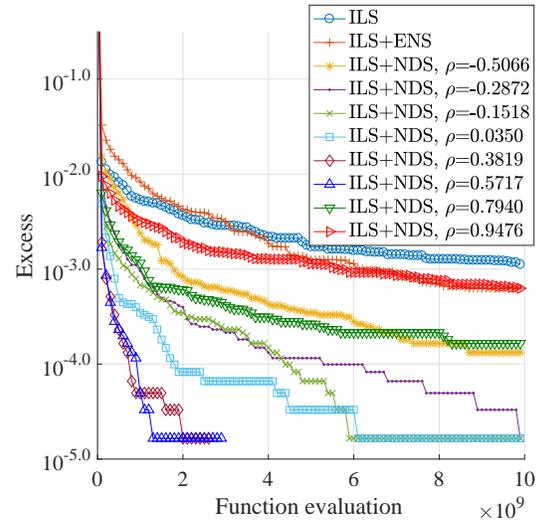}}\\
    \hspace{-0.01\linewidth}
  \subfigure[TSP: rd100]{
    \label{fig:excess_rd100}
    \includegraphics[height=0.50\linewidth]{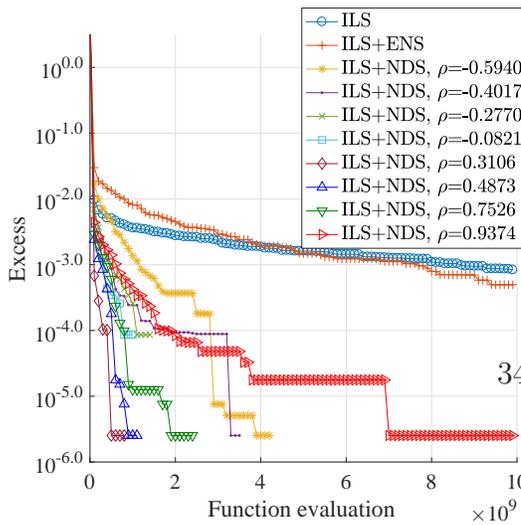}}
    \caption{Excess vs. function evaluations on 5 TSP instances}\label{fig:5TSP_excess}
\end{figure}

\begin{figure}
  \subfigure[UBQP: bqp1000.1]{
    \label{fig:excess_bqp1000_1}
    \includegraphics[height=0.50\linewidth]{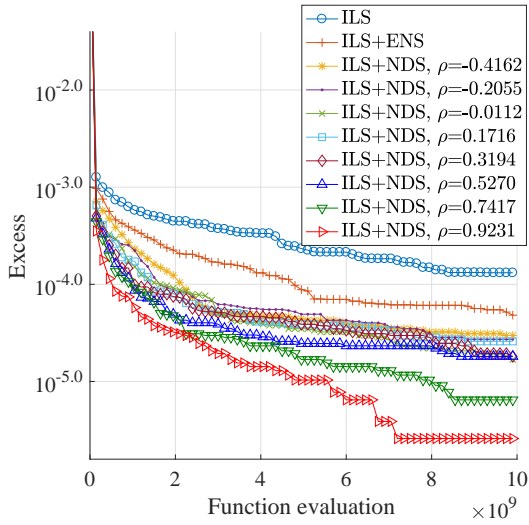}}
    \hspace{-0.01\linewidth}
  \subfigure[UBQP: bqp2500.1]{
    \label{fig:excess_bqp2500_1}
    \includegraphics[height=0.50\linewidth]{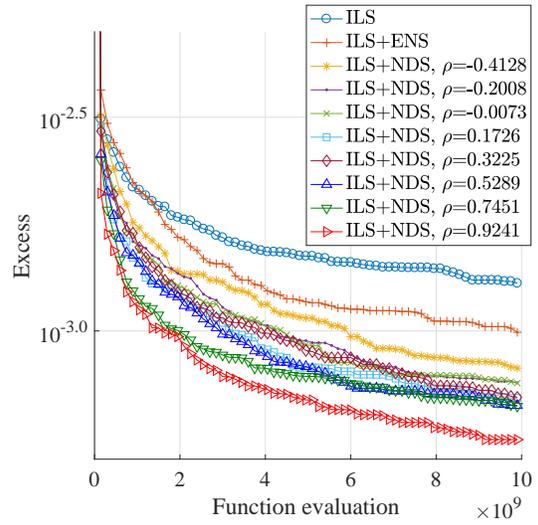}}\\
    \hspace{-0.01\linewidth}
  \subfigure[UBQP: p3000.1]{
    \label{fig:excess_p3000_1}
    \includegraphics[height=0.50\linewidth]{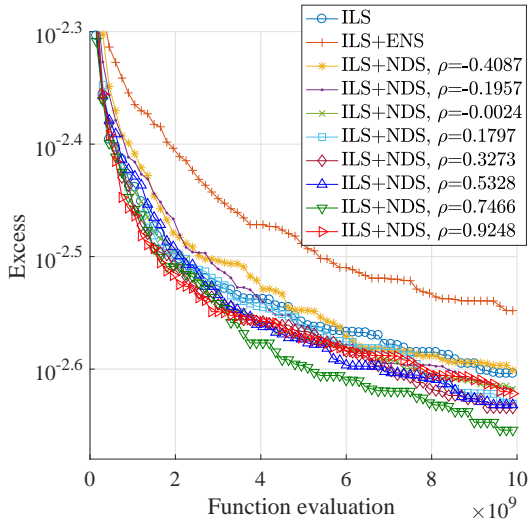}}
    \hspace{-0.01\linewidth}
  \subfigure[UBQP: p4000.1]{
    \label{fig:excess_p4000_1}
    \includegraphics[height=0.50\linewidth]{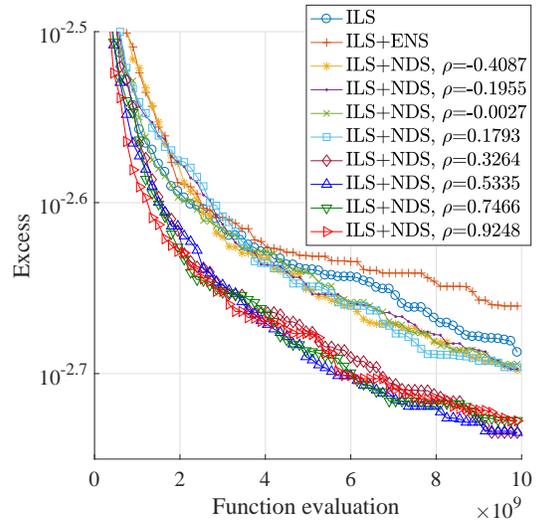}}\\
    \hspace{-0.01\linewidth}
  \subfigure[UBQP: p5000.1]{
    \label{fig:excess_p5000_1}
    \includegraphics[height=0.50\linewidth]{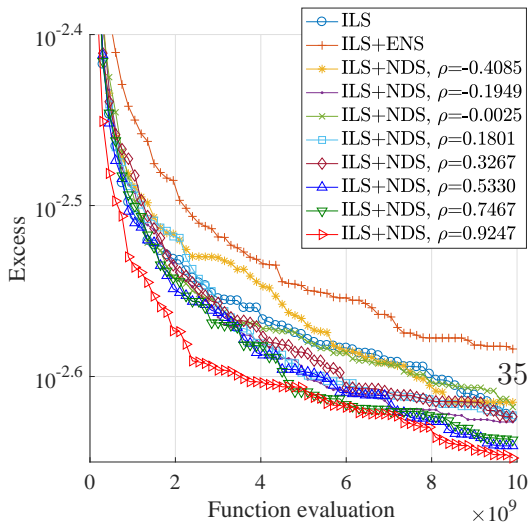}}
 \caption{Excess vs. function evaluations on 5 UBQP instances}\label{fig:5UBQP_excess}
\end{figure}

To measure the quality of the solutions found by different algorithms, we use the metric \emph{excess} which is defined by
\begin{equation}\label{eq:excess}
    \mbox{excess}(x) = \frac{|f(x) - f(x_{opt})|}{f(x_{opt})},
\end{equation}
where $x_{opt}$ is the global optimum. The lower the excess, the better. Figure~\ref{fig:5TSP_excess} and Figure~\ref{fig:5UBQP_excess} show the mean excess achieved by the compared algorithms against time on the 5 TSP instances and 5 UBQP instances respectively. In the figures, the mean excess curves are in a base 10 logarithmic scale. If a curve terminates before the final time it means that all the runs have found the global optimum before the final time. Table~\ref{tbl:ILS_final_excess_1} list the mean value and standard deviation of the final excess achieved by the ILS, ILS+ENS and ILS+NDS, in which the excess result of ILS+NDS is the best excess selected from the eight settings of $\rho$. The last two columns in Table~\ref{tbl:ILS_final_excess_1} list the best $\rho$ values of ILS+NDS and the corresponding $a$ values (if there are multiple best $\rho$ values, the first one is listed). In the ILS column and ILS+ENS column, ``$-$'' means the final excess achieved by ILS and ILS+ENS is significantly worse than that achieved by ILS+NDS based on a Mann-Whitney U-test at the 0.05 significance level and ``$=$'' means the performance is the same.

From Figure~\ref{fig:5TSP_excess} and Figure~\ref{fig:5UBQP_excess} we can see that on all the test instances, the best performance is achieved by ILS+NDS. In addition, at most setups of $\rho$, ILS+NDS performs better than ILS and ILS+ENS. On some instances (e.g. UBQP instance bqp1000.1), ILS+ENS performs better than ILS. However, on most instances, ILS+ENS performs worse than ILS. This indicates that without the guidance of the sub-objectives, the search efficiency of ILS+ENS decreases significantly. From Table~\ref{tbl:ILS_final_excess_1} we can see that the results of the Mann-Whitney U-test suggest that ILS+NDS performs significantly better than ILS and ILS+ENS on most instances. In the end ILS+NDS successfully finds the global optima of all TSP instances (i.e. the mean final excess is 0), while ILS only finds the global optima of two TSP instances and ILS+ENS finds the global optima of three TSP instances.

By comparing different ILS+NDS setups, we can observe that the sub-objective correlation significantly influences the performance of ILS+NDS. In the pervious neighborhood exploration experiment in Section~\ref{sec:nei_explore}, we observed that the higher the correlation, the lower the time taken to find a promising neighboring solution. However, in this experiment, it shows that a too high sub-objective correlation can reduce the efficiency of the algorithm. For example, on the TSP instance rd100, the results on Section~\ref{sec:nei_explore} show that when $\rho = 0.9374$, the expected function evaluation number is the lowest. However, here the ILS+NDS implementation with $\rho = 0.9374$ performs significantly worse than the ILS+NDS implementation with $\rho = 0.3106$. A possible explanation is that, the neighborhood exploration experiment in Section~\ref{sec:nei_explore} only measures the average properties of the \text{10,000} randomly collected local optima. It does not consider the quality of local optimum. During the ILS, the goal is to improve the current best local optimum. Many other factors could influence its performance. For example, the quality of the current best local optimum and the hardness to improve it. Although a too high correlation is not preferred, the results show that a positive correlation coefficient is better than a negative one on most test instances.

\begin{table}
\caption{Final excess of ILS, ILS+ENS and ILS+NDS on 5 TSP instances and 5 UBQP instances.}
\centering
\label{tbl:ILS_final_excess_1}
\resizebox{\linewidth}{!}{
\begin{tabular}{l| c c c | c | c }
\hline
\multirow{2}{*}{Instance} & \multicolumn{3}{c|}{Mean final excess (standard deviation)} & \multirow{2}{*}{best $\rho$ in ILS+NDS} &\multirow{2}{*}{corresponding $a$ } \\
\cline{2-4}
 & ILS & ILS+ENS & ILS+NDS & &\\
 \hline
TSP:eil51 & 0.00000(0.00000) = & 0.00000(0.00000) = & 0.00000(0.00000) & $-$0.5657 & $-$12 \\
\hline
TSP:st70 & 0.00000(0.00000) = & 0.00000(0.00000) = & 0.00000(0.00000) & $-$0.5847 & $-$12 \\
\hline
TSP:pr76 & 0.00002(0.00008) = & 0.00000(0.00000) = & 0.00000(0.00000) & $-$0.5462 & $-$12 \\
\hline
TSP:rat99 & 0.00112(0.00118) $-$ & 0.00063(0.00098) $-$ & 0.00000(0.00000) & 0.3819 & 2 \\
\hline
TSP:rd100 & 0.00083(0.00129) $-$ & 0.00050(0.00168) $-$ & 0.00000(0.00000) & $-$0.5940 & $-$12 \\
\hline
\hline
UBQP:bqp1000.1 & 0.00013(0.00012) $-$ & 0.00005(0.00003) $-$ & 0.00000(0.00001) & 0.9231 & 10\\
\hline
UBQP:bqp2500.1 & 0.00129(0.00026) $-$ & 0.00098(0.00025) $-$ & 0.00055(0.00019) & 0.9241 & 10\\
\hline
UBQP:p3000.1 & 0.00246(0.00051) $-$ & 0.00281(0.00061) $-$ & 0.00221(0.00057) & 0.7466 & 4\\
\hline
UBQP:p4000.1 & 0.00205(0.00031) $-$ & 0.00218(0.00041) $-$ & 0.00184(0.00038) & 0.5335 & 2\\
\hline
UBQP:p5000.1 & 0.00238(0.00031) = & 0.00261(0.00041) $-$ & 0.00225(0.00039) & 0.9247 & 10\\
\hline
\multicolumn{6}{c}{\begin{minipage}{620pt}*In the last two columns, if there are multiple best $\rho$ values, the first one is listed. ``$-$'' means the final excess achieved by ILS and ILS+ENS is significantly worse than that achieved by ILS+NDS based on a Mann-Whitney U-test at the 0.05 significance level and ``$=$'' means the performance is the same.\end{minipage}}
\end{tabular}

}
\end{table}

Beside the previous experiment, we conduct an additional comparison experiment on 12 TSP instances and 10 UBQP instances. In this experiment we also compare ILS, ILS+ENS and ILS+NDE. The difference to the previous experiment is that here we only test one setting of $\rho$ in the ILS+NDS implementation on each instance.  Table~\ref{tbl:ILS_final_excess_2} lists the mean value and standard deviation of the final excess achieved by the ILS, ILS+ENS and ILS+NDS. The last two columns in Table~\ref{tbl:ILS_final_excess_2} list the $\rho$ values in ILS+NDS and the corresponding $a$ values. In the ILS column and ILS+ENS column, ``$-$'' means the final excess achieved by ILS and ILS+ENS is significantly worse than that achieved by ILS+NDS based on a Mann-Whitney U-test at the 0.05 significance level and ``$=$'' means the performance is the same. From Table~\ref{tbl:ILS_final_excess_1} we can see that the results of the Mann-Whitney U-test suggest that ILS+NDS performs significantly better than ILS and ILS+ENS on most instances.

Based the above experimental study, we can conclude that the proposed NDS method can truly improve the original version of ILS.

\begin{table}
\caption{Final excess of ILS, ILS+ENS and ILS+NDS on 12 TSP instances and 10 UBQP instances.}
\centering
\label{tbl:ILS_final_excess_2}
\resizebox{\linewidth}{!}{
\begin{tabular}{l| c c c | c | c }
\hline
\multirow{2}{*}{Instance} & \multicolumn{3}{c|}{Mean final excess (standard deviation)} & \multirow{2}{*}{$\rho$ in ILS+NDS} & \multirow{2}{*}{corresponding $a$}\\
\cline{2-4}
 & ILS & ILS+ENS & ILS+NDS & & \\
 \hline
TSP:eil101 & 0.00712(0.00237) $-$ & 0.00474(0.00368) $-$ & 0.00064(0.00079) & 0.4864 & 3\\
\hline
TSP:pr107 & 0.00189(0.00115) $-$ & 0.00150(0.00147) $-$ & 0.00000(0.00000) & 0.5948 & 3 \\
\hline
TSP:bier127 & 0.00271(0.00131) $-$ & 0.00339(0.00353) $-$ & 0.00003(0.00008) & 0.6112 & 3 \\
\hline
TSP:ch130 & 0.00569(0.00146) $-$ & 0.00478(0.00337) $-$ & 0.00014(0.00058) & 0.4922 & 3 \\
\hline
TSP:kroA150 & 0.00636(0.00246) $-$ & 0.00606(0.00485) $-$ & 0.00002(0.00014) & 0.5404 & 3 \\
\hline
TSP:u159 & 0.00549(0.00305) $-$ & 0.00478(0.00593) $-$ & 0.00000(0.00000) & 0.5448 & 3 \\
\hline
TSP:rat195 & 0.02318(0.00359) $-$ & 0.02462(0.00893) $-$ & 0.00694(0.00231) & 0.5620 & 3 \\
\hline
TSP:d198 & 0.00987(0.00140) $-$ & 0.00676(0.00310) $-$ & 0.00170(0.00086) & 0.7246 & 3 \\
\hline
TSP:ts225 & 0.00522(0.00108) $-$ & 0.01237(0.01722) = & 0.00050(0.00036) & 0.4690 & 3 \\
\hline
TSP:gil262 & 0.02692(0.00413) $-$ & 0.03832(0.01049) $-$ & 0.00982(0.00246) & 0.4805 & 3 \\
\hline
TSP:a280 & 0.02418(0.00330) $-$ & 0.04927(0.01974) $-$ & 0.00389(0.00168) & 0.5213 & 3 \\
\hline
TSP:lin318 & 0.02739(0.00304) $-$ & 0.04248(0.01226) $-$ & 0.01265(0.00298) & 0.4927 & 3 \\
\hline
UBQP:p3000.2 & 0.00160(0.00028) $-$ & 0.00165(0.00035) $-$ & 0.00131(0.00033) & 0.9247 & 2 \\
\hline
UBQP:p3000.5 & 0.00153(0.00024) = & 0.00180(0.00029) $-$ & 0.00156(0.00028) & 0.9248 & 2 \\
\hline
UBQP:p4000.2 & 0.00200(0.00032) = & 0.00215(0.00029) $-$ & 0.00195(0.00028) & 0.9248 & 2 \\
\hline
UBQP:p4000.5 & 0.00244(0.00033) $-$ & 0.00260(0.00042) $-$ & 0.00229(0.00023) & 0.9247 & 2 \\
\hline
UBQP:p5000.2 & 0.00239(0.00029) $-$ & 0.00259(0.00033) $-$ & 0.00227(0.00025) & 0.9247 & 2 \\
\hline
UBQP:p5000.5 & 0.00244(0.00035) $-$ & 0.00260(0.00030) $-$ & 0.00231(0.00028) & 0.9247 & 2 \\
\hline
UBQP:p6000.2 & 0.00318(0.00029) $-$ & 0.00325(0.00039) $-$ & 0.00298(0.00032) & 0.9247 & 2 \\
\hline
UBQP:p6000.3 & 0.00301(0.00035) = & 0.00304(0.00035) $-$ & 0.00284(0.00039) & 0.9248 & 2 \\
\hline
UBQP:p7000.2 & 0.00286(0.00025) $-$ & 0.00305(0.00026) $-$ & 0.00277(0.00024) & 0.9247 & 2 \\
\hline
UBQP:p7000.3 & 0.00338(0.00032) = & 0.00348(0.00040) $-$ & 0.00331(0.00030) & 0.9248 & 2 \\
\hline
\multicolumn{6}{c}{\begin{minipage}{600pt}*``$-$'' means the final excess achieved by ILS and ILS+ENS is significantly worse than that achieved by ILS+NDS based on a Mann-Whitney U-test at the 0.05 significance level and ``$=$'' means the performance is the same.\end{minipage}}
\end{tabular}
}
\end{table}

\subsection{The Performance of ILK+NDE}

In this section, we test the performance of ILK+NDE on six middle-size and large-size TSP instances. To verify the effect of the proposed NDE technique, we compare ILK+NDE against the original ILK and a variant of ILK+NDE in which the guidance of the sub-objectives is removed which is named as Iterated Lin-Kernighan algorithm with further Exploitation (ILK+E). ILK+E is summarized in Algorithm~\ref{alg:ILK+E}, in which we can see that ILK+E conducts further exploitation on all the encountered LK local optima. By comparing ILK+E against ILK+NDE, we can verify whether the sub-objectives $(f_1, f_2)$ can truly improve the algorithm performance.

\begin{algorithm}
\small
\SetKwInput{KwInput}{Input}
\KwInput{$f$, $T$, $k$, $\tilde{c}$}
    Decompose $f$ into $f_1$ and $f_2$\;
    $x_0' \gets $ randomly or heuristically generated solution\;
    $x_0 \gets$ LK($x_0'\mid f$)\;
    $x_{best} \gets x_0$\;
    $j \gets 0$\;
    \While{stopping criterion is not met}{
        $x_{j+1} \gets$ FurtherExploit($x_j\mid T,k,\tilde{c}$)\;
        \If {$x_{j+1} = x_j$}{
            $x_{j+1}' \gets$ Perturbation($x_j$)\;
            $x_{j+1} \gets$ LK($x_{j+1}'\mid f$)\;
        }
        \If {$f(x_{j+1}) < f(x_{best})$} {
            $x_{best} \gets x_{j+1}$\;
        }
        $j\gets j+1$\;
    }
    \KwRet{\mbox{the historical best solution} $x_{best}$}
\caption{ILK+E}
\label{alg:ILK+E}
\end{algorithm}

In the following experiment, we compare ILK+NDE against ILK and ILK+E on ten middle-size and large-size TSP instances from the TSPLIB: \{vm1748, u1817, d2103, pr2392, pcb3038, fnl4461, pla7397, rl11849, usa13509, d18512\}. In the experiments, the implementation of the LK local search is based on the Concorde software package~\footnote{\url{http://www.math.uwaterloo.ca/tsp/concorde/}}. Following \cite{bentley1992fast}, the edge exchange in the LK implementation is restricted in a sub-graph of the original TSP graph $\cal G$. In the sub-graph, each vertex (city) only connects with its 20 nearest vertexes (cities). Following \cite{johnson1997traveling}, a double bridge kick is used in the perturbation phases of the implementations of ILK, ILK+E and ILK+NDE. For ILK+E and ILK+NDE, we set $T=1000$, $k=5$ and the penalty $\tilde{c}$ is equal to the largest edge cost in each test instance. In the ILK+NDE, first the original TSP is decomposed into two sub-objectives $(f_1,f_2)$ based on the decomposition introduced in Section~\ref{sec:nei_explore}. Since the LK local search only focuses on the edges in the nearest sub-graph of the TSP, we only decompose the edges in the sub-graph. 

With different probability distributions (see Eq.~\ref{eq:p_prime} and Eq.~\ref{eq:p}), we decompose each TSP instance into five pairs of $(f_1, f_2)$ by defining $a = -1, 0, 1, 2, 3$. The resulting $\rho$ values ranging from about $-0.5$ to about $0.6$. It is very hard to count the function evaluation number in the LK local search, hence we use the CPU runtime as the stopping criterion for the compared algorithms. The max runtime on the test instances are \{vm1748: 400s, u1817: 400s, d2103: 500s, pr2392: 500s, pcb3038: 600s, fnl4461: 900s, pla7397: 1500s, rl11849: 2400s, usa13509: 2700s, d18512: 3700s\}. In the ILK+E and ILK+NDE implementations, in the first $1/5$ runtime the ILK procedure is applied; while in the last $4/5$ runtime it is the ILK+E/ILK+NDE procedure. On each instance, each algorithm is run 50 times from different random initial solutions.

\begin{figure}
  \subfigure[vm1748]{
    \label{fig:excess_LK_vm1748}
    \includegraphics[height=0.30\linewidth]{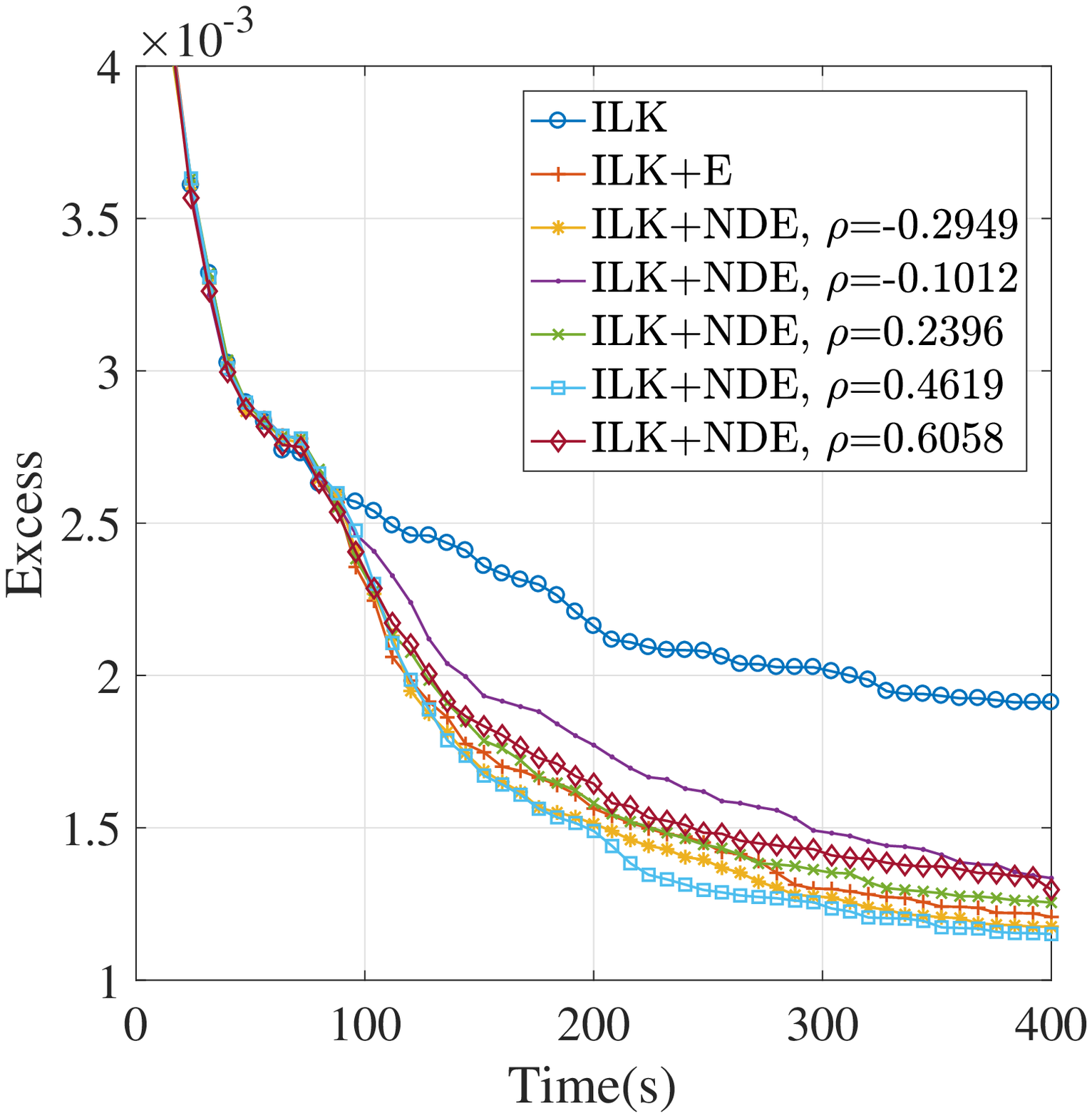}}
  \subfigure[u1817]{
    \label{fig:excess_LK_u1817}
    \includegraphics[height=0.30\linewidth]{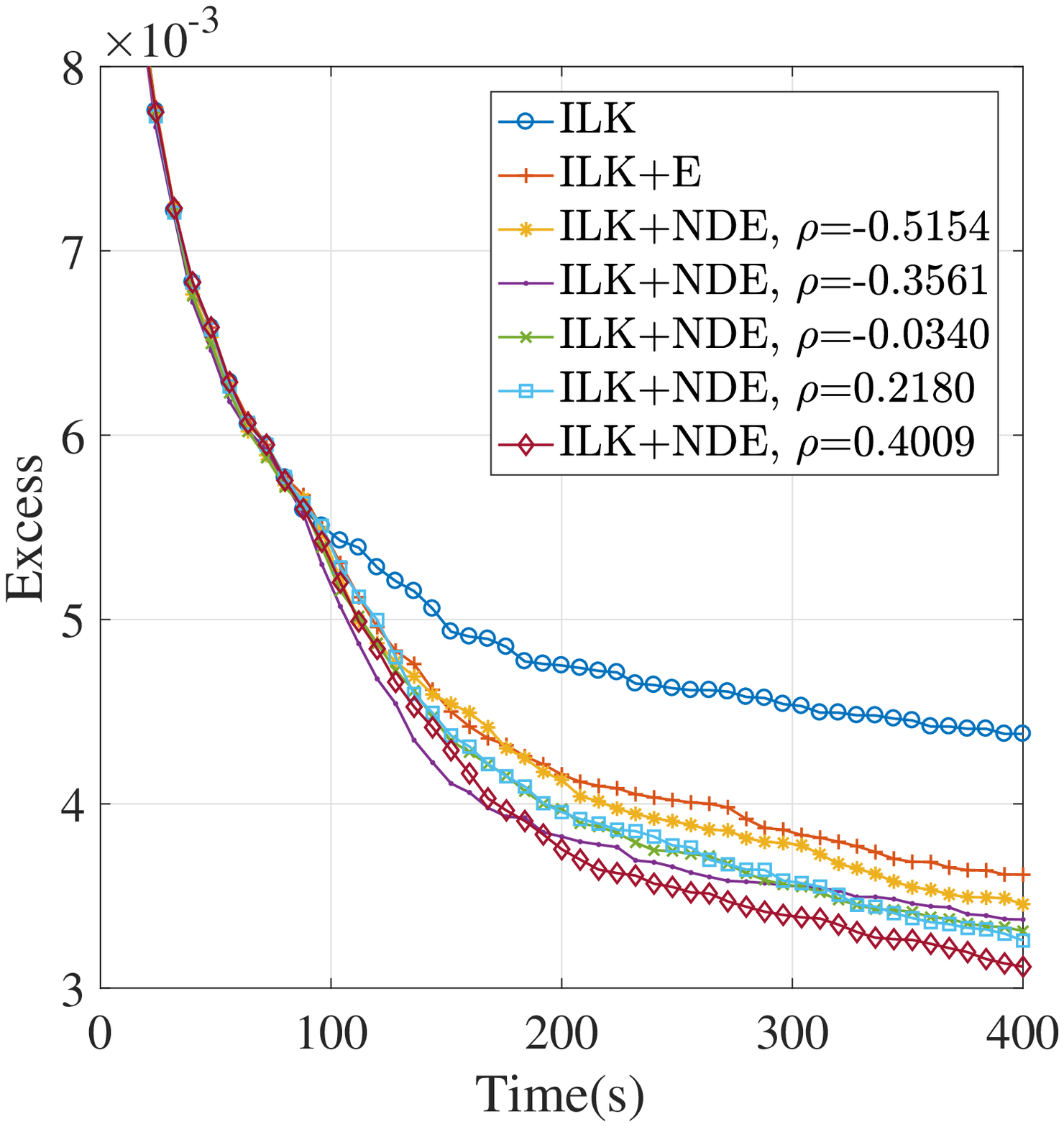}}
  \subfigure[d2103]{
    \label{fig:excess_LK_d2103}
    \includegraphics[height=0.30\linewidth]{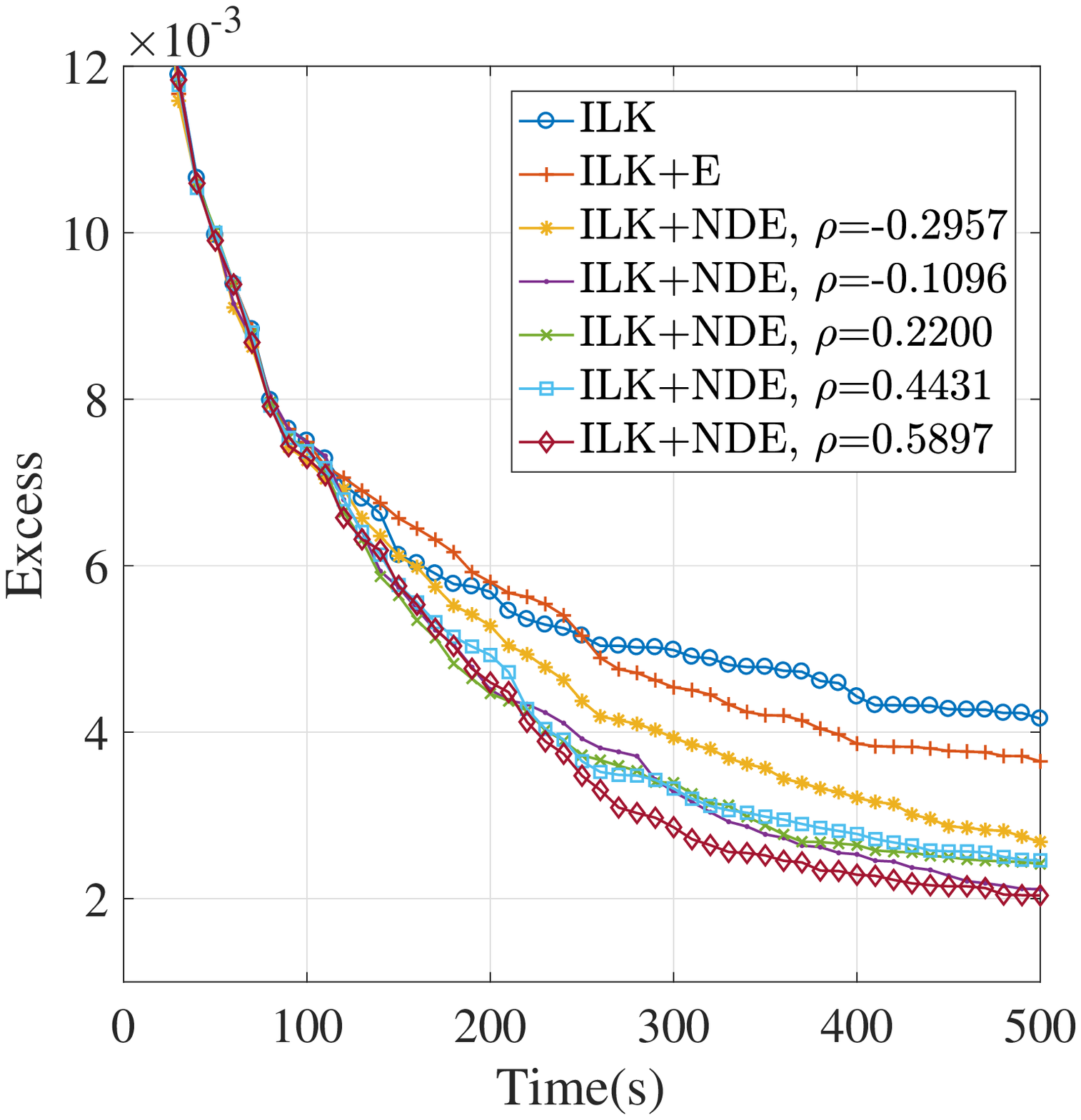}}\\
  \subfigure[pr2392]{
    \label{fig:excess_LK_pr2392}
    \includegraphics[height=0.30\linewidth]{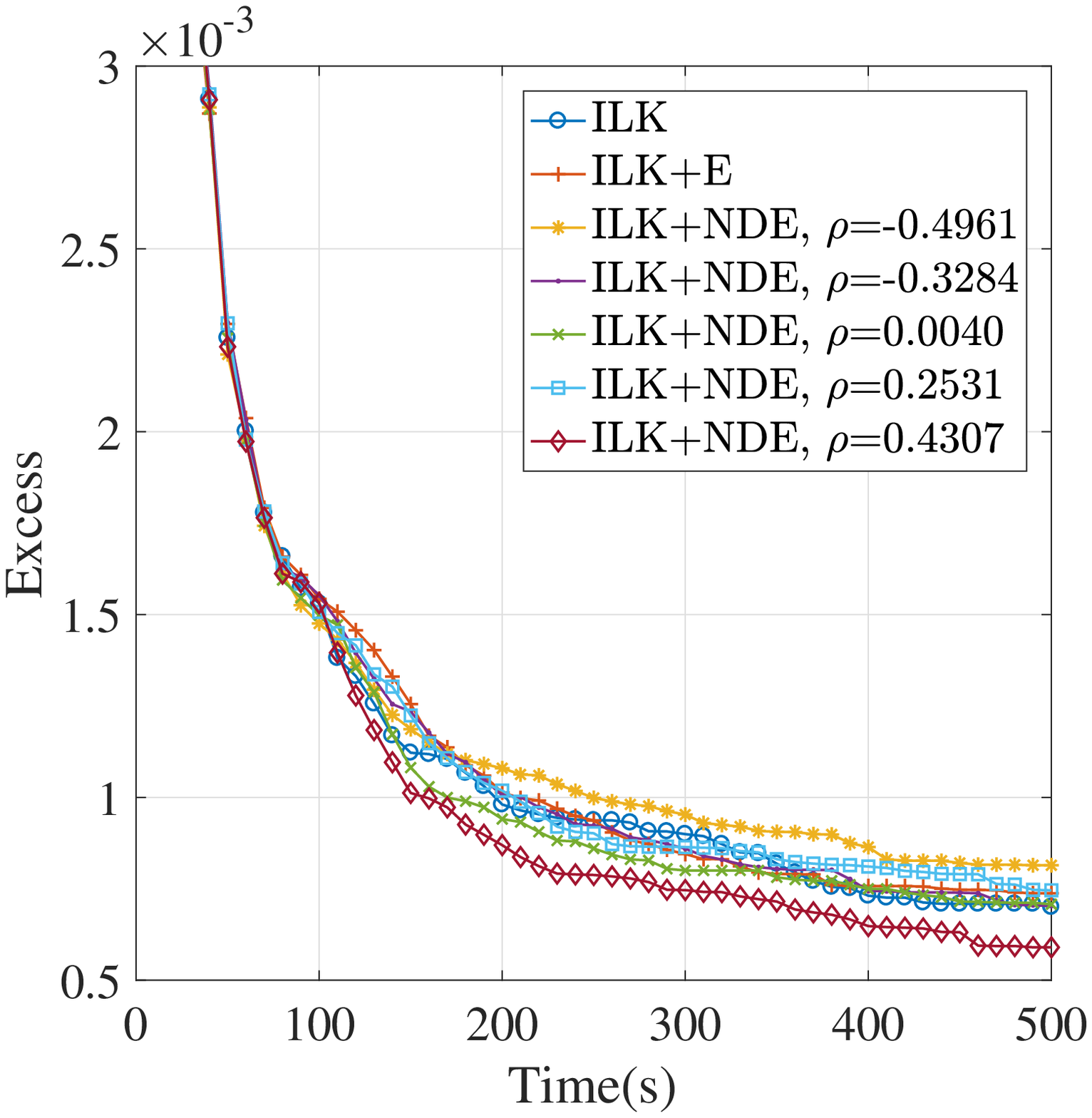}}
  \subfigure[pcb3038]{
    \label{fig:excess_LK_pcb3038}
    \includegraphics[height=0.30\linewidth]{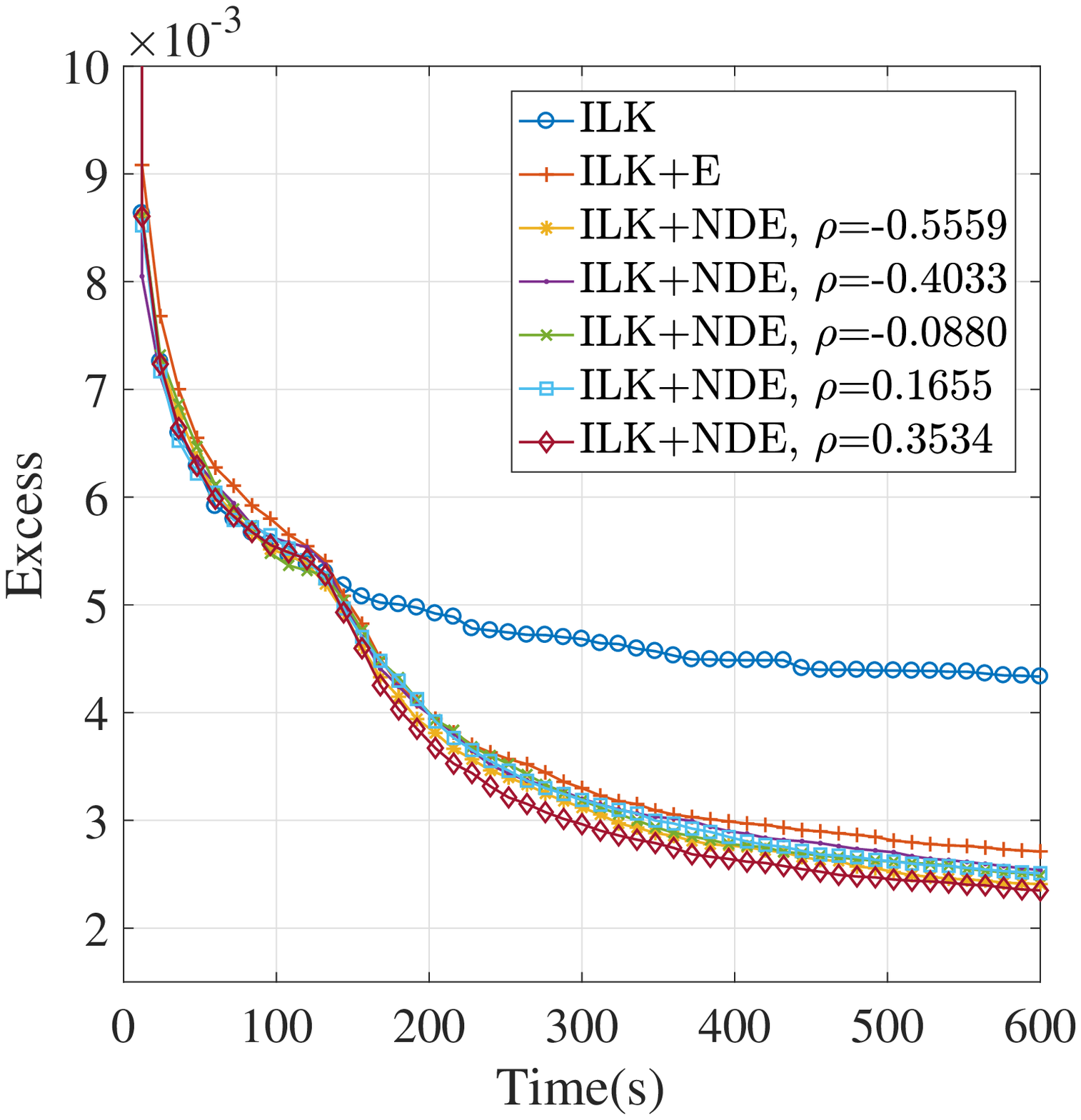}}
  \subfigure[fnl4461]{
    \hspace{-0.01\linewidth}
    \label{fig:excess_LK_fnl4461}
    \includegraphics[height=0.30\linewidth]{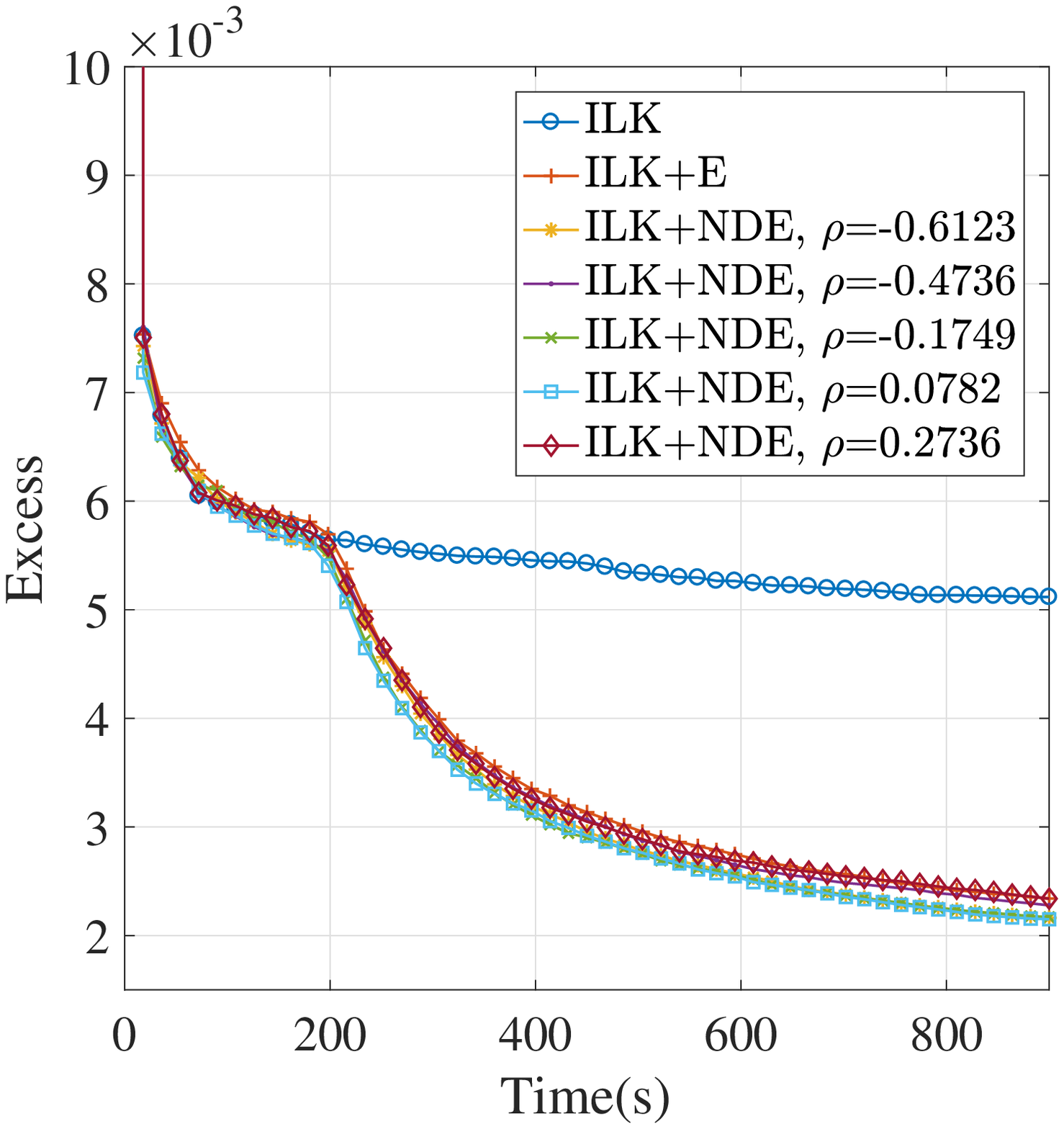}}\\
  \subfigure[pla7397]{
    \label{fig:excess_LK_pla7397}
    \includegraphics[height=0.30\linewidth]{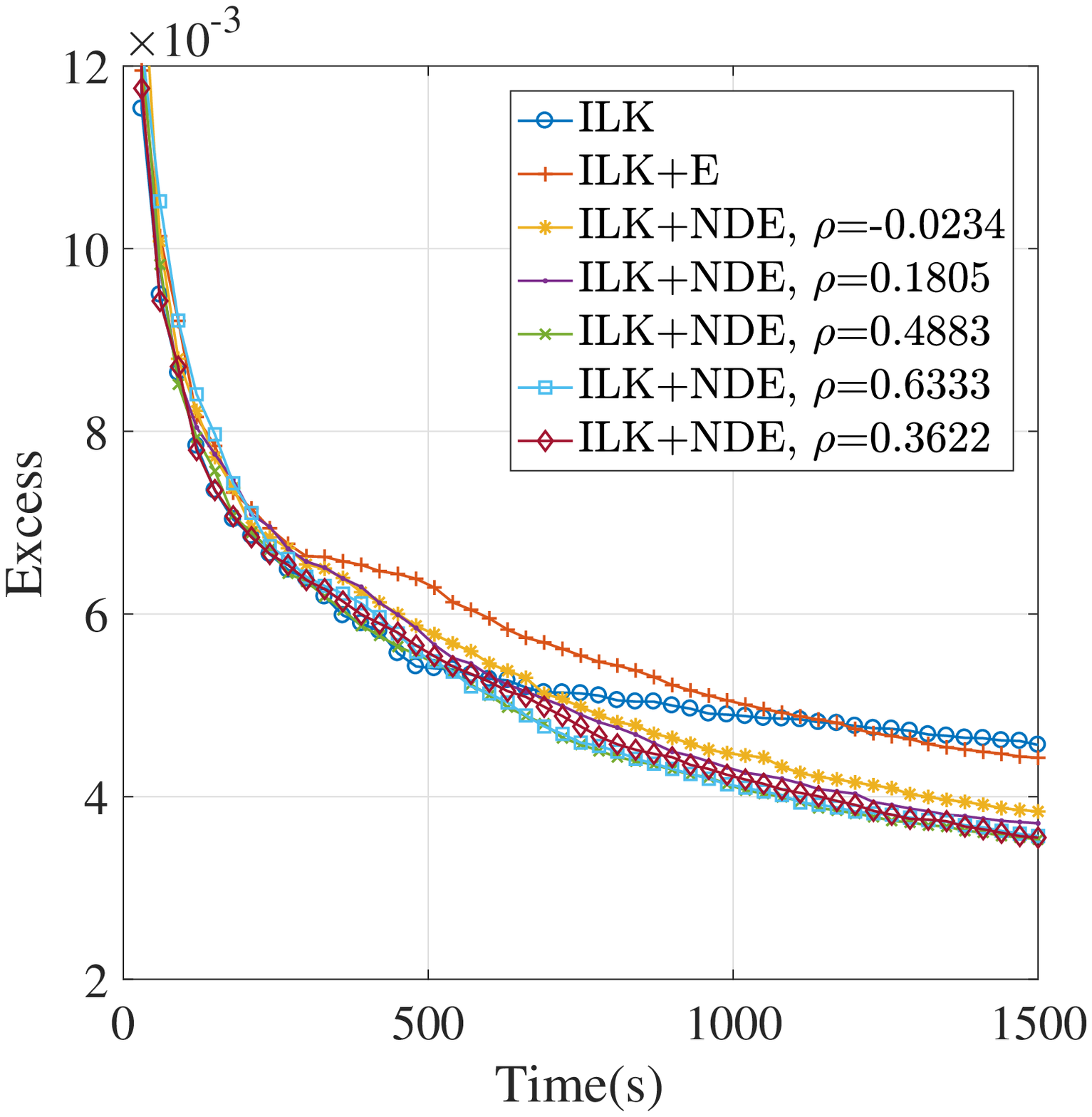}}
  \subfigure[rl11849]{
    \hspace{-0.01\linewidth}
    \label{fig:excess_LK_rl11849}
    \includegraphics[height=0.30\linewidth]{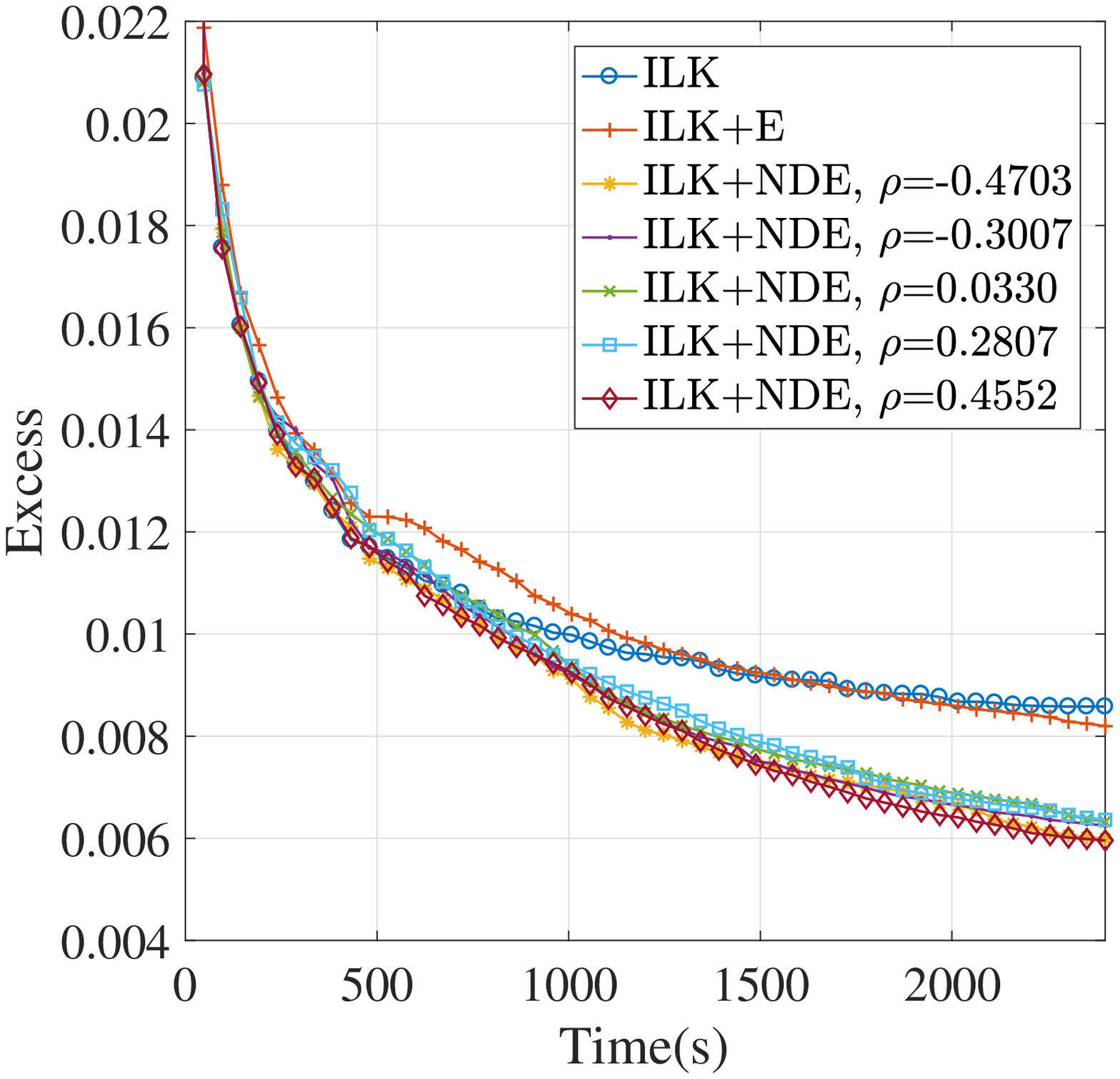}}
  \subfigure[usa13509]{
    \label{fig:excess_LK_usa13509}
    \includegraphics[height=0.30\linewidth]{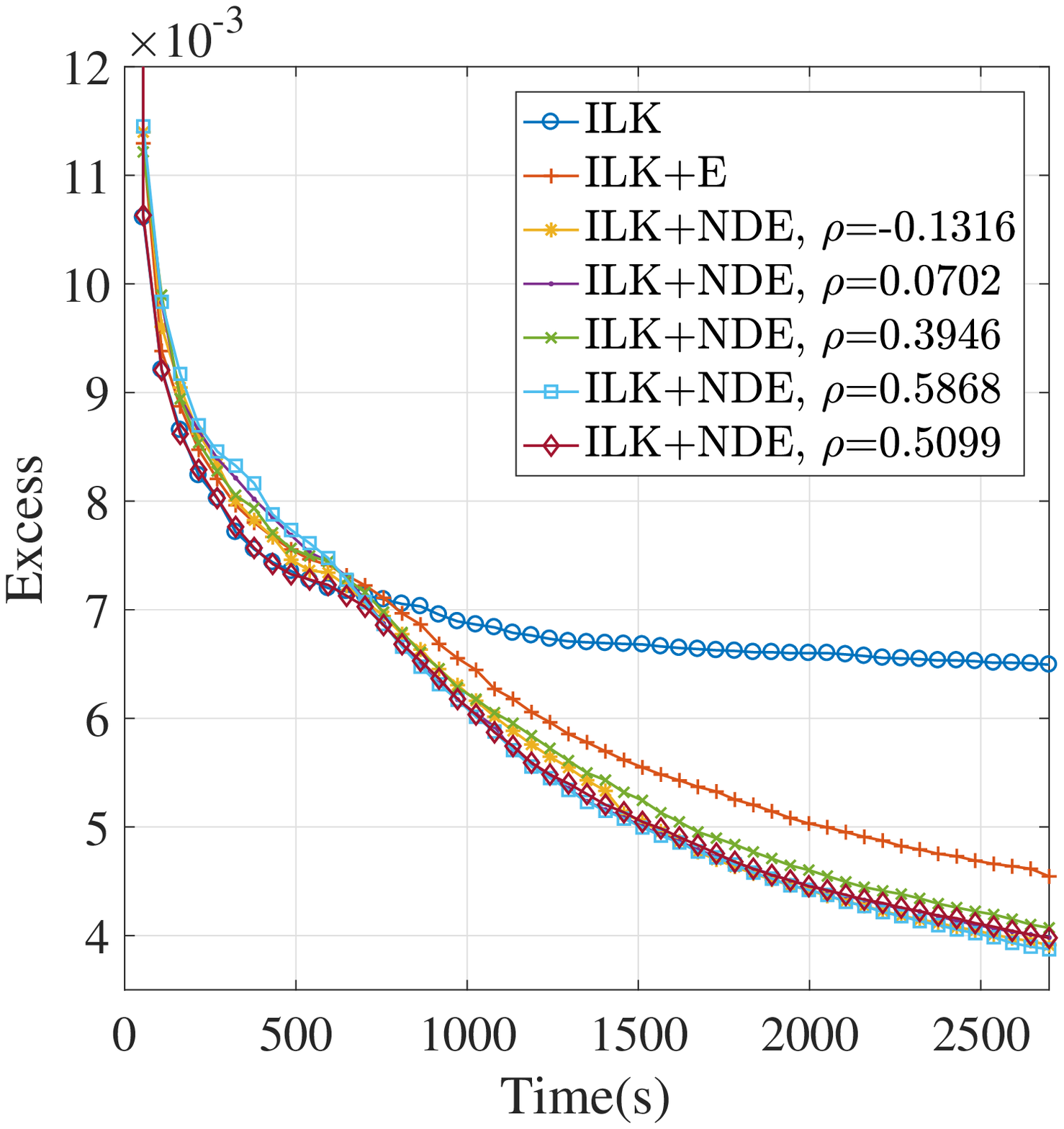}}\\
  \subfigure[d18512]{
    \hspace{-0.01\linewidth}
    \label{fig:excess_LK_d18512}
    \includegraphics[height=0.30\linewidth]{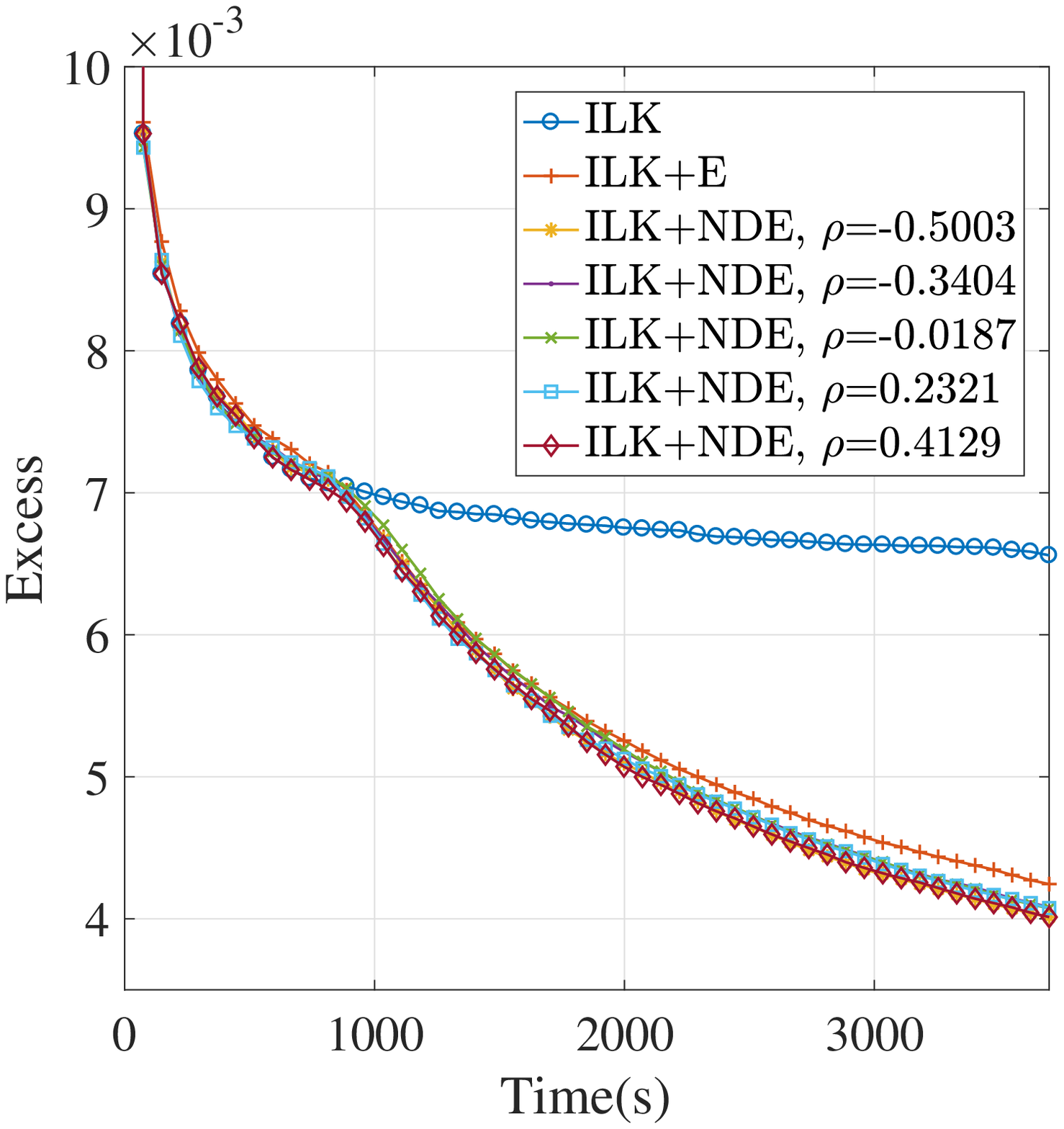}}
 \caption{Excess vs time on 10 middle-size and large-size large TSP instances}\label{fig:excess_LK}
\end{figure}

Figure~\ref{fig:excess_LK} shows the mean excess achieved by different algorithms against time. Table~\ref{tbl:ILK_final_excess} shows the obtained final excess values. From Figure~\ref{fig:excess_LK} and Table~\ref{tbl:ILK_final_excess} we can see that on all test instances, ILK+NDE performs the best. In Table~\ref{tbl:ILK_final_excess}, the superiority of ILK+NDE is confirmed by the Mann-Whitney U-test at a significance level of $0.05$. This results show that the LK local search can benefit from the sub-objectives $(f_1, f_2)$ on middle-size and large-size TSP instances.

Based the above experimental study, we can conclude that the proposed NDE method can truly improve the original version of ILK.

\begin{table}
\caption{Final excess of ILK, ILK+E and ILK+NDE on 10 middle-size and large-size TSP instances.}
\centering
\label{tbl:ILK_final_excess}
\resizebox{\linewidth}{!}{
\begin{tabular}{l| c c c | c | c }
\hline
\multirow{2}{*}{Instance} & \multicolumn{3}{c|}{Mean final excess (standard deviation)} & \multirow{2}{*}{best $\rho$ in ILK+NDE}  & \multirow{2}{*}{corresponding $a$}\\
\cline{2-4}
 & ILK & ILK+E & ILK+NDE & & \\
 \hline
TSP:vm1748 & 0.00191(0.00041) $-$ & 0.00121(0.00051) = & 0.00115(0.00056) & 0.4619 & 2\\
\hline
TSP:u1817 & 0.00438(0.00061) $-$ & 0.00362(0.00103) $-$ & 0.00312(0.00102) & 0.4009 & 3 \\
\hline
TSP:d2103 & 0.00416(0.00189) $-$ & 0.00365(0.00293) $-$ & 0.00204(0.00167) & 0.5897 & 3 \\
\hline
TSP:pr2392 & 0.00070(0.00026) $-$ & 0.00074(0.00071) = & 0.00059(0.00057) & 0.4307 & 3 \\
\hline
TSP:pcb3038 & 0.00433(0.00049) $-$ & 0.00271(0.00084) $-$ & 0.00235(0.00073) & 0.3534 & 3 \\
\hline
TSP:fnl4461 & 0.00512(0.00034) $-$ & 0.00234(0.00051) $-$ & 0.00215(0.00042) & 0.0782 & 2 \\
\hline
TSP:pla7397 & 0.00457(0.00071) $-$ & 0.00443(0.00172) $-$ & 0.00353(0.00101) & 0.4883 & 1\\
\hline
TSP:rl11849 & 0.00858(0.00106) $-$ & 0.00820(0.00387) $-$ & 0.00596(0.00169) & 0.4552 & 3\\
\hline
TSP:usa13509 & 0.00649(0.00042) $-$ & 0.00454(0.00133) $-$ & 0.00387(0.00052) & 0.5868 & 2\\
\hline
TSP:d18512 & 0.00656(0.00022) $-$ & 0.00424(0.00060) $-$ & 0.00401(0.00031) & $-$0.5003 & $-$1\\
\hline
\multicolumn{6}{c}{\begin{minipage}{590pt}*If there are multiple best $\rho$ values, the first one is listed. ``$-$'' means the final excess achieved by ILK and ILK+E is significantly worse than that achieved by ILK+NDE based on a Mann-Whitney U-test at the 0.05 significance level and ``$=$'' means the performance is the same.\end{minipage}}
\end{tabular}
}
\end{table}

\subsection{The Performance of ITS+NDS}
In this section we apply the proposed NDS method to the Iterated Tabu Search (ITS) algorithm~\cite{palubeckis2006iterated}. The basic procedure of ITS is shown in Algorithm~\ref{alg:ITS}. The ITS variant that enhanced by the NDS method is named ITS+NDS, and its procedure is shown in Algorithm~\ref{alg:ITS+NDS}. In Algorithm~\ref{alg:ITS} and Algorithm~\ref{alg:ITS+NDS}, TabuSearch($x$) means executing a tabu search process from $x$ until a pre-defined stopping criteria is met and output the best solution during the tabu search.

\begin{algorithm}
\small
    Decompose $f$ into $f_1$ and $f_2$\; 
    $x_0' \gets $ random or heuristically generated solution\;
    set $x_{best} \gets x_0'$ and $j \gets 0$\;
    \While{stopping criterion is not met}{
        $x_j \gets$ TabuSearch($x_j'$)\;
        \If {$f(x_{j}) < f(x_{best})$} {
            $x_{best} \gets x_{j}$\;
        }
        $x_{j+1}' \gets$ NDS($x_j\mid f,f_1,f_2$)\;
        \If {$x_{j+1}'=x_j$}{
            $x_{j+1}' \gets$ Perturbation($x_j$)\;
        }
        $j\gets j+1$\;
    }
    \KwRet{\mbox{the historical best solution} $x_{best}$}
\caption{ITS+NDS}
\label{alg:ITS+NDS}
\end{algorithm}

In this section, we compare ITS+NDS against ITS on 5 UBQP instances:\{bqp1000.1, bqp2500.1, p3000.1, p4000.1, p5000.1\}, which are same to the UBQP instances in Figure~\ref{fig:5UBQP_excess}. In our experiment, the TS procedure is based on $1$-bit-flip moves. At each move, TS checks the flip of variables that are not contained in the tabu list, and selects the move that leads to the best neighboring solution. The flipped variable in the selected neighboring solution is used to update the tabu list, and is forbidden to be flipped until a number of $K$ moves have elapsed. Following~\cite{glover2010diversification}, in our experiment, $K$ is sampled from the uniform distribution over $[n/100+1, n/100+10]$, and the TS process stops when the best solution cannot be improved within $20n$ moves, where $n$ is the problem size. For the ITS+NDS implementation, on each UBQP instance, we still use the eight pairs of $(f_1, f_2)$ listed in Table~\ref{tbl:nei_type_UBQP}. The repeated run number for each algorithm is $50$ and the maximum function evaluation number is $10^{10}$ for each run. The other settings are the same to the settings in the experiment of Figure~\ref{fig:5UBQP_excess} in Section~\ref{sec:epm_NDS}.

\begin{figure}
  \subfigure[UBQP: bqp1000.1]{
    \label{fig:ITS_excess_bqp1000_1}
    \includegraphics[height=0.35\linewidth]{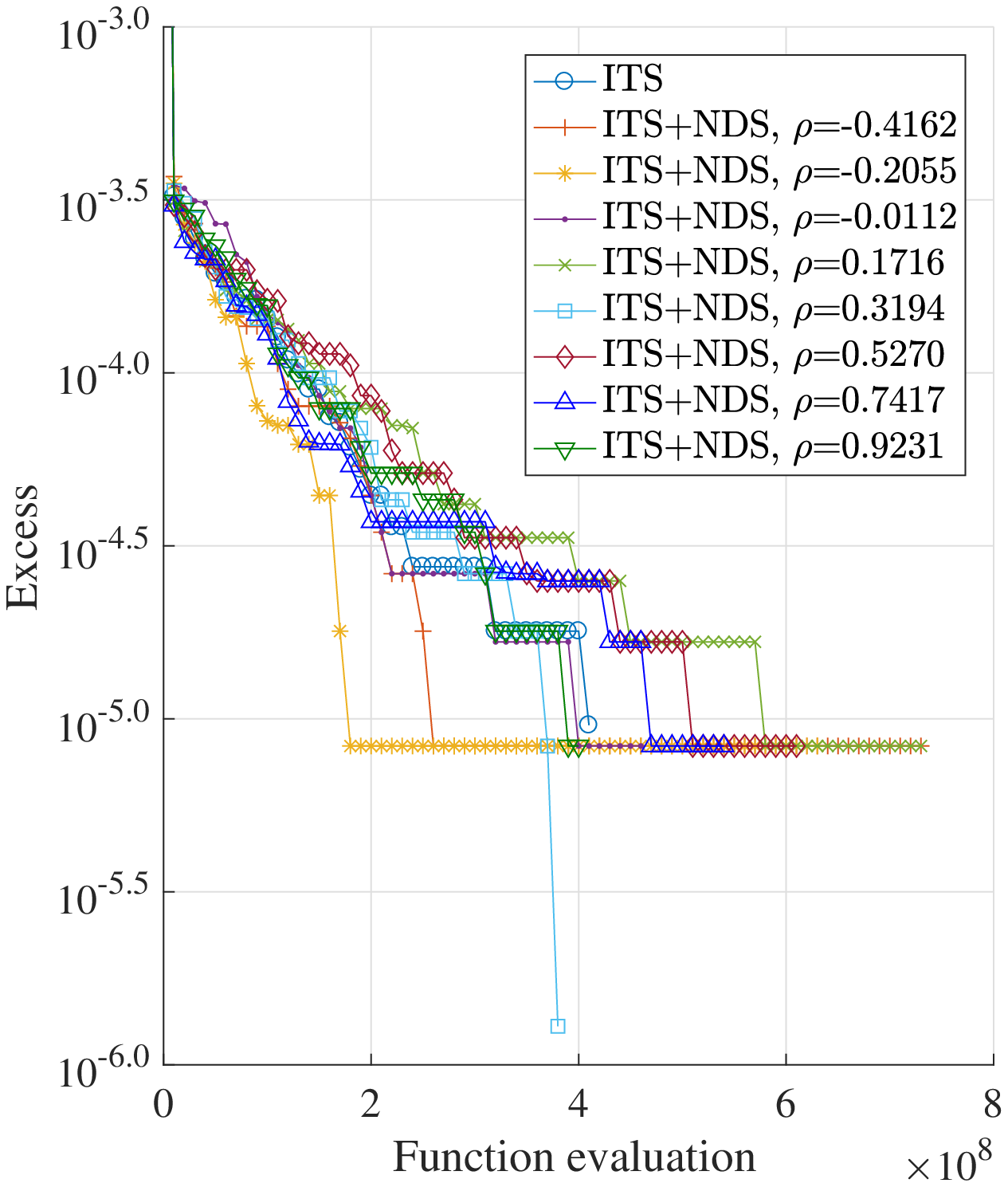}}
    \hspace{-0.01\linewidth}
  \subfigure[UBQP: bqp2500.1]{
    \label{fig:ITS_excess_bqp2500_1}
    \includegraphics[height=0.35\linewidth]{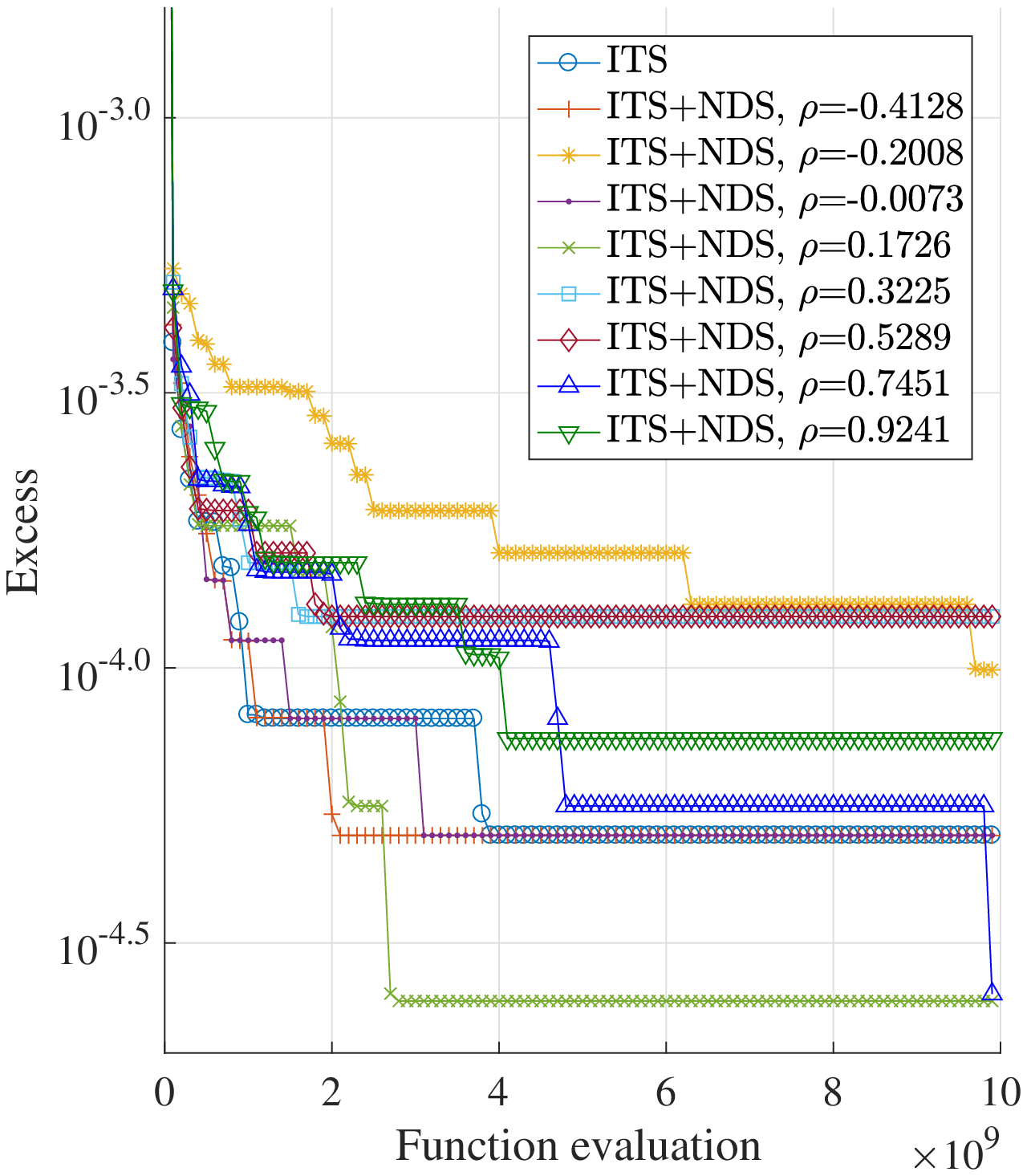}}
    \hspace{-0.01\linewidth}
  \subfigure[UBQP: p3000.1]{
    \label{fig:ITS_excess_p3000_1}
    \includegraphics[height=0.35\linewidth]{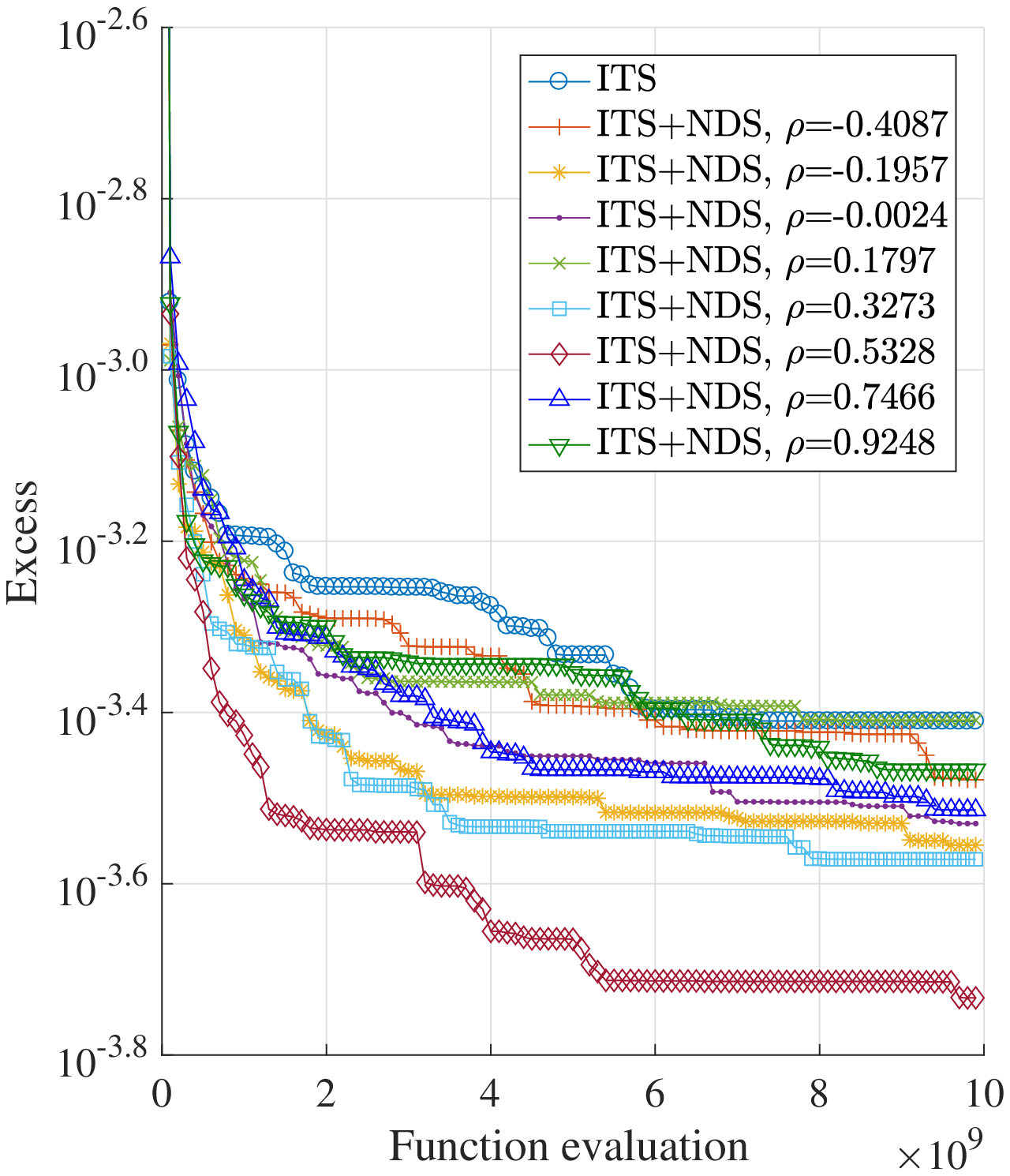}}\\
  \subfigure[UBQP: p4000.1]{
    \label{fig:ITS_excess_p4000_1}
    \includegraphics[height=0.35\linewidth]{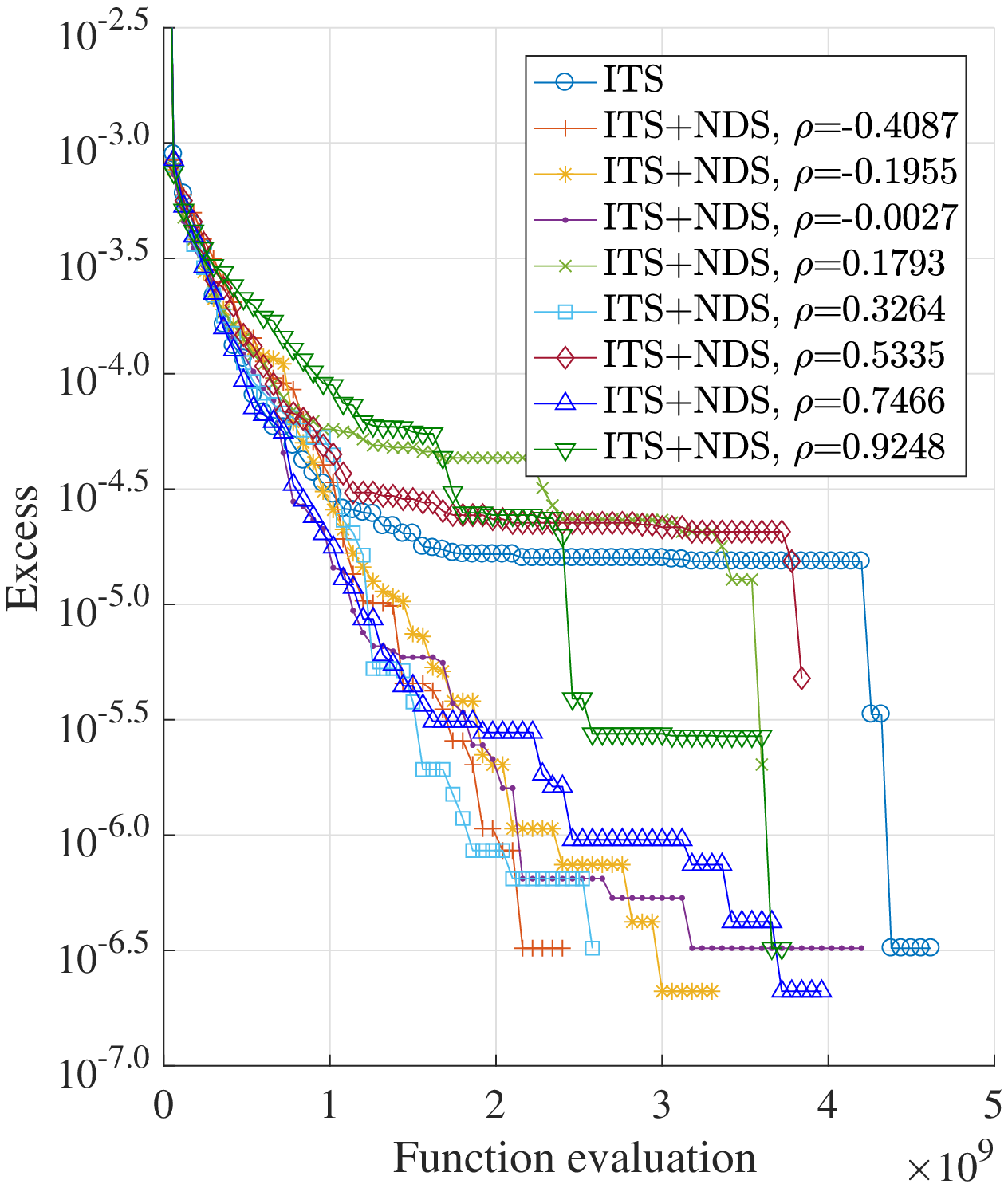}}
    \hspace{-0.01\linewidth}
  \subfigure[UBQP: p5000.1]{
    \label{fig:ITS_excess_p5000_1}
    \includegraphics[height=0.35\linewidth]{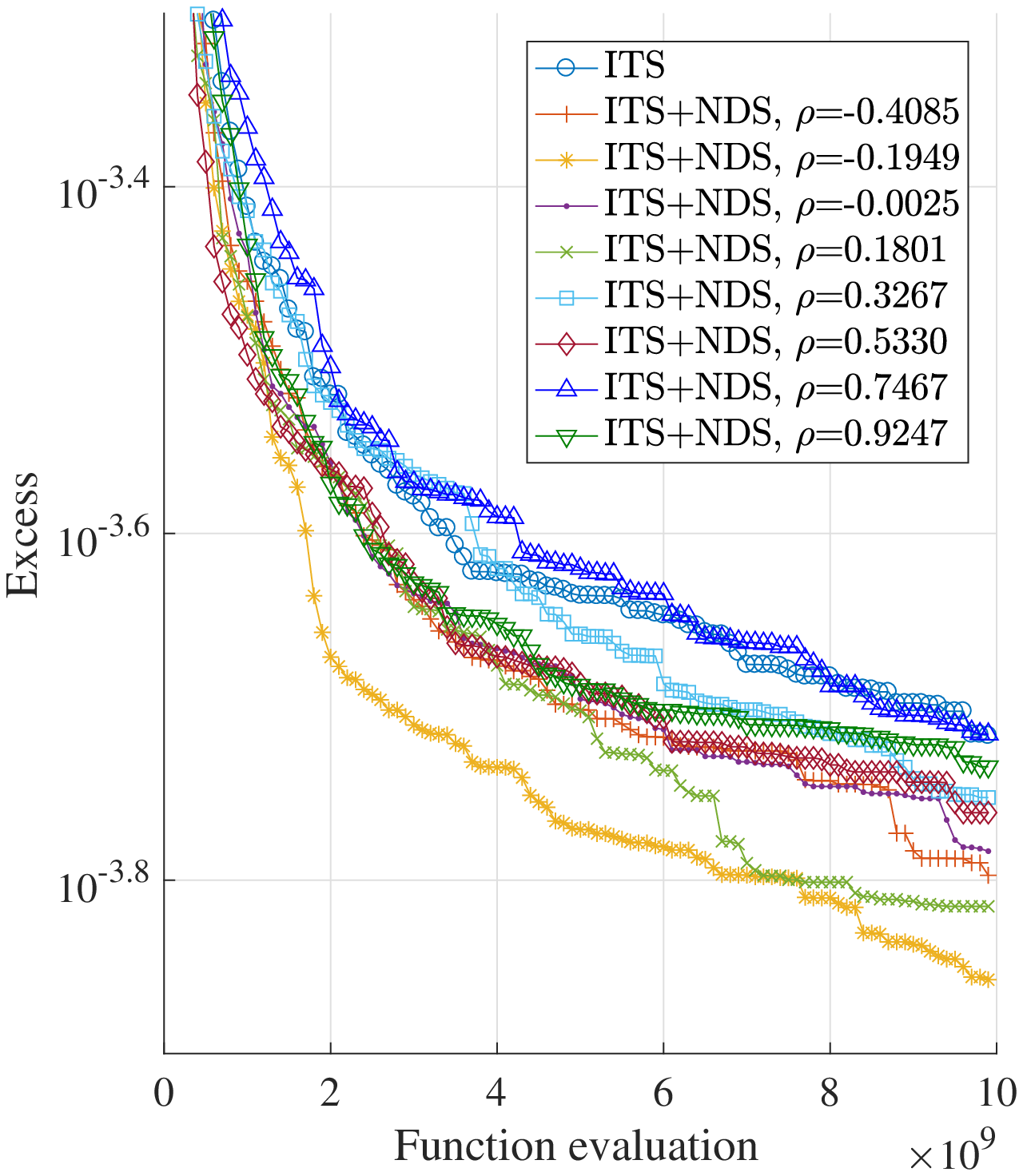}}\\
 \caption{ITS vs. ITS+NDS on 5 UBQP instances}\label{fig:ITS_excess}
\end{figure}

Figure~\ref{fig:ITS_excess} shows the mean excess achieved by ITS and ITS+NDS against time. Table~\ref{tbl:ITS_final_excess} lists the mean value and standard deviation of the final excess achieved by ITS and ITS+NDS, in which the excess result of ITS+NDS is the best excess selected from the eight settings of $\rho$. From Figure~\ref{fig:ITS_excess} we can see that on all the test instances, the best anytime performance is achieved by ILS+NDS. At most setups of $\rho$, ITS+NDS performs better than ITS. From Table~\ref{tbl:ITS_final_excess} we can see that in the end the excess achieved by ITS+NDS is significantly better than that of ITS on one UBQP instance, p3000.1. On the other UBQP instances, the statistical test suggests that the two algorithms perform the same. However, Table~\ref{tbl:ITS_final_excess} only considers the final excess values. From Figure~\ref{fig:ITS_excess} we can see that on bqp1000.1 and p4000.1 most curves of ITS+NDS terminate early than that of ITS, which means that ITS+NDS finds the global optima of bqp1000.1 and p4000.1 early than ITS. Hence, we can state that ITS+NDS performs better than ITS on three UBQP instances and in the rest two instances they perform the same.

Based the above experimental study, we can conclude that the proposed NDS method can truly improve the original version of ITS.

\begin{table}
\caption{Final excess of ITS and ITS+NDS on 5 UBQP instances.}
\centering
\label{tbl:ITS_final_excess}
\resizebox{\linewidth}{!}{
\begin{tabular}{l| c c | c | c }
\hline
\multirow{2}{*}{Instance} & \multicolumn{2}{c|}{Mean final excess (standard deviation)} & \multirow{2}{*}{best $\rho$ in ITS+NDS} & \multirow{2}{*}{corresponding $a$} \\
\cline{2-3}
 & ILS & ILS+NDS & \\
\hline
UBQP:bqp1000.1 & 0.00000(0.00000) = & 0.00000(0.00000) & $-$0.4162 & $-$15\\
\hline
UBQP:bqp2500.1 & 0.00005(0.00025) = & \textbf{0.00002}(0.00018) & 0.1726 & 0.5 \\
\hline
UBQP:p3000.1 & 0.00039(0.00044) $-$ & \textbf{0.00018}(0.00039) & 0.5328 & 2\\
\hline
UBQP:p4000.1 & 0.00000(0.00000) = & 0.00000(0.00000) & $-$0.4087 & $-$15\\
\hline
UBQP:p5000.1 & 0.00019(0.00021) = & \textbf{0.00014}(0.00014) & $-$0.1949 & $-$3 \\
\hline
\multicolumn{5}{c}{\begin{minipage}{500pt}*If there are multiple best $\rho$ values, the first one is listed. ``$-$'' means the final excess achieved by ITS is significantly worse than that achieved by ITS+NDS based on a Mann-Whitney U-test at the 0.05 significance level and ``$=$'' means the performance is the same.\end{minipage}}
\end{tabular}
}
\end{table}

\section{Conclusions} \label{sec:conclu}


In this paper, we proposed a new objective decomposition method which is suitable for a certain subclass of COPs, which we called the sum-of-the-parts COPs. We gave the formalization of the sum-of-the-parts COP and showed that the TSP and the UBQP belong to this class. The proposed method decomposes the objective function of a sum-of-the-parts COP into two sub-objectives by splitting the unit costs following a certain probability distribution. It was shown that the correlation between the decomposed sub-objectives can be controlled by the use of the probability distribution.

Based on the non-dominance relationship introduced by the decomposed sub-objectives, we proposed two new multi-objectivization inspired techniques.

The first was called Non-Dominance Search (NDS). NDS can be used as an escaping scheme from local optima for metaheuristics with fixed neighborhood structure. NDS is based on our neighborhood non-dominance hypothesis which states that the neighborhood of a non-dominated neighboring solution of a local optimum is more likely to improve the local optimum. Empirical studies on some selected TSP and UBQP instances confirm that the hypothesis holds. NDS was combined within the Iterated Local Search and Iterated Tabu Search. The resultant metaheuristics are called ILS+NDS and ITS+NDS. Experimental results on some TSP and UBQP instances showed that ILS+NDS and ITS+NDS outperform their counterparts in most cases.

The second is called Non-Dominance Exploitation (NDE), which is applicable for metaheuristics with varied neighborhood structure, such as the Lin-Kernighan (LK) heuristic for the TSP. NDE is proposed to exploit promising local optima based on the non-dominance relationship. NDE was combined with the Iterated Lin-Kernighan algorithm (ILK), called ILK+NDE. Experimental results on middle-size and large-size TSP instances showed that ILK+NDE significantly outperform the original ILK and ILK+E.


In the future, we intend to test the performance of the proposed objective decomposition method on other sum-of-the-parts COPs. 

%



\section*{References}

\bibliographystyle{elsarticle-num}


%

\end{document}